%% file: aij-digital-design.tex
\documentclass[authoryear]{elsarticle}

\usepackage{graphicx}
\usepackage{hyperref}
\usepackage{threeparttable}
\usepackage{tabularx}
\usepackage{standalone}
\usepackage{circuitikz}
\usepackage{subcaption}
\usepackage{mathtools}
\usepackage{booktabs}
\usepackage{paralist}
\usepackage{rotating}
\usepackage{pgfplots}
\usepackage{dcolumn}
\usepackage{amssymb}
\usepackage{amsmath}
\usepackage{amsthm}
\usepackage{xspace}
\usepackage{bbold}
\usepackage{tikz}
\usepackage{quantikz}
\usepackage[inline]{enumitem}
\usepackage[binary-units=true]{siunitx}
\usepackage[ruled]{algorithm2e}

\usepgfplotslibrary{groupplots}
\pgfplotsset{compat=1.17}

\usetikzlibrary{positioning}
\usetikzlibrary{matrix}
\usetikzlibrary{shapes}
\usetikzlibrary{arrows}

\makeatletter
\definecolor{antiquewhite}{rgb}{0.98, 0.92, 0.84}
\definecolor{champagne}{rgb}{0.97, 0.91, 0.81}
\definecolor{darkkhaki}{rgb}{0.74, 0.72, 0.42}
\definecolor{pastelgray}{rgb}{0.81, 0.81, 0.77}
\definecolor{whitesmoke}{rgb}{0.96, 0.96, 0.96}
\definecolor{tearose(rose)}{rgb}{0.96, 0.76, 0.76}
 \makeatletter

\ctikzset{tripoles/and gate/width/.initial=1.1}
\ctikzset{tripoles/and gate/height/.initial=.8}
\ctikzset{tripoles/and gate/port width/.initial=.7}
\ctikzset{tripoles/and gate/input height/.initial=.5}
\ctikzset{tripoles/nand gate/width/.initial=1.1}
\ctikzset{tripoles/nand gate/height/.initial=.8}
\ctikzset{tripoles/nand gate/port width/.initial=.7}
\ctikzset{tripoles/nand gate/circle width/.initial=.15}
\ctikzset{tripoles/nand gate/input height/.initial=.5}
\ctikzset{tripoles/or gate/width/.initial=1.1}
\ctikzset{tripoles/or gate/height/.initial=.8}
\ctikzset{tripoles/or gate/port width/.initial=.7}
\ctikzset{tripoles/or gate/input height/.initial=.5}
\ctikzset{tripoles/or gate/input skip/.initial=.25}
\ctikzset{tripoles/or gate/aaa/.initial=.6}
\ctikzset{tripoles/or gate/bbb/.initial=.4}
\ctikzset{tripoles/or gate/ccc/.initial=.5}
\ctikzset{tripoles/or gate/ddd/.initial=.0}
\ctikzset{tripoles/nor gate/width/.initial=1.1}
\ctikzset{tripoles/nor gate/height/.initial=.8}
\ctikzset{tripoles/nor gate/port width/.initial=.7}
\ctikzset{tripoles/nor gate/input height/.initial=.5}
\ctikzset{tripoles/nor gate/input skip/.initial=.25}
\ctikzset{tripoles/nor gate/circle width/.initial=.15}
\ctikzset{tripoles/nor gate/aaa/.initial=.6}
\ctikzset{tripoles/nor gate/bbb/.initial=.4}
\ctikzset{tripoles/nor gate/ccc/.initial=.5}
\ctikzset{tripoles/nor gate/ddd/.initial=.0}
\ctikzset{tripoles/xor gate/width/.initial=1.1}
\ctikzset{tripoles/xor gate/height/.initial=.8}
\ctikzset{tripoles/xor gate/port width/.initial=.7}
\ctikzset{tripoles/xor gate/input height/.initial=.5}
\ctikzset{tripoles/xor gate/input skip/.initial=.15}
\ctikzset{tripoles/xor gate/distance/.initial=.1}
\ctikzset{tripoles/xor gate/aaa/.initial=.6}
\ctikzset{tripoles/xor gate/bbb/.initial=.4}
\ctikzset{tripoles/xor gate/ccc/.initial=.5}
\ctikzset{tripoles/xor gate/ddd/.initial=.0}
\ctikzset{tripoles/xnor gate/width/.initial=1.1}
\ctikzset{tripoles/xnor gate/height/.initial=.8}
\ctikzset{tripoles/xnor gate/port width/.initial=.7}
\ctikzset{tripoles/xnor gate/input height/.initial=.5}
\ctikzset{tripoles/xnor gate/input skip/.initial=.15}
\ctikzset{tripoles/xnor gate/distance/.initial=.1}
\ctikzset{tripoles/xnor gate/aaa/.initial=.6}
\ctikzset{tripoles/xnor gate/bbb/.initial=.4}
\ctikzset{tripoles/xnor gate/ccc/.initial=.5}
\ctikzset{tripoles/xnor gate/ddd/.initial=.0}
\ctikzset{tripoles/xnor gate/circle width/.initial=.15}
\ctikzset{bipoles/buffer/height/.initial=1}
\ctikzset{bipoles/buffer/width/.initial=1}
\ctikzset{bipoles/not gate/width/.initial=1}
\ctikzset{bipoles/not gate/height/.initial=.8}
\ctikzset{bipoles/not gate/circle width/.initial=.15}

\ctikzset{thickness/.initial=2}
\ctikzset{color/.initial=black}
\ctikzset{fill/.initial=white}
\pgfkeys{/tikz/color/.add code={}{\ctikzset{color={#1}}}}
\pgfkeys{/tikz/fill/.add code={}{\ctikzset{fill={#1}}}}

\long\def\pgfcircdeclarelogicgate#1#2{
  \pgfdeclareshape{#1 gate}{
    \savedanchor\northwest{\pgf@y=\pgfkeysvalueof{/tikz/circuitikz/tripoles/#1 gate/height}\pgf@circ@Rlen
      \pgf@y=.5\pgf@y
      \pgf@x=-\pgfkeysvalueof{/tikz/circuitikz/tripoles/#1 gate/width}\pgf@circ@Rlen
      \pgf@x=.5\pgf@x
    }
    \anchor{in 1}{
      \pgf@y=\pgfkeysvalueof{/tikz/circuitikz/tripoles/#1 gate/height}\pgf@circ@Rlen
      \pgf@y=.5\pgf@y
      \pgf@y=\pgfkeysvalueof{/tikz/circuitikz/tripoles/#1 gate/input height}\pgf@y
      \pgf@x=-\pgfkeysvalueof{/tikz/circuitikz/tripoles/#1 gate/width}\pgf@circ@Rlen
      \pgf@x=.5\pgf@x
    }
    \anchor{in 2}{
      \pgf@y=\pgfkeysvalueof{/tikz/circuitikz/tripoles/#1 gate/height}\pgf@circ@Rlen
      \pgf@y=.5\pgf@y
      \pgf@y=\pgfkeysvalueof{/tikz/circuitikz/tripoles/#1 gate/input height}\pgf@y
      \pgf@x=-\pgfkeysvalueof{/tikz/circuitikz/tripoles/#1 gate/width}\pgf@circ@Rlen
      \pgf@x=.5\pgf@x
      \pgf@y=-\pgf@y
    }
    \anchor{out}{
      \northwest
      \pgf@y=0pt
      \pgf@x=-\pgf@x
    }
    \anchor{center}{
      \northwest
      \pgf@y=0pt
      \pgf@x=0pt
    }
    \backgroundpath{
      \pgfsetcolor{\pgfkeysvalueof{/tikz/circuitikz/color}}

      \northwest
      \pgf@circ@res@up=\pgf@y
      \pgf@circ@res@down=-\pgf@y
      \pgf@circ@res@right=-\pgf@x
      \pgf@circ@res@left=\pgf@x

      #2

    }
  }
}

\pgfcircdeclarelogicgate{and}{
  \pgfpathmoveto{\pgfpoint{\pgf@circ@res@left}
                          {\pgfkeysvalueof{/tikz/circuitikz/tripoles/and gate/input height}\pgf@circ@res@up}}
  \pgfpathlineto{\pgfpoint{\pgfkeysvalueof{/tikz/circuitikz/tripoles/and gate/port width}\pgf@circ@res@left}
                          {\pgfkeysvalueof{/tikz/circuitikz/tripoles/and gate/input height}\pgf@circ@res@up}}

  \pgfpathmoveto{\pgfpoint{\pgf@circ@res@left}
                          {\pgfkeysvalueof{/tikz/circuitikz/tripoles/and gate/input height}\pgf@circ@res@down}}
  \pgfpathlineto{\pgfpoint{\pgfkeysvalueof{/tikz/circuitikz/tripoles/and gate/port width}\pgf@circ@res@left}
                          {\pgfkeysvalueof{/tikz/circuitikz/tripoles/and gate/input height}\pgf@circ@res@down}}

  \pgfpathmoveto{\pgfpoint{\pgf@circ@res@right}{0pt}}
  \pgfpathlineto{\pgfpoint{\pgfkeysvalueof{/tikz/circuitikz/tripoles/and gate/port width}\pgf@circ@res@right}{0pt}}

  \pgfusepath{draw}

  \pgfsetlinewidth{2\pgflinewidth}
  \pgfpathmoveto{\pgfpoint{\pgfkeysvalueof{/tikz/circuitikz/tripoles/and gate/port width}\pgf@circ@res@left}{\pgf@circ@res@down}}
  \pgfpathcurveto{\pgfpoint{.0\pgf@circ@res@left}{\pgf@circ@res@down}}
                 {\pgfpoint{\pgfkeysvalueof{/tikz/circuitikz/tripoles/and gate/port width}\pgf@circ@res@right}{.5\pgf@circ@res@down}}
                 {\pgfpoint{\pgfkeysvalueof{/tikz/circuitikz/tripoles/and gate/port width}\pgf@circ@res@right}{0pt}}
  \pgfpathcurveto{\pgfpoint{\pgfkeysvalueof{/tikz/circuitikz/tripoles/and gate/port width}\pgf@circ@res@right}{.5\pgf@circ@res@up}}
                 {\pgfpoint{.0\pgf@circ@res@left}{\pgf@circ@res@up}}
                 {\pgfpoint{\pgfkeysvalueof{/tikz/circuitikz/tripoles/and gate/port width}\pgf@circ@res@left}{\pgf@circ@res@up}}
  \pgfclosepath
  \pgfsetcolor{\pgfkeysvalueof{/tikz/circuitikz/fill}}
  \pgfsetstrokecolor{black}
  \pgfusepath{draw,fill}
}

\pgfcircdeclarelogicgate{nand}{
  \pgfpathmoveto{\pgfpoint{\pgf@circ@res@left}{\pgfkeysvalueof{/tikz/circuitikz/tripoles/nand gate/input height}\pgf@circ@res@up}}
  \pgfpathlineto{\pgfpoint{\pgfkeysvalueof{/tikz/circuitikz/tripoles/nand gate/port width}\pgf@circ@res@left}{\pgfkeysvalueof{/tikz/circuitikz/tripoles/nand gate/input height}\pgf@circ@res@up}}

  \pgfpathmoveto{\pgfpoint{\pgf@circ@res@left}{\pgfkeysvalueof{/tikz/circuitikz/tripoles/nand gate/input height}\pgf@circ@res@down}}
  \pgfpathlineto{\pgfpoint{\pgfkeysvalueof{/tikz/circuitikz/tripoles/nand gate/port width}\pgf@circ@res@left}{\pgfkeysvalueof{/tikz/circuitikz/tripoles/nand gate/input height}\pgf@circ@res@down}}

  \pgfpathmoveto{\pgfpoint{\pgf@circ@res@right}{0pt}}
  \pgfpathlineto{\pgfpoint{\pgfkeysvalueof{/tikz/circuitikz/tripoles/nand gate/port width}\pgf@circ@res@right}{0pt}}

  \pgfusepath{draw}

  \pgfsetlinewidth{2\pgflinewidth}

  \pgf@circ@res@step=\pgfkeysvalueof{/tikz/circuitikz/tripoles/nand gate/circle width}\pgf@circ@res@right
  \pgf@circ@res@other=\pgfkeysvalueof{/tikz/circuitikz/tripoles/nand gate/port width}\pgf@circ@res@right

  \pgfpathmoveto{\pgfpoint{-\pgf@circ@res@other}{\pgf@circ@res@down}}
  \pgfpathcurveto{\pgfpoint{.0\pgf@circ@res@left}{\pgf@circ@res@down}}
                 {\pgfpoint{\pgf@circ@res@other-\pgf@circ@res@step}{.5\pgf@circ@res@down}}
                 {\pgfpoint{\pgf@circ@res@other-\pgf@circ@res@step}{0pt}}
  \pgfpathcurveto{\pgfpoint{\pgf@circ@res@other-\pgf@circ@res@step}{.5\pgf@circ@res@up}}
                 {\pgfpoint{.0\pgf@circ@res@left}{\pgf@circ@res@up}}
                 {\pgfpoint{\pgfkeysvalueof{/tikz/circuitikz/tripoles/nand gate/port width}\pgf@circ@res@left}{\pgf@circ@res@up}}
  \pgfclosepath

  \pgfsetcolor{\pgfkeysvalueof{/tikz/circuitikz/fill}}
  \pgfsetstrokecolor{black}
  \pgfusepath{draw,fill}

  \pgfpathellipse{\pgfpoint{\pgf@circ@res@other-.5\pgf@circ@res@step}{0pt}}
                 {\pgfpoint{.5\pgf@circ@res@step}{0pt}}
                 {\pgfpoint{0pt}{.5\pgf@circ@res@step}}

  \pgfsetstrokecolor{black}
  \pgfusepath{draw}
}

\pgfcircdeclarelogicgate{nor}{
  \pgfpathmoveto{\pgfpoint{\pgf@circ@res@left}{\pgfkeysvalueof{/tikz/circuitikz/tripoles/nor gate/input height}\pgf@circ@res@up}}
  \pgfpathlineto{\pgfpoint{(\pgfkeysvalueof{/tikz/circuitikz/tripoles/nor gate/port width}-\pgfkeysvalueof{/tikz/circuitikz/tripoles/nor gate/input skip})*\pgf@circ@res@left}{\pgfkeysvalueof{/tikz/circuitikz/tripoles/nor gate/input height}\pgf@circ@res@up}}

  \pgfpathmoveto{\pgfpoint{\pgf@circ@res@left}{\pgfkeysvalueof{/tikz/circuitikz/tripoles/nor gate/input height}\pgf@circ@res@down}}
  \pgfpathlineto{\pgfpoint{(\pgfkeysvalueof{/tikz/circuitikz/tripoles/nor gate/port width}-\pgfkeysvalueof{/tikz/circuitikz/tripoles/nor gate/input skip})*\pgf@circ@res@left}{\pgfkeysvalueof{/tikz/circuitikz/tripoles/nor gate/input height}\pgf@circ@res@down}}

  \pgfpathmoveto{\pgfpoint{\pgf@circ@res@right}{0pt}}
  \pgfpathlineto{\pgfpoint{\pgfkeysvalueof{/tikz/circuitikz/tripoles/nor gate/port width}\pgf@circ@res@right}{0pt}}

  \pgfusepath{draw}

  \pgf@circ@res@other=\pgfkeysvalueof{/tikz/circuitikz/tripoles/nor gate/port width}\pgf@circ@res@right
  \pgf@circ@res@step=\pgfkeysvalueof{/tikz/circuitikz/tripoles/nand gate/circle width}\pgf@circ@res@right

  \pgfsetlinewidth{2\pgflinewidth}
  \pgfpathmoveto{\pgfpoint{-\pgf@circ@res@other}{\pgf@circ@res@up}}
  \pgfpathcurveto{\pgfpoint{\pgfkeysvalueof{/tikz/circuitikz/tripoles/nor gate/aaa}\pgf@circ@res@left}{\pgf@circ@res@up}}
                 {\pgfpoint{\pgfkeysvalueof{/tikz/circuitikz/tripoles/nor gate/bbb}\pgf@circ@res@left}{\pgfkeysvalueof{/tikz/circuitikz/tripoles/nor gate/ccc}\pgf@circ@res@up}}
                 {\pgfpoint{\pgfkeysvalueof{/tikz/circuitikz/tripoles/nor gate/bbb}\pgf@circ@res@left}{0pt}}
  \pgfpathcurveto{\pgfpoint{\pgfkeysvalueof{/tikz/circuitikz/tripoles/nor gate/bbb}\pgf@circ@res@left}{\pgfkeysvalueof{/tikz/circuitikz/tripoles/nor gate/ccc}\pgf@circ@res@down}}
                 {\pgfpoint{\pgfkeysvalueof{/tikz/circuitikz/tripoles/nor gate/aaa}\pgf@circ@res@left}{\pgf@circ@res@down}}
                 {\pgfpoint{-\pgf@circ@res@other}{\pgf@circ@res@down}}

  \pgfpathcurveto{\pgfpoint{\pgfkeysvalueof{/tikz/circuitikz/tripoles/nor gate/ddd}\pgf@circ@res@left}{\pgf@circ@res@down}}
                 {\pgfpoint{\pgf@circ@res@other-\pgf@circ@res@step}{\pgfkeysvalueof{/tikz/circuitikz/tripoles/nor gate/ccc}\pgf@circ@res@down}}
                 {\pgfpoint{\pgf@circ@res@other-\pgf@circ@res@step}{0pt}}
  \pgfpathcurveto{\pgfpoint{\pgf@circ@res@other-\pgf@circ@res@step}{\pgfkeysvalueof{/tikz/circuitikz/tripoles/nor gate/ccc}\pgf@circ@res@up}}
                 {\pgfpoint{\pgfkeysvalueof{/tikz/circuitikz/tripoles/nor gate/ddd}\pgf@circ@res@left}{\pgf@circ@res@up}}
                 {\pgfpoint{-\pgf@circ@res@other}{\pgf@circ@res@up}}

  \pgfsetcolor{\pgfkeysvalueof{/tikz/circuitikz/fill}}
  \pgfsetstrokecolor{black}
  \pgfusepath{draw,fill}

  \pgfpathellipse{\pgfpoint{\pgf@circ@res@other-.5\pgf@circ@res@step}{0pt}}
                 {\pgfpoint{.5\pgf@circ@res@step}{0pt}}
                 {\pgfpoint{0pt}{.5\pgf@circ@res@step}}

  \pgfusepath{draw}
}

\pgfcircdeclarelogicgate{or}{
  \pgfpathmoveto{\pgfpoint{\pgf@circ@res@left}{\pgfkeysvalueof{/tikz/circuitikz/tripoles/or gate/input height}\pgf@circ@res@up}}
  \pgfpathlineto{\pgfpoint{(\pgfkeysvalueof{/tikz/circuitikz/tripoles/or gate/port width}-\pgfkeysvalueof{/tikz/circuitikz/tripoles/or gate/input skip})*\pgf@circ@res@left}{\pgfkeysvalueof{/tikz/circuitikz/tripoles/or gate/input height}\pgf@circ@res@up}}

  \pgfpathmoveto{\pgfpoint{\pgf@circ@res@left}{\pgfkeysvalueof{/tikz/circuitikz/tripoles/or gate/input height}\pgf@circ@res@down}}
  \pgfpathlineto{\pgfpoint{(\pgfkeysvalueof{/tikz/circuitikz/tripoles/or gate/port width}-\pgfkeysvalueof{/tikz/circuitikz/tripoles/or gate/input skip})*\pgf@circ@res@left}{\pgfkeysvalueof{/tikz/circuitikz/tripoles/or gate/input height}\pgf@circ@res@down}}

  \pgfpathmoveto{\pgfpoint{\pgf@circ@res@right}{0pt}}\pgfpathlineto{\pgfpoint{\pgfkeysvalueof{/tikz/circuitikz/tripoles/or gate/port width}\pgf@circ@res@right}{0pt}}

  \pgfusepath{draw}
  \pgf@circ@res@other=\pgfkeysvalueof{/tikz/circuitikz/tripoles/or gate/port width}\pgf@circ@res@right
  \pgfsetlinewidth{2\pgflinewidth}
  \pgfpathmoveto{\pgfpoint{-\pgf@circ@res@other}{\pgf@circ@res@up}}
  \pgfpathcurveto{\pgfpoint{\pgfkeysvalueof{/tikz/circuitikz/tripoles/or gate/aaa}\pgf@circ@res@left}{\pgf@circ@res@up}}
                 {\pgfpoint{\pgfkeysvalueof{/tikz/circuitikz/tripoles/or gate/bbb}\pgf@circ@res@left}{\pgfkeysvalueof{/tikz/circuitikz/tripoles/or gate/ccc}\pgf@circ@res@up}}
                 {\pgfpoint{\pgfkeysvalueof{/tikz/circuitikz/tripoles/or gate/bbb}\pgf@circ@res@left}{0pt}}
  \pgfpathcurveto{\pgfpoint{\pgfkeysvalueof{/tikz/circuitikz/tripoles/or gate/bbb}\pgf@circ@res@left}{\pgfkeysvalueof{/tikz/circuitikz/tripoles/or gate/ccc}\pgf@circ@res@down}}
                 {\pgfpoint{\pgfkeysvalueof{/tikz/circuitikz/tripoles/or gate/aaa}\pgf@circ@res@left}{\pgf@circ@res@down}}
                 {\pgfpoint{-\pgf@circ@res@other}{\pgf@circ@res@down}}

  \pgfpathcurveto{\pgfpoint{\pgfkeysvalueof{/tikz/circuitikz/tripoles/or gate/ddd}\pgf@circ@res@left}{\pgf@circ@res@down}}
                 {\pgfpoint{\pgf@circ@res@other}{\pgfkeysvalueof{/tikz/circuitikz/tripoles/or gate/ccc}\pgf@circ@res@down}}
                 {\pgfpoint{\pgf@circ@res@other}{0pt}}
  \pgfpathcurveto{\pgfpoint{\pgf@circ@res@other}{\pgfkeysvalueof{/tikz/circuitikz/tripoles/or gate/ccc}\pgf@circ@res@up}}
                 {\pgfpoint{\pgfkeysvalueof{/tikz/circuitikz/tripoles/or gate/ddd}\pgf@circ@res@left}{\pgf@circ@res@up}}
                 {\pgfpoint{-\pgf@circ@res@other}{\pgf@circ@res@up}}
  \pgfsetcolor{\pgfkeysvalueof{/tikz/circuitikz/fill}}
  \pgfsetstrokecolor{black}
  \pgfusepath{draw,fill}
}

\pgfcircdeclarelogicgate{xor}{
  \pgfpathmoveto{\pgfpoint{\pgf@circ@res@left}{\pgfkeysvalueof{/tikz/circuitikz/tripoles/xor gate/input height}\pgf@circ@res@up}}
  \pgfpathlineto{\pgfpoint{(\pgfkeysvalueof{/tikz/circuitikz/tripoles/xor gate/port width}-\pgfkeysvalueof{/tikz/circuitikz/tripoles/xor gate/input skip})*\pgf@circ@res@left}{\pgfkeysvalueof{/tikz/circuitikz/tripoles/xor gate/input height}\pgf@circ@res@up}}

  \pgfpathmoveto{\pgfpoint{\pgf@circ@res@left}{\pgfkeysvalueof{/tikz/circuitikz/tripoles/xor gate/input height}\pgf@circ@res@down}}
  \pgfpathlineto{\pgfpoint{(\pgfkeysvalueof{/tikz/circuitikz/tripoles/xor gate/port width}-\pgfkeysvalueof{/tikz/circuitikz/tripoles/xor gate/input skip})*\pgf@circ@res@left}{\pgfkeysvalueof{/tikz/circuitikz/tripoles/xor gate/input height}\pgf@circ@res@down}}

  \pgfpathmoveto{\pgfpoint{\pgf@circ@res@right}{0pt}}\pgfpathlineto{\pgfpoint{\pgfkeysvalueof{/tikz/circuitikz/tripoles/xor gate/port width}\pgf@circ@res@right}{0pt}}

  \pgfusepath{draw}
  \pgf@circ@res@other=\pgfkeysvalueof{/tikz/circuitikz/tripoles/xor gate/port width}\pgf@circ@res@right
  \pgfsetlinewidth{2\pgflinewidth}
  \pgfpathmoveto{\pgfpoint{-\pgf@circ@res@other}{\pgf@circ@res@up}}
  \pgfpathcurveto{\pgfpoint{\pgfkeysvalueof{/tikz/circuitikz/tripoles/xor gate/aaa}\pgf@circ@res@left}{\pgf@circ@res@up}}
                 {\pgfpoint{\pgfkeysvalueof{/tikz/circuitikz/tripoles/xor gate/bbb}\pgf@circ@res@left}{\pgfkeysvalueof{/tikz/circuitikz/tripoles/xor gate/ccc}\pgf@circ@res@up}}
                 {\pgfpoint{\pgfkeysvalueof{/tikz/circuitikz/tripoles/xor gate/bbb}\pgf@circ@res@left}{0pt}}
  \pgfpathcurveto{\pgfpoint{\pgfkeysvalueof{/tikz/circuitikz/tripoles/xor gate/bbb}\pgf@circ@res@left}{\pgfkeysvalueof{/tikz/circuitikz/tripoles/xor gate/ccc}\pgf@circ@res@down}}
                 {\pgfpoint{\pgfkeysvalueof{/tikz/circuitikz/tripoles/xor gate/aaa}\pgf@circ@res@left}{\pgf@circ@res@down}}
                 {\pgfpoint{-\pgf@circ@res@other}{\pgf@circ@res@down}}

  \pgfpathcurveto{\pgfpoint{\pgfkeysvalueof{/tikz/circuitikz/tripoles/xor gate/ddd}\pgf@circ@res@left}{\pgf@circ@res@down}}
                 {\pgfpoint{\pgf@circ@res@other}{\pgfkeysvalueof{/tikz/circuitikz/tripoles/xor gate/ccc}\pgf@circ@res@down}}
                 {\pgfpoint{\pgf@circ@res@other}{0pt}}
  \pgfpathcurveto{\pgfpoint{\pgf@circ@res@other}{\pgfkeysvalueof{/tikz/circuitikz/tripoles/xor gate/ccc}\pgf@circ@res@up}}
                 {\pgfpoint{\pgfkeysvalueof{/tikz/circuitikz/tripoles/xor gate/ddd}\pgf@circ@res@left}{\pgf@circ@res@up}}
                 {\pgfpoint{-\pgf@circ@res@other}{\pgf@circ@res@up}}

  \pgfsetcolor{\pgfkeysvalueof{/tikz/circuitikz/fill}}
  \pgfsetstrokecolor{black}
  \pgfusepath{draw,fill}

  \def\pgf@circ@temp{-\pgfkeysvalueof{/tikz/circuitikz/tripoles/xor gate/distance}\pgf@circ@res@right+}
  \pgfpathmoveto{\pgfpoint{\pgf@circ@temp-\pgf@circ@res@other}{\pgf@circ@res@up}}

  \pgfpathcurveto{\pgfpoint{\pgf@circ@temp\pgfkeysvalueof{/tikz/circuitikz/tripoles/xor gate/aaa}\pgf@circ@res@left}{.95*\pgf@circ@res@up}}
                 {\pgfpoint{\pgf@circ@temp\pgfkeysvalueof{/tikz/circuitikz/tripoles/xor gate/bbb}\pgf@circ@res@left}{\pgfkeysvalueof{/tikz/circuitikz/tripoles/xor gate/ccc}\pgf@circ@res@up}}
                 {\pgfpoint{\pgf@circ@temp\pgfkeysvalueof{/tikz/circuitikz/tripoles/xor gate/bbb}\pgf@circ@res@left}{0pt}}
  \pgfpathcurveto{\pgfpoint{\pgf@circ@temp\pgfkeysvalueof{/tikz/circuitikz/tripoles/xor gate/bbb}\pgf@circ@res@left}{\pgfkeysvalueof{/tikz/circuitikz/tripoles/xor gate/ccc}\pgf@circ@res@down}}
                 {\pgfpoint{\pgf@circ@temp\pgfkeysvalueof{/tikz/circuitikz/tripoles/xor gate/aaa}\pgf@circ@res@left}{.95*\pgf@circ@res@down}}
                 {\pgfpoint{\pgf@circ@temp-\pgf@circ@res@other}{.95*\pgf@circ@res@down}}

  \pgfusepath{draw}
}

\pgfcircdeclarelogicgate{xnor}{
  \pgfpathmoveto{\pgfpoint{\pgf@circ@res@left}{\pgfkeysvalueof{/tikz/circuitikz/tripoles/xnor gate/input height}\pgf@circ@res@up}}
  \pgfpathlineto{\pgfpoint{(\pgfkeysvalueof{/tikz/circuitikz/tripoles/xnor gate/port width}-\pgfkeysvalueof{/tikz/circuitikz/tripoles/xnor gate/input skip})*\pgf@circ@res@left}{\pgfkeysvalueof{/tikz/circuitikz/tripoles/xnor gate/input height}\pgf@circ@res@up}}

  \pgfpathmoveto{\pgfpoint{\pgf@circ@res@left}{\pgfkeysvalueof{/tikz/circuitikz/tripoles/xnor gate/input height}\pgf@circ@res@down}}
  \pgfpathlineto{\pgfpoint{(\pgfkeysvalueof{/tikz/circuitikz/tripoles/xnor gate/port width}-\pgfkeysvalueof{/tikz/circuitikz/tripoles/xnor gate/input skip})*\pgf@circ@res@left}{\pgfkeysvalueof{/tikz/circuitikz/tripoles/xnor gate/input height}\pgf@circ@res@down}}

  \pgfpathmoveto{\pgfpoint{\pgf@circ@res@right}{0pt}}\pgfpathlineto{\pgfpoint{\pgfkeysvalueof{/tikz/circuitikz/tripoles/xnor gate/port width}\pgf@circ@res@right}{0pt}}

  \pgfusepath{draw}
  \pgf@circ@res@other=\pgfkeysvalueof{/tikz/circuitikz/tripoles/xnor gate/port width}\pgf@circ@res@right
  \pgf@circ@res@step=\pgfkeysvalueof{/tikz/circuitikz/tripoles/xnor gate/circle width}\pgf@circ@res@right

  \pgfsetlinewidth{2\pgflinewidth}
  \pgfpathmoveto{\pgfpoint{-\pgf@circ@res@other}{\pgf@circ@res@up}}
  \pgfpathcurveto{\pgfpoint{\pgfkeysvalueof{/tikz/circuitikz/tripoles/xnor gate/aaa}\pgf@circ@res@left}{\pgf@circ@res@up}}
                 {\pgfpoint{\pgfkeysvalueof{/tikz/circuitikz/tripoles/xnor gate/bbb}\pgf@circ@res@left}{\pgfkeysvalueof{/tikz/circuitikz/tripoles/xnor gate/ccc}\pgf@circ@res@up}}
                 {\pgfpoint{\pgfkeysvalueof{/tikz/circuitikz/tripoles/xnor gate/bbb}\pgf@circ@res@left}{0pt}}
  \pgfpathcurveto{\pgfpoint{\pgfkeysvalueof{/tikz/circuitikz/tripoles/xnor gate/bbb}\pgf@circ@res@left}{\pgfkeysvalueof{/tikz/circuitikz/tripoles/xnor gate/ccc}\pgf@circ@res@down}}
                 {\pgfpoint{\pgfkeysvalueof{/tikz/circuitikz/tripoles/xnor gate/aaa}\pgf@circ@res@left}{\pgf@circ@res@down}}
                 {\pgfpoint{-\pgf@circ@res@other}{\pgf@circ@res@down}}

  \pgfpathcurveto{\pgfpoint{\pgfkeysvalueof{/tikz/circuitikz/tripoles/xnor gate/ddd}\pgf@circ@res@left}{\pgf@circ@res@down}}
                 {\pgfpoint{\pgf@circ@res@other-\pgf@circ@res@step}{\pgfkeysvalueof{/tikz/circuitikz/tripoles/xnor gate/ccc}\pgf@circ@res@down}}
                 {\pgfpoint{\pgf@circ@res@other-\pgf@circ@res@step}{0pt}}
  \pgfpathcurveto{\pgfpoint{\pgf@circ@res@other-\pgf@circ@res@step}{\pgfkeysvalueof{/tikz/circuitikz/tripoles/xnor gate/ccc}\pgf@circ@res@up}}
                 {\pgfpoint{\pgfkeysvalueof{/tikz/circuitikz/tripoles/xnor gate/ddd}\pgf@circ@res@left}{\pgf@circ@res@up}}
                 {\pgfpoint{-\pgf@circ@res@other}{\pgf@circ@res@up}}

  \pgfsetcolor{\pgfkeysvalueof{/tikz/circuitikz/fill}}
  \pgfsetstrokecolor{black}
  \pgfusepath{draw,fill}

  \def\pgf@circ@temp{-\pgfkeysvalueof{/tikz/circuitikz/tripoles/xnor gate/distance}\pgf@circ@res@right+}
  \pgfpathmoveto{\pgfpoint{\pgf@circ@temp-\pgf@circ@res@other}{\pgf@circ@res@up}}

  \pgfpathcurveto{\pgfpoint{\pgf@circ@temp\pgfkeysvalueof{/tikz/circuitikz/tripoles/xnor gate/aaa}\pgf@circ@res@left}{.95*\pgf@circ@res@up}}
                 {\pgfpoint{\pgf@circ@temp\pgfkeysvalueof{/tikz/circuitikz/tripoles/xnor gate/bbb}\pgf@circ@res@left}{\pgfkeysvalueof{/tikz/circuitikz/tripoles/xnor gate/ccc}\pgf@circ@res@up}}
                 {\pgfpoint{\pgf@circ@temp\pgfkeysvalueof{/tikz/circuitikz/tripoles/xnor gate/bbb}\pgf@circ@res@left}{0pt}}
  \pgfpathcurveto{\pgfpoint{\pgf@circ@temp\pgfkeysvalueof{/tikz/circuitikz/tripoles/xnor gate/bbb}\pgf@circ@res@left}{\pgfkeysvalueof{/tikz/circuitikz/tripoles/xnor gate/ccc}\pgf@circ@res@down}}
                 {\pgfpoint{\pgf@circ@temp\pgfkeysvalueof{/tikz/circuitikz/tripoles/xnor gate/aaa}\pgf@circ@res@left}{.95*\pgf@circ@res@down}}
                 {\pgfpoint{\pgf@circ@temp-\pgf@circ@res@other}{.95*\pgf@circ@res@down}}

  \pgfusepath{draw}

  \pgfpathellipse{\pgfpoint{\pgf@circ@res@other-.5\pgf@circ@res@step}{0pt}}
                 {\pgfpoint{.5\pgf@circ@res@step}{0pt}}
                 {\pgfpoint{0pt}{.5\pgf@circ@res@step}}

  \pgfusepath{draw}
}

\pgfdeclareshape{not gate}{
  \anchor{center}{\pgfpointorigin}
  \savedanchor\northwest{
    \pgf@y=\pgfkeysvalueof{/tikz/circuitikz/bipoles/not gate/height}\pgf@circ@Rlen
    \pgf@y=.5\pgf@y
    \pgf@x=-\pgfkeysvalueof{/tikz/circuitikz/bipoles/not gate/width}\pgf@circ@Rlen
    \pgf@x=.5\pgf@x
  }
  \anchor{in}{
    \northwest
    \pgf@y=0pt
  }
  \anchor{out}{
    \northwest
    \pgf@y=0pt
    \pgf@x=-\pgf@x
  }
  \backgroundpath{
    \pgfsetcolor{\pgfkeysvalueof{/tikz/circuitikz/color}}

    \northwest
    \pgf@circ@res@up=\pgf@y
    \pgf@circ@res@down=-\pgf@y
    \pgf@circ@res@right=-\pgf@x
    \pgf@circ@res@left=\pgf@x

    \pgf@circ@res@other=\pgfkeysvalueof{/tikz/circuitikz/bipoles/not gate/circle width}\pgf@circ@res@right

    \pgfscope
    \pgfsetlinewidth{2\pgflinewidth}
    \pgftransformxshift{.7\pgf@circ@res@left}
    \pgf@circ@res@step=\pgf@circ@res@right
    \advance\pgf@circ@res@step by -\pgf@circ@res@left
    \pgf@circ@res@step=.7\pgf@circ@res@step

    \pgfpathmoveto{\pgfpoint{\pgf@circ@res@step-\pgf@circ@res@other}{0pt}}
    \pgfpathlineto{\pgfpoint{0pt}{\pgf@circ@res@up}}
    \pgfpathlineto{\pgfpoint{0pt}{\pgf@circ@res@down}}
    \pgfclosepath
    \pgfsetcolor{\pgfkeysvalueof{/tikz/circuitikz/fill}}
    \pgfsetstrokecolor{black}
    \pgfusepath{draw,fill}

    \pgfpathellipse{\pgfpoint{\pgf@circ@res@step-.5\pgf@circ@res@other}{0pt}}
                   {\pgfpoint{.5\pgf@circ@res@other}{0pt}}
                   {\pgfpoint{0pt}{.5\pgf@circ@res@other}}
    \pgfsetcolor{white}
    \pgfsetstrokecolor{black}
    \pgfusepath{draw,fill}
    \endpgfscope

    \pgfpathmoveto{\pgfpoint{\pgf@circ@res@left}{0pt}}
    \pgfpathlineto{\pgfpoint{.7\pgf@circ@res@left}{0pt}}

    \pgfpathmoveto{\pgfpoint{\pgf@circ@res@right}{0pt}}
    \pgfpathlineto{\pgfpoint{.7\pgf@circ@res@right}{0pt}}

    \pgfusepath{draw}
  }
}

\pgfdeclareshape{buffer gate}{
  \anchor{center}{\pgfpointorigin}
  \savedanchor\northwest{\pgf@y=\pgfkeysvalueof{/tikz/circuitikz/bipoles/not gate/height}\pgf@circ@Rlen
    \pgf@y=.5\pgf@y
    \pgf@x=-\pgfkeysvalueof{/tikz/circuitikz/bipoles/not gate/width}\pgf@circ@Rlen
    \pgf@x=.5\pgf@x
  }
  \savedanchor\left{\pgf@y=0pt
  }
  \anchor{in}{
    \northwest
    \pgf@y=0pt
  }
  \anchor{out}{
    \northwest
    \pgf@y=0pt
    \pgf@x=-\pgf@x
  }
  \backgroundpath{
    \pgfsetcolor{\pgfkeysvalueof{/tikz/circuitikz/color}}

    \northwest
    \pgf@circ@res@up=\pgf@y
    \pgf@circ@res@down=-\pgf@y
    \pgf@circ@res@right=-\pgf@x
    \pgf@circ@res@left=\pgf@x

    \pgf@circ@res@other=\pgfkeysvalueof{/tikz/circuitikz/bipoles/not gate/circle width}\pgf@circ@res@right

    \pgfscope
    \pgfsetlinewidth{2\pgflinewidth}
    \pgftransformxshift{.7\pgf@circ@res@left}
    \pgf@circ@res@step=\pgf@circ@res@right
    \advance\pgf@circ@res@step by -\pgf@circ@res@left
    \pgf@circ@res@step=.7\pgf@circ@res@step

    \pgfpathmoveto{\pgfpoint{\pgf@circ@res@step-\pgf@circ@res@other}{0pt}}
    \pgfpathlineto{\pgfpoint{0pt}{\pgf@circ@res@up}}
    \pgfpathlineto{\pgfpoint{0pt}{\pgf@circ@res@down}}
    \pgfpathlineto{\pgfpoint{\pgf@circ@res@step-\pgf@circ@res@other}{0pt}}
    \pgfsetcolor{\pgfkeysvalueof{/tikz/circuitikz/fill}}
    \pgfsetstrokecolor{black}
    \pgfusepath{draw,fill}
    \endpgfscope

    \pgfpathmoveto{\pgfpoint{\pgf@circ@res@left}{0pt}}
    \pgfpathlineto{\pgfpoint{.7\pgf@circ@res@left}{0pt}}

    \pgfpathmoveto{\pgfpoint{\pgf@circ@res@right}{0pt}}
    \pgfpathlineto{\pgfpoint{.5\pgf@circ@res@right}{0pt}}

    \pgfusepath{draw}

  }
}

\pgfdeclareshape{tristate buffer gate}{
  \anchor{center}{\pgfpointorigin}
  \savedanchor\northwest{
    \pgf@y=\pgfkeysvalueof{/tikz/circuitikz/bipoles/not gate/height}\pgf@circ@Rlen
    \pgf@y=.5\pgf@y
    \pgf@x=-\pgfkeysvalueof{/tikz/circuitikz/bipoles/not gate/width}\pgf@circ@Rlen
    \pgf@x=.5\pgf@x
  }
  \savedanchor\left{
    \pgf@y=0pt
  }
  \anchor{in}{
    \northwest
    \pgf@y=0pt
  }
  \anchor{en}{
    \northwest
    \pgf@x=0pt
  }
  \anchor{out}{
    \northwest
    \pgf@y=0pt
    \pgf@x=-\pgf@x
  }
  \anchor{center}{
    \pgfpointorigin
  }
  \backgroundpath{
    \pgfsetcolor{\pgfkeysvalueof{/tikz/circuitikz/color}}

    \northwest
    \pgf@circ@res@up=\pgf@y
    \pgf@circ@res@down=-\pgf@y
    \pgf@circ@res@right=-\pgf@x
    \pgf@circ@res@left=\pgf@x

    \pgf@circ@res@other=\pgfkeysvalueof{/tikz/circuitikz/bipoles/not gate/circle width}\pgf@circ@res@right

    \pgfscope
    \pgfsetlinewidth{2\pgflinewidth}
    \pgftransformxshift{.7\pgf@circ@res@left}
    \pgf@circ@res@step=\pgf@circ@res@right
    \advance\pgf@circ@res@step by -\pgf@circ@res@left
    \pgf@circ@res@step=.7\pgf@circ@res@step

    \pgfpathmoveto{\pgfpoint{\pgf@circ@res@step-\pgf@circ@res@other}{0pt}}
    \pgfpathlineto{\pgfpoint{0pt}{\pgf@circ@res@up}}
    \pgfpathlineto{\pgfpoint{0pt}{\pgf@circ@res@down}}
    \pgfpathlineto{\pgfpoint{\pgf@circ@res@step-\pgf@circ@res@other}{0pt}}
    \pgfsetcolor{\pgfkeysvalueof{/tikz/circuitikz/fill}}
    \pgfsetstrokecolor{black}
    \pgfusepath{draw,fill}
    \endpgfscope

    \pgfpathmoveto{\pgfpoint{\pgf@circ@res@left}{0pt}}
    \pgfpathlineto{\pgfpoint{.7\pgf@circ@res@left}{0pt}}

    \pgfpathmoveto{\pgfpoint{\pgf@circ@res@right}{0pt}}
    \pgfpathlineto{\pgfpoint{.5\pgf@circ@res@right}{0pt}}

    \pgfpathmoveto{\pgfpoint{0pt}{.45\pgf@circ@res@up}}
    \pgfpathlineto{\pgfpoint{0pt}{\pgf@circ@res@up}}

    \pgfusepath{draw}
  }
}

\makeatother
 \makeatletter

\ctikzset{variable and gate/width/.initial=1.1}
\ctikzset{variable and gate/height/.initial=1.2}
\ctikzset{variable and gate/port width/.initial=.7}
\ctikzset{variable and gate/input height/.initial=.6}
\ctikzset{variable and gate/input height one/.initial=0.6}
\ctikzset{variable and gate/input height two/.initial=0.2}
\ctikzset{variable and gate/input height three/.initial=0.6}

\ctikzset{variable and gate alt/width/.initial=1.1}
\ctikzset{variable and gate alt/height/.initial=1.4}
\ctikzset{variable and gate alt/port width/.initial=.7}
\ctikzset{variable and gate alt/input height/.initial=.6}
\ctikzset{variable and gate alt/input height one/.initial=0.7}
\ctikzset{variable and gate alt/input height two/.initial=0.4}
\ctikzset{variable and gate alt/input height three/.initial=0.4}
\ctikzset{variable and gate alt/input height four/.initial=0.7}

\ctikzset{variable or gate/width/.initial=1.1}
\ctikzset{variable or gate/height/.initial=1.2}
\ctikzset{variable or gate/port width/.initial=.7}
\ctikzset{variable or gate/input height/.initial=.6}
\ctikzset{variable or gate/input height one/.initial=0.6}
\ctikzset{variable or gate/input height two/.initial=0.2}
\ctikzset{variable or gate/input height three/.initial=0.6}
\ctikzset{variable or gate/input skip/.initial=.225}
\ctikzset{variable or gate/input skip two/.initial=.275}
\ctikzset{variable or gate/aaa/.initial=.6}
\ctikzset{variable or gate/bbb/.initial=.4} 
\ctikzset{variable or gate/ccc/.initial=.5} 
\ctikzset{variable or gate/ddd/.initial=.0}

\ctikzset{three and gate/width/.initial=1.1}
\ctikzset{three and gate/height/.initial=1}
\ctikzset{three and gate/port width/.initial=.7}
\ctikzset{three and gate/input height/.initial=.6}
\ctikzset{three and gate/input height one/.initial=0.6}
\ctikzset{three and gate/input height two/.initial=0}
\ctikzset{three and gate/input height three/.initial=0.6}

\long\def\pgfcircdeclarevarlogicport#1#2{
  \pgfdeclareshape{#1}{
    \anchor{center}{\pgfpointorigin}
    \savedanchor\northwest{
      \pgf@y=\pgfkeysvalueof{/tikz/circuitikz/#1/height}\pgf@circ@Rlen
      \pgf@y=.5\pgf@y
      \pgf@x=-\pgfkeysvalueof{/tikz/circuitikz/#1/width}\pgf@circ@Rlen
      \pgf@x=.5\pgf@x
    }
    \anchor{in 1}{
      \pgf@x=-\pgfkeysvalueof{/tikz/circuitikz/#1/width}\pgf@circ@Rlen
      \pgf@x=.5\pgf@x
      \pgf@y=\pgfkeysvalueof{/tikz/circuitikz/#1/height}\pgf@circ@Rlen
      \pgf@y=.5\pgf@y
      \pgf@y=\pgfkeysvalueof{/tikz/circuitikz/#1/input height one}\pgf@y
    }
    \anchor{in 2}{
      \pgf@x=-\pgfkeysvalueof{/tikz/circuitikz/#1/width}\pgf@circ@Rlen
      \pgf@x=.5\pgf@x
      \pgf@y=\pgfkeysvalueof{/tikz/circuitikz/#1/height}\pgf@circ@Rlen
      \pgf@y=.5\pgf@y
      \pgf@y=\pgfkeysvalueof{/tikz/circuitikz/#1/input height two}\pgf@y
    }
    \anchor{in 3}{
      \pgf@x=-\pgfkeysvalueof{/tikz/circuitikz/#1/width}\pgf@circ@Rlen
      \pgf@x=.5\pgf@x
      \pgf@y=-\pgfkeysvalueof{/tikz/circuitikz/#1/height}\pgf@circ@Rlen
      \pgf@y=.5\pgf@y
      \pgf@y=\pgfkeysvalueof{/tikz/circuitikz/#1/input height three}\pgf@y
    }
    \anchor{out}{
      \northwest
      \pgf@y=0pt
      \pgf@x=-\pgf@x
    }
    \backgroundpath{          
      \pgfsetcolor{\pgfkeysvalueof{/tikz/circuitikz/color}}   
      \northwest
      \pgf@circ@res@up=\pgf@y 
      \pgf@circ@res@down=-\pgf@y
      \pgf@circ@res@right=-\pgf@x
      \pgf@circ@res@left=\pgf@x
      #2
    }
  }
}

\long\def\pgfcircdeclarevarlogicportalt#1#2{
  \pgfdeclareshape{#1}{
    \anchor{center}{\pgfpointorigin}
    \savedanchor\northwest{
      \pgf@y=\pgfkeysvalueof{/tikz/circuitikz/#1/height}\pgf@circ@Rlen
      \pgf@y=.5\pgf@y
      \pgf@x=-\pgfkeysvalueof{/tikz/circuitikz/#1/width}\pgf@circ@Rlen
      \pgf@x=.5\pgf@x
    }
    \anchor{in 1}{
      \pgf@x=-\pgfkeysvalueof{/tikz/circuitikz/#1/width}\pgf@circ@Rlen
      \pgf@x=.5\pgf@x
      \pgf@y=\pgfkeysvalueof{/tikz/circuitikz/#1/height}\pgf@circ@Rlen
      \pgf@y=.5\pgf@y
      \pgf@y=\pgfkeysvalueof{/tikz/circuitikz/#1/input height one}\pgf@y
    }
    \anchor{in 2}{
      \pgf@x=-\pgfkeysvalueof{/tikz/circuitikz/#1/width}\pgf@circ@Rlen
      \pgf@x=.5\pgf@x
      \pgf@y=\pgfkeysvalueof{/tikz/circuitikz/#1/height}\pgf@circ@Rlen
      \pgf@y=.5\pgf@y
      \pgf@y=\pgfkeysvalueof{/tikz/circuitikz/#1/input height two}\pgf@y
    }
    \anchor{in 3}{
      \pgf@x=-\pgfkeysvalueof{/tikz/circuitikz/#1/width}\pgf@circ@Rlen
      \pgf@x=.5\pgf@x
      \pgf@y=-\pgfkeysvalueof{/tikz/circuitikz/#1/height}\pgf@circ@Rlen
      \pgf@y=.5\pgf@y
      \pgf@y=\pgfkeysvalueof{/tikz/circuitikz/#1/input height three}\pgf@y
    }
    \anchor{in 4}{
      \pgf@x=-\pgfkeysvalueof{/tikz/circuitikz/#1/width}\pgf@circ@Rlen
      \pgf@x=.5\pgf@x
      \pgf@y=-\pgfkeysvalueof{/tikz/circuitikz/#1/height}\pgf@circ@Rlen
      \pgf@y=.5\pgf@y
      \pgf@y=\pgfkeysvalueof{/tikz/circuitikz/#1/input height four}\pgf@y
    }
    \anchor{out}{
      \northwest
      \pgf@y=0pt
      \pgf@x=-\pgf@x
    }
    \backgroundpath{          
      \pgfsetcolor{\pgfkeysvalueof{/tikz/circuitikz/color}}   
      \northwest
      \pgf@circ@res@up=\pgf@y 
      \pgf@circ@res@down=-\pgf@y
      \pgf@circ@res@right=-\pgf@x
      \pgf@circ@res@left=\pgf@x
      #2
    }
  }
}

\pgfcircdeclarevarlogicport{variable and gate}{
  \pgfpathmoveto{\pgfpoint{\pgf@circ@res@left}{\pgfkeysvalueof{/tikz/circuitikz/variable and gate/input height one}\pgf@circ@res@up}}
  \pgfpathlineto{\pgfpoint{\pgfkeysvalueof{/tikz/circuitikz/variable and gate/port width}\pgf@circ@res@left}{\pgfkeysvalueof{/tikz/circuitikz/variable and gate/input height one}\pgf@circ@res@up}}

  \pgfpathmoveto{\pgfpoint{\pgf@circ@res@left}{\pgfkeysvalueof{/tikz/circuitikz/variable and gate/input height two}\pgf@circ@res@up}}
  \pgfpathlineto{\pgfpoint{\pgfkeysvalueof{/tikz/circuitikz/variable and gate/port width}\pgf@circ@res@left}{\pgfkeysvalueof{/tikz/circuitikz/variable and gate/input height two}\pgf@circ@res@up}}

  \pgfpathmoveto{\pgfpoint{\pgf@circ@res@left}{\pgfkeysvalueof{/tikz/circuitikz/variable and gate/input height three}\pgf@circ@res@down}}
  \pgfpathlineto{\pgfpoint{\pgfkeysvalueof{/tikz/circuitikz/variable and gate/port width}\pgf@circ@res@left}{\pgfkeysvalueof{/tikz/circuitikz/variable and gate/input height three}\pgf@circ@res@down}}

  \pgfpathmoveto{\pgfpoint{\pgf@circ@res@right}{0pt}}
  \pgfpathlineto{\pgfpoint{\pgfkeysvalueof{/tikz/circuitikz/variable and gate/port width}\pgf@circ@res@right}{0pt}}

  \pgfusepath{draw}

  \pgfsetlinewidth{2\pgflinewidth}
  \pgfpathmoveto{\pgfpoint{\pgfkeysvalueof{/tikz/circuitikz/variable and gate/port width}\pgf@circ@res@left}{\pgf@circ@res@down}}
  \pgfpathcurveto{\pgfpoint{.0\pgf@circ@res@left}{\pgf@circ@res@down}}{\pgfpoint{\pgfkeysvalueof{/tikz/circuitikz/variable and gate/port width}\pgf@circ@res@right}{.5\pgf@circ@res@down}}{\pgfpoint{\pgfkeysvalueof{/tikz/circuitikz/variable and gate/port width}\pgf@circ@res@right}{0pt}}
  \pgfpathcurveto{\pgfpoint{\pgfkeysvalueof{/tikz/circuitikz/variable and gate/port width}\pgf@circ@res@right}{.5\pgf@circ@res@up}}{\pgfpoint{.0\pgf@circ@res@left}{\pgf@circ@res@up}}{\pgfpoint{\pgfkeysvalueof{/tikz/circuitikz/variable and gate/port width}\pgf@circ@res@left}{\pgf@circ@res@up}}
  \pgfclosepath
  \pgfsetcolor{\pgfkeysvalueof{/tikz/circuitikz/fill}}
  \pgfsetstrokecolor{black}
  \pgfusepath{draw,fill}
}

\pgfcircdeclarevarlogicport{variable or gate}{
  \pgfpathmoveto{\pgfpoint{\pgf@circ@res@left}{\pgfkeysvalueof{/tikz/circuitikz/variable or gate/input height one}\pgf@circ@res@up}}
  \pgfpathlineto{\pgfpoint{(\pgfkeysvalueof{/tikz/circuitikz/variable or gate/port width}-\pgfkeysvalueof{/tikz/circuitikz/variable or gate/input skip})*\pgf@circ@res@left}
                {\pgfkeysvalueof{/tikz/circuitikz/variable or gate/input height one}\pgf@circ@res@up}}

  \pgfpathmoveto{\pgfpoint{\pgf@circ@res@left}{\pgfkeysvalueof{/tikz/circuitikz/variable or gate/input height two}\pgf@circ@res@up}}
  \pgfpathlineto{\pgfpoint{(\pgfkeysvalueof{/tikz/circuitikz/variable or gate/port width}-\pgfkeysvalueof{/tikz/circuitikz/variable or gate/input skip two})*\pgf@circ@res@left}
                          {\pgfkeysvalueof{/tikz/circuitikz/variable or gate/input height two}\pgf@circ@res@up}}

  \pgfpathmoveto{\pgfpoint{\pgf@circ@res@left}{\pgfkeysvalueof{/tikz/circuitikz/variable or gate/input height three}\pgf@circ@res@down}}
  \pgfpathlineto{\pgfpoint{(\pgfkeysvalueof{/tikz/circuitikz/variable or gate/port width}-\pgfkeysvalueof{/tikz/circuitikz/variable or gate/input skip})*\pgf@circ@res@left}
                          {\pgfkeysvalueof{/tikz/circuitikz/variable or gate/input height three}\pgf@circ@res@down}}

  \pgfpathmoveto{\pgfpoint{\pgf@circ@res@right}{0pt}}\pgfpathlineto{\pgfpoint{\pgfkeysvalueof{/tikz/circuitikz/variable or gate/port width}\pgf@circ@res@right}{0pt}}  

  \pgfusepath{draw}

  \pgfsetlinewidth{2\pgflinewidth}

  \pgf@circ@res@other=\pgfkeysvalueof{/tikz/circuitikz/variable or gate/port width}\pgf@circ@res@right
  \pgfpathmoveto{\pgfpoint{-\pgf@circ@res@other}{\pgf@circ@res@up}}
  \pgfpathcurveto{\pgfpoint{\pgfkeysvalueof{/tikz/circuitikz/variable or gate/aaa}\pgf@circ@res@left}{\pgf@circ@res@up}}{\pgfpoint{\pgfkeysvalueof{/tikz/circuitikz/variable or gate/bbb}\pgf@circ@res@left}{\pgfkeysvalueof{/tikz/circuitikz/variable or gate/ccc}\pgf@circ@res@up}}{\pgfpoint{\pgfkeysvalueof{/tikz/circuitikz/variable or gate/bbb}\pgf@circ@res@left}{0pt}}
  \pgfpathcurveto{\pgfpoint{\pgfkeysvalueof{/tikz/circuitikz/variable or gate/bbb}\pgf@circ@res@left}{\pgfkeysvalueof{/tikz/circuitikz/variable or gate/ccc}\pgf@circ@res@down}}{\pgfpoint{\pgfkeysvalueof{/tikz/circuitikz/variable or gate/aaa}\pgf@circ@res@left}{\pgf@circ@res@down}}{\pgfpoint{-\pgf@circ@res@other}{\pgf@circ@res@down}}
  \pgfpathcurveto{\pgfpoint{\pgfkeysvalueof{/tikz/circuitikz/variable or gate/ddd}\pgf@circ@res@left}{\pgf@circ@res@down}}{\pgfpoint{\pgf@circ@res@other}{\pgfkeysvalueof{/tikz/circuitikz/variable or gate/ccc}\pgf@circ@res@down}}{\pgfpoint{\pgf@circ@res@other}{0pt}}
  \pgfpathcurveto{\pgfpoint{\pgf@circ@res@other}{\pgfkeysvalueof{/tikz/circuitikz/variable or gate/ccc}\pgf@circ@res@up}}{\pgfpoint{\pgfkeysvalueof{/tikz/circuitikz/variable or gate/ddd}\pgf@circ@res@left}{\pgf@circ@res@up}}{\pgfpoint{-\pgf@circ@res@other}{\pgf@circ@res@up}}
  \pgfclosepath

  \pgfsetcolor{\pgfkeysvalueof{/tikz/circuitikz/fill}}
  \pgfsetstrokecolor{black}
  \pgfusepath{draw,fill}
}

\pgfcircdeclarevarlogicport{three and gate}{
  \pgfpathmoveto{\pgfpoint{\pgf@circ@res@left}{\pgfkeysvalueof{/tikz/circuitikz/three and gate/input height one}\pgf@circ@res@up}}
  \pgfpathlineto{\pgfpoint{\pgfkeysvalueof{/tikz/circuitikz/three and gate/port width}\pgf@circ@res@left}{\pgfkeysvalueof{/tikz/circuitikz/three and gate/input height one}\pgf@circ@res@up}}

  \pgfpathmoveto{\pgfpoint{\pgf@circ@res@left}{\pgfkeysvalueof{/tikz/circuitikz/three and gate/input height two}\pgf@circ@res@up}}
  \pgfpathlineto{\pgfpoint{\pgfkeysvalueof{/tikz/circuitikz/three and gate/port width}\pgf@circ@res@left}{\pgfkeysvalueof{/tikz/circuitikz/three and gate/input height two}\pgf@circ@res@up}}

  \pgfpathmoveto{\pgfpoint{\pgf@circ@res@left}{\pgfkeysvalueof{/tikz/circuitikz/three and gate/input height three}\pgf@circ@res@down}}
  \pgfpathlineto{\pgfpoint{\pgfkeysvalueof{/tikz/circuitikz/three and gate/port width}\pgf@circ@res@left}{\pgfkeysvalueof{/tikz/circuitikz/three and gate/input height three}\pgf@circ@res@down}}

  \pgfpathmoveto{\pgfpoint{\pgf@circ@res@right}{0pt}}
  \pgfpathlineto{\pgfpoint{\pgfkeysvalueof{/tikz/circuitikz/three and gate/port width}\pgf@circ@res@right}{0pt}}

  \pgfusepath{draw}

  \pgfsetlinewidth{2\pgflinewidth}
  \pgfpathmoveto{\pgfpoint{\pgfkeysvalueof{/tikz/circuitikz/three and gate/port width}\pgf@circ@res@left}{\pgf@circ@res@down}}
  \pgfpathcurveto{\pgfpoint{.0\pgf@circ@res@left}{\pgf@circ@res@down}}{\pgfpoint{\pgfkeysvalueof{/tikz/circuitikz/three and gate/port width}\pgf@circ@res@right}{.5\pgf@circ@res@down}}{\pgfpoint{\pgfkeysvalueof{/tikz/circuitikz/three and gate/port width}\pgf@circ@res@right}{0pt}}
  \pgfpathcurveto{\pgfpoint{\pgfkeysvalueof{/tikz/circuitikz/three and gate/port width}\pgf@circ@res@right}{.5\pgf@circ@res@up}}{\pgfpoint{.0\pgf@circ@res@left}{\pgf@circ@res@up}}{\pgfpoint{\pgfkeysvalueof{/tikz/circuitikz/three and gate/port width}\pgf@circ@res@left}{\pgf@circ@res@up}}
  \pgfclosepath
  \pgfsetcolor{\pgfkeysvalueof{/tikz/circuitikz/fill}}
  \pgfsetstrokecolor{black}
  \pgfusepath{draw,fill}
}

\pgfcircdeclarevarlogicportalt{variable and gate alt}{
  \pgfpathmoveto{\pgfpoint{\pgf@circ@res@left}{\pgfkeysvalueof{/tikz/circuitikz/variable and gate alt/input height one}\pgf@circ@res@up}}
  \pgfpathlineto{\pgfpoint{\pgfkeysvalueof{/tikz/circuitikz/variable and gate alt/port width}\pgf@circ@res@left}{\pgfkeysvalueof{/tikz/circuitikz/variable and gate alt/input height one}\pgf@circ@res@up}}

  \pgfpathmoveto{\pgfpoint{\pgf@circ@res@left}{\pgfkeysvalueof{/tikz/circuitikz/variable and gate alt/input height two}\pgf@circ@res@up}}
  \pgfpathlineto{\pgfpoint{\pgfkeysvalueof{/tikz/circuitikz/variable and gate alt/port width}\pgf@circ@res@left}{\pgfkeysvalueof{/tikz/circuitikz/variable and gate alt/input height two}\pgf@circ@res@up}}

  \pgfpathmoveto{\pgfpoint{\pgf@circ@res@left}{\pgfkeysvalueof{/tikz/circuitikz/variable and gate alt/input height three}\pgf@circ@res@down}}
  \pgfpathlineto{\pgfpoint{\pgfkeysvalueof{/tikz/circuitikz/variable and gate alt/port width}\pgf@circ@res@left}{\pgfkeysvalueof{/tikz/circuitikz/variable and gate alt/input height three}\pgf@circ@res@down}}

  \pgfpathmoveto{\pgfpoint{\pgf@circ@res@left}{\pgfkeysvalueof{/tikz/circuitikz/variable and gate alt/input height four}\pgf@circ@res@down}}
  \pgfpathlineto{\pgfpoint{\pgfkeysvalueof{/tikz/circuitikz/variable and gate alt/port width}\pgf@circ@res@left}{\pgfkeysvalueof{/tikz/circuitikz/variable and gate alt/input height four}\pgf@circ@res@down}}

  \pgfpathmoveto{\pgfpoint{\pgf@circ@res@right}{0pt}}
  \pgfpathlineto{\pgfpoint{\pgfkeysvalueof{/tikz/circuitikz/variable and gate alt/port width}\pgf@circ@res@right}{0pt}}

  \pgfusepath{draw}

  \pgfsetlinewidth{2\pgflinewidth}
  \pgfpathmoveto{\pgfpoint{\pgfkeysvalueof{/tikz/circuitikz/variable and gate alt/port width}\pgf@circ@res@left}{\pgf@circ@res@down}}
  \pgfpathcurveto{\pgfpoint{.0\pgf@circ@res@left}{\pgf@circ@res@down}}{\pgfpoint{\pgfkeysvalueof{/tikz/circuitikz/variable and gate alt/port width}\pgf@circ@res@right}{.5\pgf@circ@res@down}}{\pgfpoint{\pgfkeysvalueof{/tikz/circuitikz/variable and gate alt/port width}\pgf@circ@res@right}{0pt}}
  \pgfpathcurveto{\pgfpoint{\pgfkeysvalueof{/tikz/circuitikz/variable and gate alt/port width}\pgf@circ@res@right}{.5\pgf@circ@res@up}}{\pgfpoint{.0\pgf@circ@res@left}{\pgf@circ@res@up}}{\pgfpoint{\pgfkeysvalueof{/tikz/circuitikz/variable and gate alt/port width}\pgf@circ@res@left}{\pgf@circ@res@up}}
  \pgfclosepath
  \pgfsetcolor{\pgfkeysvalueof{/tikz/circuitikz/fill}}
  \pgfsetstrokecolor{black}
  \pgfusepath{draw,fill}
}

\makeatother
 \makeatletter

\ctikzset{thickness/.initial=2}
\ctikzset{color/.initial=black}
\ctikzset{fill/.initial=white}
\pgfkeys{/tikz/color/.add code={}{\ctikzset{color={#1}}}}
\pgfkeys{/tikz/fill/.add code={}{\ctikzset{fill={#1}}}}
\pgfkeys{/tikz/refsize/.initial=1cm}

\newdimen\pgf@ref@up
\newdimen\pgf@ref@down
\newdimen\pgf@ref@left
\newdimen\pgf@ref@right

\pgfdeclareshape{ref}{
  \nodeparts{text}

  \saveddimen{\pgf@ref@right}{\pgf@x=\pgfkeysvalueof{/tikz/refsize}}

  \anchor{center}{\pgfpointorigin}
  \anchor{west}{\pgfpoint{\pgf@ref@left}{0}}

  \anchor{in}{\pgfpoint{\pgf@ref@left}{0}}
  \anchor{out}{\pgfpoint{\pgf@ref@right}{0}}

  \anchor{text}{
    \pgf@x=1.5pt
    \pgf@y=-.5\ht\pgfnodeparttextbox
  }

  \backgroundpath{
    \pgfsetcolor{\pgfkeysvalueof{/tikz/circuitikz/color}}

    \pgfscope
    \pgfsetlinewidth{2\pgflinewidth}

    \pgfpathmoveto{\pgfpoint{\pgf@ref@left}{\pgf@ref@up}}
    \pgfpathlineto{\pgfpoint{\pgf@ref@right-0.225cm}{\pgf@ref@up}}
    \pgfpathlineto{\pgfpoint{\pgf@ref@right}{0}}
    \pgfpathlineto{\pgfpoint{\pgf@ref@right-0.225cm}{\pgf@ref@down}}
    \pgfpathlineto{\pgfpoint{\pgf@ref@left}{\pgf@ref@down}}
    \pgfclosepath

    \pgfsetcolor{\pgfkeysvalueof{/tikz/circuitikz/fill}}
    \pgfsetstrokecolor{black}
    \pgfusepath{draw,fill}

    \endpgfscope
  }
}

\makeatother
 \newcommand{\smallVerticalDots}[1]{
  \node[fill=black, circle, inner sep=0.25pt] at ($#1 + (0, 0.1)$) {};
  \node[fill=black, circle, inner sep=0.25pt] at #1 {};
  \node[fill=black, circle, inner sep=0.25pt] at ($#1 - (0, 0.1)$) {};
}

\newcommand{\mediumDiagonalDots}[1]{
  \node[fill=black, circle, inner sep=0.5pt] at ($#1 + (-0.125, 0.125)$) {};
  \node[fill=black, circle, inner sep=0.5pt] at #1 {};
  \node[fill=black, circle, inner sep=0.5pt] at ($#1 + (0.125, -0.125)$) {};
}

\newcommand{\smallHorizontalDots}[1]{
  \node[fill=black, circle, inner sep=0.25pt] at ($#1 + (-0.1, 0)$) {};
  \node[fill=black, circle, inner sep=0.25pt] at #1 {};
  \node[fill=black, circle, inner sep=0.25pt] at ($#1 + (0.1, 0)$) {};
}

\newcommand{\mediumHorizontalDots}[1]{
  \node[fill=black, circle, inner sep=0.5pt] at ($#1 + (-0.15, 0)$) {};
  \node[fill=black, circle, inner sep=0.5pt] at #1 {};
  \node[fill=black, circle, inner sep=0.5pt] at ($#1 + (0.15, 0)$) {};
}

\newcommand{\mediumVerticalDots}[1]{
  \node[fill=black, circle, inner sep=0.5pt] at ($#1 + (0, -0.15)$) {};
  \node[fill=black, circle, inner sep=0.5pt] at #1 {};
  \node[fill=black, circle, inner sep=0.5pt] at ($#1 + (0, 0.15)$) {};
}
 \makeatletter

\ctikzset{thickness/.initial=2}
\ctikzset{color/.initial=black}
\ctikzset{fill/.initial=white}
\pgfkeys{/tikz/color/.add code={}{\ctikzset{color={#1}}}}
\pgfkeys{/tikz/fill/.add code={}{\ctikzset{fill={#1}}}}

\pgfdeclareshape{rc adder}{
  \anchor{center}{\pgfpointorigin}
  \savedanchor\northwest{
    \pgf@y=1.75\pgf@circ@Rlen
    \pgf@y=.5\pgf@y
    \pgf@x=-1.75\pgf@circ@Rlen
    \pgf@x=.5\pgf@x
  }
 \anchor{in 1}{
   \northwest
   \pgf@x=-0.4\pgf@x
   \pgf@y=\pgf@y
 }
 \anchor{in 2}{
   \northwest
   \pgf@x=0.4\pgf@x
   \pgf@y=\pgf@y
 }
 \anchor{out}{
   \northwest
   \pgf@x=0pt
   \pgf@y=-\pgf@y
 }
 \anchor{in c}{
   \northwest
   \pgf@x=-\pgf@x
   \pgf@y=0pt
 }
 \anchor{out c}{
   \northwest
   \pgf@x=\pgf@x
   \pgf@y=0pt
 }

  \backgroundpath{

    \pgfsetcolor{\pgfkeysvalueof{/tikz/circuitikz/color}}

    \northwest
    \pgf@circ@res@up=\pgf@y 
    \pgf@circ@res@down=-\pgf@y
    \pgf@circ@res@right=-\pgf@x
    \pgf@circ@res@left=\pgf@x

    \pgfscope

    \pgfpathmoveto{\pgfpoint{0.4\pgf@circ@res@left}{\pgf@circ@res@up}} \pgfpathlineto{\pgfpoint{0.4\pgf@circ@res@left}{0.5\pgf@circ@res@up}}
    \pgfsetstrokecolor{black}
    \pgfusepath{draw}

    \pgfpathmoveto{\pgfpoint{0.4\pgf@circ@res@right}{\pgf@circ@res@up}} \pgfpathlineto{\pgfpoint{0.4\pgf@circ@res@right}{0.5\pgf@circ@res@up}}
    \pgfsetstrokecolor{black}
    \pgfusepath{draw}

    \pgfpathmoveto{\pgfpoint{0}{\pgf@circ@res@down}} \pgfpathlineto{\pgfpoint{0}{0.5\pgf@circ@res@down}}
    \pgfsetstrokecolor{black}
    \pgfusepath{draw}

    \pgfpathmoveto{\pgfpoint{\pgf@circ@res@right}{0}} \pgfpathlineto{\pgfpoint{0.75\pgf@circ@res@right}{0}}
    \pgfsetstrokecolor{black}
    \pgfusepath{draw}

    \pgfpathmoveto{\pgfpoint{\pgf@circ@res@left}{0}} \pgfpathlineto{\pgfpoint{0.75\pgf@circ@res@left}{0}}
    \pgfsetstrokecolor{black}
    \pgfusepath{draw}

    \pgfsetlinewidth{2\pgflinewidth}
    \pgfpathmoveto{\pgfpoint{0.75\pgf@circ@res@left}{0.5\pgf@circ@res@up}}
    \pgfpathlineto{\pgfpoint{0.75\pgf@circ@res@right}{0.5\pgf@circ@res@up}}
    \pgfpathlineto{\pgfpoint{0.75\pgf@circ@res@right}{0.5\pgf@circ@res@down}}
    \pgfpathlineto{\pgfpoint{0.75\pgf@circ@res@left}{0.5\pgf@circ@res@down}}
    \pgfclosepath
    \pgfsetcolor{\pgfkeysvalueof{/tikz/circuitikz/fill}}
    \pgfsetstrokecolor{black}
    \pgfusepath{draw,fill}

    \endpgfscope
  }
}
\makeatother
 \makeatletter

\ctikzset{thickness/.initial=2}
\ctikzset{color/.initial=black}
\ctikzset{fill/.initial=white}
\pgfkeys{/tikz/color/.add code={}{\ctikzset{color={#1}}}}
\pgfkeys{/tikz/fill/.add code={}{\ctikzset{fill={#1}}}}

\pgfdeclareshape{rc subtractor}{
  \anchor{center}{\pgfpointorigin}
  \savedanchor\northwest{
    \pgf@y=1.75\pgf@circ@Rlen
    \pgf@y=.5\pgf@y
    \pgf@x=-1.75\pgf@circ@Rlen
    \pgf@x=.5\pgf@x
  }
 \anchor{in 1}{
   \northwest
   \pgf@x=-0.4\pgf@x
   \pgf@y=\pgf@y
 }
 \anchor{in 2}{
   \northwest
   \pgf@x=0.4\pgf@x
   \pgf@y=\pgf@y
 }
 \anchor{out}{
   \northwest
   \pgf@x=0pt
   \pgf@y=-\pgf@y
 }
 \anchor{in b}{
   \northwest
   \pgf@x=-\pgf@x
   \pgf@y=0pt
 }
 \anchor{out b}{
   \northwest
   \pgf@x=\pgf@x
   \pgf@y=0pt
 }

  \backgroundpath{

    \pgfsetcolor{\pgfkeysvalueof{/tikz/circuitikz/color}}

    \northwest
    \pgf@circ@res@up=\pgf@y 
    \pgf@circ@res@down=-\pgf@y
    \pgf@circ@res@right=-\pgf@x
    \pgf@circ@res@left=\pgf@x

    \pgfscope

    \pgfpathmoveto{\pgfpoint{0.4\pgf@circ@res@left}{\pgf@circ@res@up}} \pgfpathlineto{\pgfpoint{0.4\pgf@circ@res@left}{0.5\pgf@circ@res@up}}
    \pgfsetstrokecolor{black}
    \pgfusepath{draw}

    \pgfpathmoveto{\pgfpoint{0.4\pgf@circ@res@right}{\pgf@circ@res@up}} \pgfpathlineto{\pgfpoint{0.4\pgf@circ@res@right}{0.5\pgf@circ@res@up}}
    \pgfsetstrokecolor{black}
    \pgfusepath{draw}

    \pgfpathmoveto{\pgfpoint{0}{\pgf@circ@res@down}} \pgfpathlineto{\pgfpoint{0}{0.5\pgf@circ@res@down}}
    \pgfsetstrokecolor{black}
    \pgfusepath{draw}

    \pgfpathmoveto{\pgfpoint{\pgf@circ@res@right}{0}} \pgfpathlineto{\pgfpoint{0.75\pgf@circ@res@right}{0}}
    \pgfsetstrokecolor{black}
    \pgfusepath{draw}

    \pgfpathmoveto{\pgfpoint{\pgf@circ@res@left}{0}} \pgfpathlineto{\pgfpoint{0.75\pgf@circ@res@left}{0}}
    \pgfsetstrokecolor{black}
    \pgfusepath{draw}

    \pgfsetlinewidth{2\pgflinewidth}
    \pgfpathmoveto{\pgfpoint{0.75\pgf@circ@res@left}{0.5\pgf@circ@res@up}}
    \pgfpathlineto{\pgfpoint{0.75\pgf@circ@res@right}{0.5\pgf@circ@res@up}}
    \pgfpathlineto{\pgfpoint{0.75\pgf@circ@res@right}{0.5\pgf@circ@res@down}}
    \pgfpathlineto{\pgfpoint{0.75\pgf@circ@res@left}{0.5\pgf@circ@res@down}}
    \pgfclosepath
    \pgfsetcolor{\pgfkeysvalueof{/tikz/circuitikz/fill}}
    \pgfsetstrokecolor{black}
    \pgfusepath{draw,fill}

    \endpgfscope
  }
}
\makeatother
 \makeatletter

\ctikzset{thickness/.initial=2}
\ctikzset{color/.initial=black}
\ctikzset{fill/.initial=white}
\pgfkeys{/tikz/color/.add code={}{\ctikzset{color={#1}}}}
\pgfkeys{/tikz/fill/.add code={}{\ctikzset{fill={#1}}}}

\pgfdeclareshape{mux two}{
  \anchor{center}{\pgfpointorigin}
  \savedanchor\northwest{
    \pgf@y=1.75\pgf@circ@Rlen
    \pgf@y=.5\pgf@y
    \pgf@x=-1.75\pgf@circ@Rlen
    \pgf@x=.5\pgf@x
  }
  \anchor{in 1}{
    \northwest
    \pgf@x=\pgf@x
    \pgf@y=0.4\pgf@y
  }
  \anchor{in 2}{
    \northwest
    \pgf@x=\pgf@x
    \pgf@y=-0.4\pgf@y
  }
  \anchor{out}{
    \northwest
    \pgf@x=-\pgf@x
    \pgf@y=0.4\pgf@y
  }
  \anchor{sel in}{
    \northwest
    \pgf@x=0pt
    \pgf@y=-\pgf@y
  }
  \anchor{sel out}{
    \northwest
    \pgf@x=0pt
    \pgf@y=\pgf@y
  }

  \backgroundpath{

    \pgfsetcolor{\pgfkeysvalueof{/tikz/circuitikz/color}}

    \northwest
    \pgf@circ@res@up=\pgf@y 
    \pgf@circ@res@down=-\pgf@y
    \pgf@circ@res@right=-\pgf@x
    \pgf@circ@res@left=\pgf@x

    \pgfscope

    \pgfpathmoveto{\pgfpoint{0.5\pgf@circ@res@left}{0.4\pgf@circ@res@up}} \pgfpathlineto{\pgfpoint{\pgf@circ@res@left}{0.4\pgf@circ@res@up}}
    \pgfsetstrokecolor{black}
    \pgfusepath{draw}

    \pgfpathmoveto{\pgfpoint{0.5\pgf@circ@res@left}{0.4\pgf@circ@res@down}} \pgfpathlineto{\pgfpoint{\pgf@circ@res@left}{0.4\pgf@circ@res@down}}
    \pgfsetstrokecolor{black}
    \pgfusepath{draw}

    \pgfpathmoveto{\pgfpoint{0.5\pgf@circ@res@right}{0.4\pgf@circ@res@up}} \pgfpathlineto{\pgfpoint{\pgf@circ@res@right}{0.4\pgf@circ@res@up}}
    \pgfsetstrokecolor{black}
    \pgfusepath{draw}

    \pgfpathmoveto{\pgfpoint{0}{\pgf@circ@res@up}} \pgfpathlineto{\pgfpoint{0}{0.75\pgf@circ@res@up}}
    \pgfsetstrokecolor{black}
    \pgfusepath{draw}

    \pgfpathmoveto{\pgfpoint{0}{\pgf@circ@res@down}} \pgfpathlineto{\pgfpoint{0}{0.75\pgf@circ@res@down}}
    \pgfsetstrokecolor{black}
    \pgfusepath{draw}

    \pgfsetlinewidth{2\pgflinewidth}
    \pgfpathmoveto{\pgfpoint{0.5\pgf@circ@res@left}{\pgf@circ@res@up}}
    \pgfpathlineto{\pgfpoint{0.5\pgf@circ@res@right}{0.5\pgf@circ@res@up}}
    \pgfpathlineto{\pgfpoint{0.5\pgf@circ@res@right}{0.5\pgf@circ@res@down}}
    \pgfpathlineto{\pgfpoint{0.5\pgf@circ@res@left}{\pgf@circ@res@down}}
    \pgfclosepath
    \pgfsetcolor{\pgfkeysvalueof{/tikz/circuitikz/fill}}
    \pgfsetstrokecolor{black}
    \pgfusepath{draw,fill}

    \endpgfscope
  }
}
\makeatother
 \makeatletter

\ctikzset{thickness/.initial=2}
\ctikzset{color/.initial=black}
\ctikzset{fill/.initial=white}
\pgfkeys{/tikz/color/.add code={}{\ctikzset{color={#1}}}}
\pgfkeys{/tikz/fill/.add code={}{\ctikzset{fill={#1}}}}

\newdimen\pgf@adder@up
\newdimen\pgf@adder@down
\newdimen\pgf@adder@left
\newdimen\pgf@adder@right

\pgfdeclareshape{full adder}{
  \anchor{center}{\pgfpointorigin}
  \anchor{north}{\pgfpoint{0}{\pgf@adder@up}}
  \anchor{south}{\pgfpoint{0}{\pgf@adder@down}}
  \anchor{west}{\pgfpoint{\pgf@adder@left}{0}}
  \anchor{east}{\pgfpoint{\pgf@adder@right}{0}}
  \anchor{north west}{\pgfpoint{\pgf@adder@left}{\pgf@adder@up}}

  \backgroundpath{
    \pgfsetcolor{\pgfkeysvalueof{/tikz/circuitikz/color}}

    \pgfscope
    \pgfsetlinewidth{2\pgflinewidth}

    \pgfpathmoveto{\pgfpoint{.65\pgf@adder@left}{.65\pgf@adder@up}}
    \pgfpathlineto{\pgfpoint{.65\pgf@adder@right}{.65\pgf@adder@up}}
    \pgfpathlineto{\pgfpoint{.65\pgf@adder@right}{.65\pgf@adder@down}}
    \pgfpathlineto{\pgfpoint{.65\pgf@adder@left}{.65\pgf@adder@down}}
    \pgfclosepath

    \pgfsetcolor{\pgfkeysvalueof{/tikz/circuitikz/fill}}
    \pgfsetstrokecolor{black}
    \pgfusepath{draw,fill}
    \endpgfscope

    \pgfpathmoveto{\pgfpoint{.65\pgf@adder@right}{0}}
    \pgfpathlineto{\pgfpoint{\pgf@adder@right}{0}}
    \pgfsetstrokecolor{black}
    \pgfusepath{draw}

    \pgfpathmoveto{\pgfpoint{0}{.65\pgf@adder@up}}
    \pgfpathlineto{\pgfpoint{0}{\pgf@adder@up}}
    \pgfsetstrokecolor{black}
    \pgfusepath{draw}

    \pgfpathmoveto{\pgfpoint{0}{.65\pgf@adder@down}}
    \pgfpathlineto{\pgfpoint{0}{\pgf@adder@down}}
    \pgfsetstrokecolor{black}
    \pgfusepath{draw}

    \pgfpathmoveto{\pgfpoint{.65\pgf@adder@left}{0}}
    \pgfpathlineto{\pgfpoint{\pgf@adder@left}{0}}
    \pgfsetstrokecolor{black}
    \pgfusepath{draw}

    \pgfpathmoveto{\pgfpoint{.65\pgf@adder@left}{.65\pgf@adder@up}}
    \pgfpathlineto{\pgfpoint{\pgf@adder@left}{\pgf@adder@up}}
    \pgfsetstrokecolor{black}
    \pgfusepath{draw}
  }
}

\pgfdeclareshape{half adder north west}{
  \anchor{center}{\pgfpointorigin}
  \anchor{north}{\pgfpoint{0}{\pgf@adder@up}}
  \anchor{south}{\pgfpoint{0}{\pgf@adder@down}}
  \anchor{west}{\pgfpoint{\pgf@adder@left}{0}}
  \anchor{east}{\pgfpoint{\pgf@adder@right}{0}}
  \anchor{north west}{\pgfpoint{\pgf@adder@left}{\pgf@adder@up}}

  \backgroundpath{
    \pgfsetcolor{\pgfkeysvalueof{/tikz/circuitikz/color}}

    \pgfscope
    \pgfsetlinewidth{2\pgflinewidth}

    \pgfpathmoveto{\pgfpoint{.65\pgf@adder@left}{.65\pgf@adder@up}}
    \pgfpathlineto{\pgfpoint{.65\pgf@adder@right}{.65\pgf@adder@up}}
    \pgfpathlineto{\pgfpoint{.65\pgf@adder@right}{.65\pgf@adder@down}}
    \pgfpathlineto{\pgfpoint{.65\pgf@adder@left}{.65\pgf@adder@down}}
    \pgfclosepath

    \pgfsetcolor{\pgfkeysvalueof{/tikz/circuitikz/fill}}
    \pgfsetstrokecolor{black}
    \pgfusepath{draw,fill}
    \endpgfscope

    \pgfpathmoveto{\pgfpoint{.65\pgf@adder@right}{0}}
    \pgfpathlineto{\pgfpoint{\pgf@adder@right}{0}}
    \pgfsetstrokecolor{black}
    \pgfusepath{draw}

    \pgfpathmoveto{\pgfpoint{0}{.65\pgf@adder@down}}
    \pgfpathlineto{\pgfpoint{0}{\pgf@adder@down}}
    \pgfsetstrokecolor{black}
    \pgfusepath{draw}

    \pgfpathmoveto{\pgfpoint{.65\pgf@adder@left}{0}}
    \pgfpathlineto{\pgfpoint{\pgf@adder@left}{0}}
    \pgfsetstrokecolor{black}
    \pgfusepath{draw}

    \pgfpathmoveto{\pgfpoint{.65\pgf@adder@left}{.65\pgf@adder@up}}
    \pgfpathlineto{\pgfpoint{\pgf@adder@left}{\pgf@adder@up}}
    \pgfsetstrokecolor{black}
    \pgfusepath{draw}
  }
}

\pgfdeclareshape{half adder}{
  \anchor{center}{\pgfpointorigin}
  \anchor{north}{\pgfpoint{0}{\pgf@adder@up}}
  \anchor{south}{\pgfpoint{0}{\pgf@adder@down}}
  \anchor{west}{\pgfpoint{\pgf@adder@left}{0}}
  \anchor{east}{\pgfpoint{\pgf@adder@right}{0}}
  \anchor{north west}{\pgfpoint{\pgf@adder@left}{\pgf@adder@up}}

  \backgroundpath{
    \pgfsetcolor{\pgfkeysvalueof{/tikz/circuitikz/color}}

    \pgfscope
    \pgfsetlinewidth{2\pgflinewidth}

    \pgfpathmoveto{\pgfpoint{.65\pgf@adder@left}{.65\pgf@adder@up}}
    \pgfpathlineto{\pgfpoint{.65\pgf@adder@right}{.65\pgf@adder@up}}
    \pgfpathlineto{\pgfpoint{.65\pgf@adder@right}{.65\pgf@adder@down}}
    \pgfpathlineto{\pgfpoint{.65\pgf@adder@left}{.65\pgf@adder@down}}
    \pgfclosepath

    \pgfsetcolor{\pgfkeysvalueof{/tikz/circuitikz/fill}}
    \pgfsetstrokecolor{black}
    \pgfusepath{draw,fill}
    \endpgfscope

    \pgfpathmoveto{\pgfpoint{0}{.65\pgf@adder@up}}
    \pgfpathlineto{\pgfpoint{0}{\pgf@adder@up}}
    \pgfsetstrokecolor{black}
    \pgfusepath{draw}

    \pgfpathmoveto{\pgfpoint{0}{.65\pgf@adder@down}}
    \pgfpathlineto{\pgfpoint{0}{\pgf@adder@down}}
    \pgfsetstrokecolor{black}
    \pgfusepath{draw}

    \pgfpathmoveto{\pgfpoint{.65\pgf@adder@left}{0}}
    \pgfpathlineto{\pgfpoint{\pgf@adder@left}{0}}
    \pgfsetstrokecolor{black}
    \pgfusepath{draw}

    \pgfpathmoveto{\pgfpoint{.65\pgf@adder@left}{.65\pgf@adder@up}}
    \pgfpathlineto{\pgfpoint{\pgf@adder@left}{\pgf@adder@up}}
    \pgfsetstrokecolor{black}
    \pgfusepath{draw}
  }
}

\makeatother
 \makeatletter

\ctikzset{thickness/.initial=2}
\ctikzset{color/.initial=black}
\ctikzset{fill/.initial=white}
\pgfkeys{/tikz/color/.add code={}{\ctikzset{color={#1}}}}
\pgfkeys{/tikz/fill/.add code={}{\ctikzset{fill={#1}}}}

\newdimen\pgf@ca@up
\newdimen\pgf@ca@down
\newdimen\pgf@ca@left
\newdimen\pgf@ca@right

\pgfdeclareshape{ca}{
  \nodeparts{text}

  \anchor{center}{\pgfpointorigin}
  \anchor{a 1}{\pgfpoint{0}{\pgf@ca@up}}
  \anchor{o 1}{\pgfpoint{0}{\pgf@ca@down}}
  \anchor{ci}{\pgfpoint{\pgf@ca@left}{0}}
  \anchor{co}{\pgfpoint{\pgf@ca@right}{0}}
  \anchor{text}{
    \pgf@x=-.5\wd\pgfnodeparttextbox
    \pgf@y=-.5\ht\pgfnodeparttextbox
  }

  \backgroundpath{
    \pgfsetcolor{\pgfkeysvalueof{/tikz/circuitikz/color}}

    \pgfscope
    \pgfsetlinewidth{2\pgflinewidth}

    \pgfpathmoveto{\pgfpoint{\pgf@ca@left}{\pgf@ca@up}}
    \pgfpathlineto{\pgfpoint{\pgf@ca@right}{\pgf@ca@up}}
    \pgfpathlineto{\pgfpoint{\pgf@ca@right}{\pgf@ca@down}}
    \pgfpathlineto{\pgfpoint{\pgf@ca@left}{\pgf@ca@down}}
    \pgfclosepath

    \pgfsetcolor{\pgfkeysvalueof{/tikz/circuitikz/fill}}
    \pgfsetstrokecolor{black}
    \pgfusepath{draw,fill}

    \endpgfscope

    \pgfpathmoveto{\pgfpoint{0}{0}}
  }
}

\newdimen\pgf@cb@up
\newdimen\pgf@cb@down
\newdimen\pgf@cb@left
\newdimen\pgf@cb@right

\pgfdeclareshape{cb}{
  \nodeparts{text}

  \anchor{center}{\pgfpointorigin}
  \anchor{a 1}{\pgfpoint{0.5\pgf@cb@left}{\pgf@cb@up}}
  \anchor{a 2}{\pgfpoint{0.5\pgf@cb@right}{\pgf@cb@up}}
  \anchor{o 1}{\pgfpoint{0.5\pgf@cb@left}{\pgf@cb@down}}
  \anchor{o 2}{\pgfpoint{0.5\pgf@cb@right}{\pgf@cb@down}}
  \anchor{ci}{\pgfpoint{\pgf@cb@left}{0}}
  \anchor{co}{\pgfpoint{\pgf@cb@right}{0}}
  \anchor{text}{
    \pgf@x=-.5\wd\pgfnodeparttextbox
    \pgf@y=-.5\ht\pgfnodeparttextbox
  }

  \backgroundpath{
    \pgfsetcolor{\pgfkeysvalueof{/tikz/circuitikz/color}}

    \pgfscope
    \pgfsetlinewidth{2\pgflinewidth}

    \pgfpathmoveto{\pgfpoint{\pgf@cb@left}{\pgf@cb@up}}
    \pgfpathlineto{\pgfpoint{\pgf@cb@right}{\pgf@cb@up}}
    \pgfpathlineto{\pgfpoint{\pgf@cb@right}{\pgf@cb@down}}
    \pgfpathlineto{\pgfpoint{\pgf@cb@left}{\pgf@cb@down}}
    \pgfclosepath

    \pgfsetcolor{\pgfkeysvalueof{/tikz/circuitikz/fill}}
    \pgfsetstrokecolor{black}
    \pgfusepath{draw,fill}
    \endpgfscope
  }
}

\newdimen\pgf@cc@up
\newdimen\pgf@cc@down
\newdimen\pgf@cc@left
\newdimen\pgf@cc@right

\pgfdeclareshape{cc}{
  \nodeparts{text}

  \anchor{center}{\pgfpointorigin}
  \anchor{a 1}{\pgfpoint{0.625\pgf@cc@left}{\pgf@cc@up}}
  \anchor{a 2}{\pgfpoint{0}{\pgf@cc@up}}
  \anchor{a 3}{\pgfpoint{0.625\pgf@cc@right}{\pgf@cc@up}}
  \anchor{o 1}{\pgfpoint{0.625\pgf@cc@left}{\pgf@cc@down}}
  \anchor{o 2}{\pgfpoint{0}{\pgf@cc@down}}
  \anchor{o 3}{\pgfpoint{0.625\pgf@cc@right}{\pgf@cc@down}}
  \anchor{ci}{\pgfpoint{\pgf@cc@left}{0}}
  \anchor{co}{\pgfpoint{\pgf@cc@right}{0}}
  \anchor{text}{
    \pgf@x=-.5\wd\pgfnodeparttextbox
    \pgf@y=-.5\ht\pgfnodeparttextbox
  }

  \backgroundpath{
    \pgfsetcolor{\pgfkeysvalueof{/tikz/circuitikz/color}}

    \pgfscope
    \pgfsetlinewidth{2\pgflinewidth}

    \pgfpathmoveto{\pgfpoint{\pgf@cc@left}{\pgf@cc@up}}
    \pgfpathlineto{\pgfpoint{\pgf@cc@right}{\pgf@cc@up}}
    \pgfpathlineto{\pgfpoint{\pgf@cc@right}{\pgf@cc@down}}
    \pgfpathlineto{\pgfpoint{\pgf@cc@left}{\pgf@cc@down}}
    \pgfclosepath

    \pgfsetcolor{\pgfkeysvalueof{/tikz/circuitikz/fill}}
    \pgfsetstrokecolor{black}
    \pgfusepath{draw,fill}
    \endpgfscope
  }
}

\newdimen\pgf@cd@up
\newdimen\pgf@cd@down
\newdimen\pgf@cd@left
\newdimen\pgf@cd@right

\pgfdeclareshape{cd}{
  \nodeparts{text}

  \anchor{center}{\pgfpointorigin}

  \anchor{a 1}{\pgfpoint{0.75\pgf@cd@left}{\pgf@cd@up}}
  \anchor{a 2}{\pgfpoint{0.25\pgf@cd@left}{\pgf@cd@up}}
  \anchor{a n}{\pgfpoint{0.75\pgf@cd@right}{\pgf@cd@up}}

  \anchor{o 1}{\pgfpoint{0.75\pgf@cd@left}{\pgf@cd@down}}
  \anchor{o 2}{\pgfpoint{0.25\pgf@cd@left}{\pgf@cd@down}}
  \anchor{o n}{\pgfpoint{0.75\pgf@cd@right}{\pgf@cd@down}}

  \anchor{ci}{\pgfpoint{\pgf@cd@left}{0}}
  \anchor{co}{\pgfpoint{\pgf@cd@right}{0}}

  \anchor{text}{
    \pgf@x=-.5\wd\pgfnodeparttextbox
    \pgf@y=-.5\ht\pgfnodeparttextbox
  }

  \backgroundpath{
    \pgfsetcolor{\pgfkeysvalueof{/tikz/circuitikz/color}}

    \pgfscope
    \pgfsetlinewidth{2\pgflinewidth}

    \pgfpathmoveto{\pgfpoint{\pgf@cd@left}{\pgf@cd@up}}
    \pgfpathlineto{\pgfpoint{\pgf@cd@right}{\pgf@cd@up}}
    \pgfpathlineto{\pgfpoint{\pgf@cd@right}{\pgf@cd@down}}
    \pgfpathlineto{\pgfpoint{\pgf@cd@left}{\pgf@cd@down}}
    \pgfclosepath

    \pgfsetcolor{\pgfkeysvalueof{/tikz/circuitikz/fill}}
    \pgfsetstrokecolor{black}
    \pgfusepath{draw,fill}
    \endpgfscope
  }
}

\makeatother
 \makeatother

\theoremstyle{definition}
\newtheorem{definition}{Definition}
\newtheorem{problem}{Problem}
\theoremstyle{plain}
\newtheorem{theorem}{Theorem}
\newtheorem{property}{Property}

\newenvironment{sketch}[1]{\begin{proof}#1}{\end{proof}}

\newcommand{\python}{\textsc{Python}\xspace}
\newcommand{\C}{\textsc{C}\xspace}
\newcommand{\CPP}{\textsc{C++}\xspace}
\newcommand{\depqbf}{\textsc{DepQBF}\xspace}
\newcommand{\rareqs}{\textsc{RAReQS}\xspace}
\newcommand{\bloqqer}{\textsc{Bloqqer}\xspace}
\newcommand{\plq}{\textsc{PLQ}\xspace}
\newcommand{\qfun}{\textsc{QFun}\xspace}
\newcommand{\kissat}{\textsc{Kissat}\xspace}

\newcommand{\inputs}{\ensuremath{\mathrm{IN}}\xspace}
\newcommand{\outputs}{\ensuremath{\mathrm{OUT}}\xspace}
\newcommand{\gates}{\ensuremath{\mathrm{GATES}}\xspace}

\definecolor{antiquewhite}{rgb}{0.98, 0.92, 0.84}
\definecolor{pastelgray}{rgb}{0.81, 0.81, 0.77}
\definecolor{champagne}{rgb}{0.97, 0.91, 0.81}
\definecolor{cream}{rgb}{1.0, 0.99, 0.82}
\definecolor{darkkhaki}{rgb}{0.74, 0.72, 0.42}
\definecolor{tearose(rose)}{rgb}{0.96, 0.76, 0.76} 
\journal{Journal of Artificial Intelligence}

\begin{document}

\begin{frontmatter}

\title{Design Space Exploration as Quantified Satisfaction}

\author{Alexander Feldman}
\author{Johan de Kleer}
\author{Ion Matei}
\address{e-mail: \small\texttt{a.feldman,dekleer,imatei@parc.com}}
\address{Palo Alto Research Center Inc.}
\address{3333 Coyote Hill Road, Palo Alto, CA 94304, USA}

\begin{abstract}

  We propose novel algorithms for design and design space
  exploration. The designs computed by these algorithms are
  compositions of function types specified in component libraries. Our
  algorithms reduce the design problem to quantified satisfiability
  and use advanced solvers to find solutions that represent useful
  systems.

  The algorithms we present in this paper are sound and complete and
  are guaranteed to discover correct designs of optimal size, if they
  exist. We apply our method to the design of Boolean systems and
  discover new and more optimal classical and quantum circuits for
  common arithmetic functions such as addition and multiplication.

  The performance of our algorithms is evaluated through extensive
  experimentation. We have first created a benchmark consisting of
  specifications of scalable synthetic digital circuits and real-world
  microchips. We have then generated multiple circuits functionally
  equivalent to the ones in the benchmark. The quantified
  satisfiability method shows more than four orders of magnitude
  speed-up, compared to a generate and test method that enumerates all
  non-isomorphic circuit topologies.

  Our approach generalizes circuit optimization. It uses arbitrary
  component libraries and has applications to areas such as digital
  circuit design, diagnostics, abductive reasoning, test vector
  generation, and combinatorial optimization.
  
\end{abstract}


\end{frontmatter}

\section{Introduction}

Design is a next frontier in artificial intelligence. Providing
algorithms and tools for conceiving novel designs benefits many areas
such as analog and digital chip design, software development,
mechanical design, and systems engineering. Human designers will be
assisted in better navigating complex trade-offs such as speed versus
number of transistors versus heat dissipation in an Integrated Circuit
(IC). Users will choose from a richer base of trade-offs and this will
lead to dramatic improvements in micro-electronics and computing.

Computation, representation, and tools have improved tremendously over
the last decades so now, one can consider systematic enumeration of
the design space. This paper provides a novel encoding scheme for
efficient exploration of the design space of digital circuits.

The algorithms presented in this paper are more computationally
intensive compared to heuristic search \citep{hansen2007anytime} and
genetic algorithms \citep{miller2000principles} but provide sound and
complete enumeration of the design space. Our algorithms exhaustively
``prove'' that certain designs can or cannot be made with $k$ components
where components are drawn from an arbitrary component library.

Traditional books on digital design, for example, teach the
construction of a full-subtractor with seven components
\citep{maini2007digital} and we found one with only five gates. The
five and seven component version of the subtractor will have the same
number of transistors but there are other technologies (such as 3-D,
or quantum) where the five-component version will have smaller
footprint and faster propagation times.

As a special case, the circuit generation algorithm presented in this
paper, reduces to circuit minimization but its performance should not
be compared to other optimization algorithms such as Quine-McCluskey
\citep{mccluskey1956minimization} or \textsc{Espresso}
\citep{brayton1984logic}. To illustrate the generality of our approach
we have used it to design a reversible quantum circuits of minimal size
\citep{nielsen2010quantum}.

Modern satisfiability (SAT) theory \citep{biere2009handbook} is widely
used in research and in industry. There are SAT solvers that can solve
industrial problems with millions of variables
\citep{jarvisalo2012international}. The algorithms in this paper
construct circuit designs by solving Quantified Boolean Formulas
(QBFs). QBF satisfiability is a generalization over satisfiability of
propositional formulas where universal and existential quantifiers are
allowed. The QBFs that our algorithms generate are of interest to
designers of quantified satisfiability (QSAT) algorithms as there is
always the need of benchmarks with practical applications
\citep{janota2016qbfgallery}.

The algorithms of this paper are validated on an extensive benchmark
of combinational circuits with more than seventy successful
experiments. We have designed generators of combinational circuits of
various size such as adders, multipliers, and multiplexers. These
circuits are the basic building blocks of Arithmetic Logic Units and
Field Programmable Gate Arrays (FPGAs). In addition to that we
consider four digital Integrated Circuits (ICs) from the well-known
74XXX family. We have shown that our QBF-based circuit generation
algorithm is multiple orders of magnitude faster compared to a
graph-based generate and test algorithm to find minimal circuits.
 \section{Design Generation and Exploration}

Technical designs materialize from requirements, specifications, and
the designers' experience. The design process is iterative with
versions continuously improving and being refined. Incomplete designs
often do not meet the requirements and designers ``debug'' and fix
them. The formal underlying problem behind finishing an incomplete
design \citep{gitina2013equivalence} has been studied in the logic and
verification communities \citep{finkbeiner2014fast}. In addition to
producing an initial design from scratch or continuing an incomplete
one, designers often create multiple alternatives for the users and
builders to choose from. The later process is called design
exploration.

A design is typically specified in some kind of
requirements. Depending on the design domain, the requirements can be
a mechanical blueprint, an electrical diagram, algorithmic pseudo-code
or human readable text. To automate the generation and enumeration of
designs, which is the main goal of this paper, we need some formal
specification of a function or a design itself.

\begin{figure}[hbt]
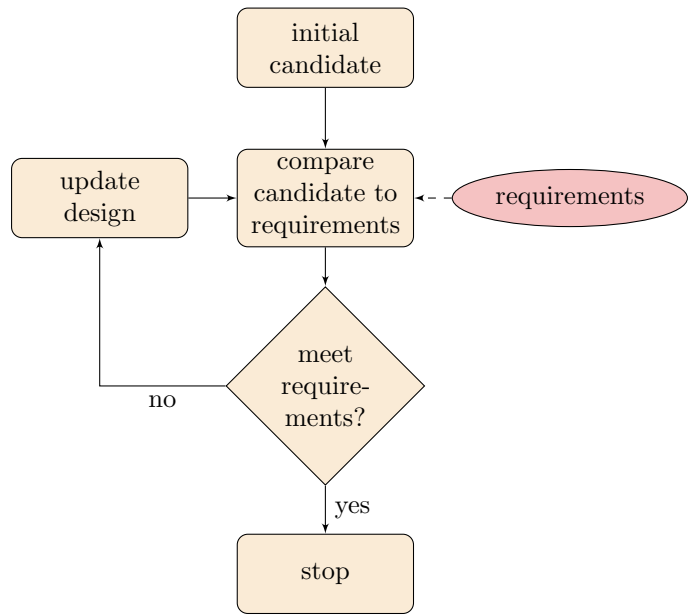

  \centering
  \includestandalone{figures/generate_test}  
  \caption{The design process as ``generate and test''}
  \label{fig:generate_test}
\end{figure}

Figure~\ref{fig:generate_test} illustrates the design generation
process. The process is usually supported by Computer Aided Design
(CAD) tools, Artificial Intelligence (AI), and combinatorial
optimization algorithms. In some cases it is possible to consider the
whole design space and completely exhaust the search. Complete
algorithms for design and design exploration are the subject of this
paper.

\begin{figure}[hbt]
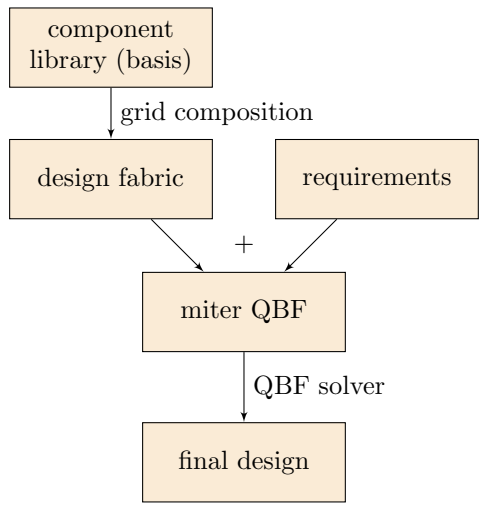

  \centering
  \includestandalone{figures/information_flow}  
  \caption{Information flow during the design process}
  \label{fig:information_flow}
\end{figure}

The information flow in solving a design problem is shown in
Figure~\ref{fig:information_flow}. The component library (basis) is
specified as a set of Boolean functions. An automated procedure is
then used to generate a regular fabric of configurable components and
topological interconnections (wires). The configurable fabric is
appended to the user requirements which are also specified as a
Boolean circuit or a Boolean function. The result is a miter: a
formula that checks for Boolean function equivalence. The miter
formula is fed to a QBF solver. The QBF solver computes a certificate
that contains the configuration of the fabric. The final design is
constructed from the certificate of the miter formula.

There is only one computationally intensive step in generating a
design: solving the QBF miter formula. Finding a satisfiable solution
of a QBF is relevant to both satisfiability and game theory and is a
prototypical PSPACE-complete problem \citep{garey90computers}.

Consider an arbitrary QBF formula:
\begin{eqnarray}
Q_1 x_1 Q_2 x_2 \ldots Q_n x_n \varphi (x_1, x_2, \ldots, x_n)
\end{eqnarray}
where $Q_1, Q_2, \ldots, Q_n$ are either existential ($\exists$) or
universal ($\forall$) quantifiers. It can be decided if a formula is
true or not by iteratively ``unpeeling'' the outermost quantifier
until no quantifiers remain. If we condition on the value of the first
quantifier, we have:
\begin{eqnarray}
A = Q_2 x_2 \ldots Q_n x_n \varphi (0, x_2, \ldots, x_n) \\
  B = Q_2 x_2 \ldots Q_n x_n \varphi (1, x_2, \ldots, x_n)
\end{eqnarray}
The formula is then reduced to $A \wedge B$ if $Q_1$ is $\forall$ and
$A \vee B$ if $Q_1$ is $\exists$. This process of recursive formula
evaluation resembles a game where alternating the quantifier types
forces the solver between making the solver look for primal and dual
solution of the formula $\varphi$.

The recursive procedure suggested above is inefficient. Modern QBF
solvers \cite{janota2016solving} use advanced search methods such as
QCDCL (Quantified Conflict-Driven Clause Learning). QBF solvers
benefit from knowledge compilation such as OBDD
\citep{coste2005propositional}, conflict learning, and even machine
learning \citep{samulowitz2007learning}. Some solvers cater to a
subclass of QBF formulas such as 2-QBF where there is only one switch
between existential and universal quantifiers, others
\citep{janota2018towards} are non-clausal and take directly circuits
as their input.

Looking deeper, the QBF solving process resembles the high-level
generate and test process of design. Although it is not trivial to
reduce design generation and exploration to solving a QBF, in this
paper we manage to do that and use the advances in QBF solving to
discover novel circuits or circuit topologies.
 \section{Fundamental Concepts}

Definitions~\ref{def:boolean_function}--\ref{def:boolean_circuit}
are directly adopted from \citet{vollmer2013introduction} and formally
introduce the notions of a Boolean function and a Boolean circuit.

\begin{definition}[Boolean Function]
  \label{def:boolean_function}

  A multi-output Boolean function is a function
  $f: \{0, 1\}^m \to \{0, 1\}^n$ for some $\{m, n\} \in \mathbb{N}$.

\end{definition}

Notice that, while in \citet{vollmer2013introduction} a Boolean
function has a single output, we do not have this restriction. Another
difference is that we do not use function families, i.e., all our
objects are finite.

Some common Boolean functions are negation ($\neg$), disjunction
($\vee$), conjunction ($\wedge$), exclusive or ($\oplus$),
implication\footnote{This paper, similar to many others, shares the
  same symbol ($\rightarrow$) for implication and for function
  mapping. The use is clear from the context.} ($\rightarrow$), and
equivalence ($\leftrightarrow$). This paper uses everywhere infix, as
opposed to prefix, notation. For example, $p \vee q$ is used instead
of $\vee\left(p, q\right)$.

We also use equivalence ($\leftrightarrow$) instead of the equal sign
($=$) to specify Boolean functions. The function output is on the left
while the inputs are on the right. For example, the Boolean function
$r = p \vee q$ is written as $r \leftrightarrow p \vee q$. When there
are multiple outputs, we give a formula for each one of them.

\begin{figure}[hbt]
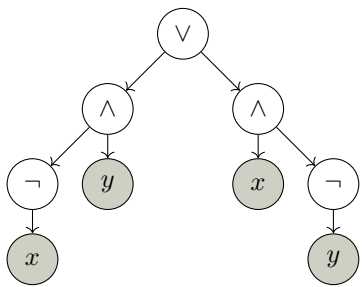

  \centering
  \includestandalone{figures/boolean_function}\caption{An example of a Boolean function}
  \label{fig:boolean_function}
\end{figure}

Figure~\ref{fig:boolean_function} shows the Boolean function
$f \leftrightarrow \neg x \wedge y \vee x \wedge \neg y$ as a
tree. Notice that only the leaf nodes are variables while all
non-leafs are operators.

\begin{definition}[Basis]
  \label{def:basis}

  A basis $B$ is defined as a finite set of Boolean functions.

\end{definition}

\begin{figure}[hbt]
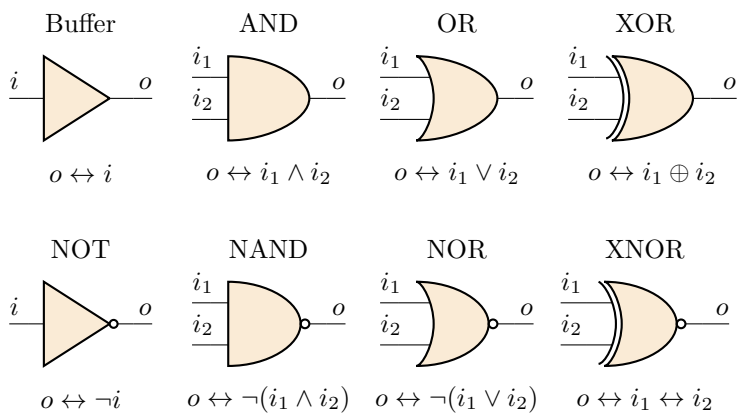

  \centering
  \includestandalone{figures/standard_basis}\caption{The standard basis}
  \label{fig:standard_basis}
\end{figure}

Later in this section we discuss the fine differences between a
Boolean circuit and a Boolean function as the two concepts are similar
in many ways. One of the most important differences is that circuits
use bases while functions do not. A basis $B$ can be thought of as the
elementary unit of sharing or as an abstract \textbf{component
  library}. Unlike in the real world, though, each basis function can
be used infinitely many times and all functions in a basis have the
same cost. Figure~\ref{fig:standard_basis} shows a basis consisting of
typical unary and binary Boolean functions.

\begin{figure}[hbt]
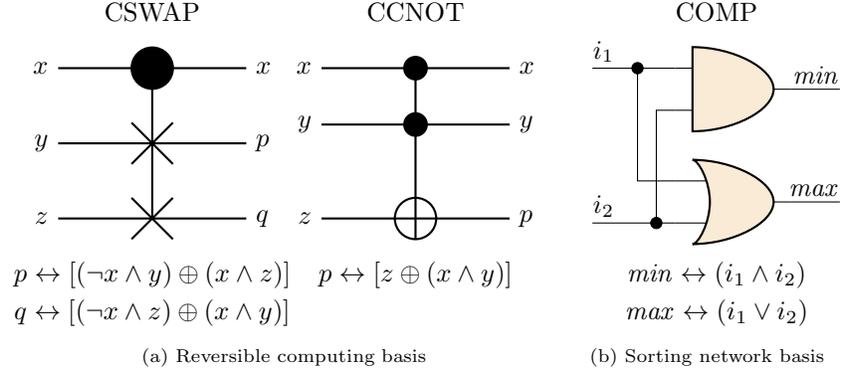

  \centering
  \begin{subfigure}[b]{7.5cm}
    \centering
    \includestandalone{figures/reversible_basis}\caption{Reversible computing basis}
    \label{fig:reversible_basis}
  \end{subfigure}\begin{subfigure}[b]{3.75cm}
    \centering
    \includestandalone{figures/sorting_basis}\caption{Sorting network basis}
    \label{fig:comparator}
  \end{subfigure}
  \caption{Non-standard bases}
  \label{fig:non_standard_bases}
\end{figure}

Figure~\ref{fig:non_standard_bases} shows bases with
multi-input/multi-output components. Figure~\ref{fig:reversible_basis}
shows a basis consisting of two multi-output functions. They implement
the Fredkin and the Toffoli gates
\citep{fredkin1982conservative,toffoli1980reversible}. These gates,
also known as CSWAP and CCNOT gates, have application in reversible
and quantum computing.

Figure~\ref{fig:comparator} shows a basis that contains one component
only: a one-bit comparator. Sorting networks are made of chains of
comparators. Proving lower-bounds on the number of comparators
necessary for the building of a $k$-input sorting network is an
ongoing challenge \citep{codish2014twenty}. The methods described in
this paper provide novel methods for the optimal design and analysis
of sorting networks.

\begin{figure}[hbt]
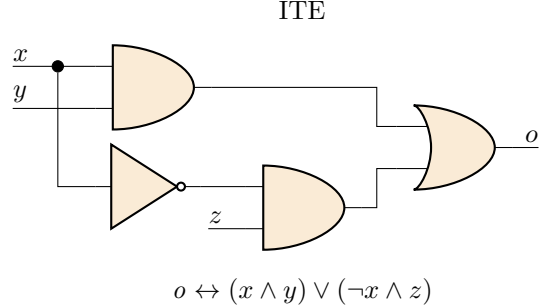

  \centering
  \includestandalone{figures/ite_basis}\caption{The ``if-then-else'' basis}
  \label{fig:ite_basis}
\end{figure}

It is possible to construct an ``if-then-else'' basis from the
function shown in Figure~\ref{fig:ite_basis} and the two Boolean
constants ($\top$ and $\perp$). If a circuit uses this base and the
output of each gate is connected to exactly one input of another gate,
then the problem of synthesizing minimal Binary Decision Diagrams
\cite{akers1978binary} can be cast as circuit design.

It is possible to work with higher-level components. In the design of
an Arithmetic-Logic Unit (ALU), for example, one can consider a basis
extending the standard gates with multi-bit adders, multipliers,
barrel shifters, etc.

\begin{definition}[Boolean Circuit]
  \label{def:boolean_circuit}

  Given a basis $B$, a Boolean circuit $C$ over $B$ is defined as
  $C = \langle V, E, \alpha, \beta, \chi, \omega \rangle$, where
  $\langle V, E \rangle$ is a finite directed acyclic graph,
  $\alpha: E \to \mathbb{N}$ is an injective function,
  $\beta: V \to B \cup \{\star\}$,
  $\chi: V \to \{x_1, x_2, \ldots, x_n\} \cup \{\star\}$, and
  $\omega: V \to \{y_1, y_2, \ldots, y_m\} \cup \{\star\}$.  The
  following conditions must hold:

  \begin{enumerate}
    \item{If $v \in V$ has an in-degree $0$, then $\chi(v) \in \{x_1,
        x_2, \ldots, x_n\}$ or $\beta(v)$ is a $0$-ary Boolean
        function (i.e., a Boolean constant) in $B$;}
    \item{If $v \in V$ has an in-degree $k > 0$, then $\beta(v)$ is a
        $k$-ary Boolean function from $B$;}
    \item{For every $i, 1 \le i \le n$, there is exactly one node $v
        \in V$ such that $\chi(v) = x_i$;}
    \item{For every $i, 1 \le i \le m$, there is exactly one node $v
        \in V$ such that $\omega(v) = y_i$.}
  \end{enumerate}
  
\end{definition}

The function $\alpha$ determines the ordering of the edges that go
into a node when the ordering matters (such as in
implication). The function $\alpha$ is not necessary if $B$ consists
of symmetric functions only.

The function $\beta$ determines the type of each node in the circuit:
a function in the basis $B$. The function $\chi$ specifies the set of
input nodes $\{x_1, x_2, \ldots, x_n\}$. The function $\omega$
specifies the set of output nodes $\{y_1, y_2, \ldots, y_n\}$. A node
$v$ is non-output, or computational, if $\chi(v) = \star$ and
$\omega(v) = \star$.

\begin{figure}[hbt]
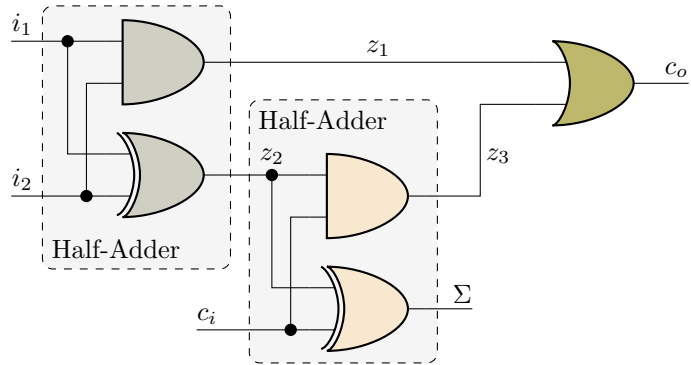

  \centering
  \includestandalone{figures/full_adder_classical}
  \caption{A full-adder}
  \label{fig:full_adder}
\end{figure}

Figure~\ref{fig:full_adder} shows a simple and frequently used circuit
that is used for adding the two binary numbers $i_1$ and $i_2$ and a
carry input bit $c_i$. The output is found in the sum bit $\Sigma$ and
in the carry output $c_o$. Notice that there are two identical
subcircuits in Figure~\ref{fig:full_adder}. These are the two
half-adders.

\begin{figure}[hbt]
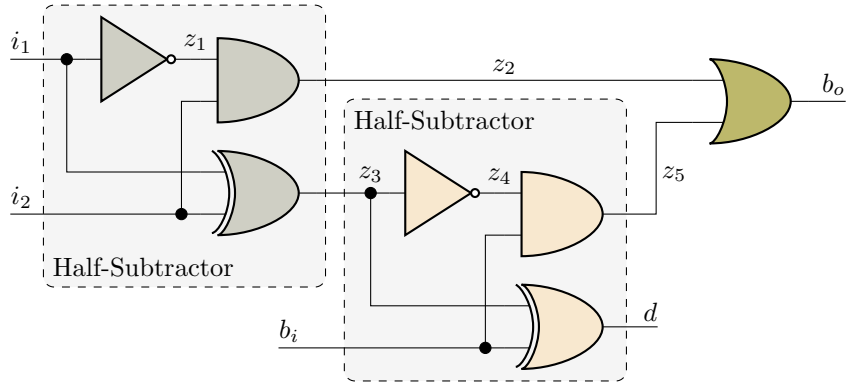

  \centering
  \includestandalone{figures/full_subtractor_classical}
  \caption{A full-subtractor}
  \label{fig:full_subtractor}
\end{figure}

Figure~\ref{fig:full_subtractor} shows another circuit
that is used for subtracting two binary numbers ($i_1$ and $i_2$) and
a borrow input bit $b_i$. The output nodes are the difference $d$ and
the borrow output $b_0$.

The circuits shown in Figure~\ref{fig:full_adder} and
Figure~\ref{fig:full_subtractor} use the standard basis. They are used
as running examples for the rest of the paper.

Notice that, in a circuit, we use the term gate instead of component.
Also, in a circuit the output of each gate is connected to the inputs
of one or more other gates, i.e., a gate drives multiple other
gates. The number of gates that are connected to a certain output is
the gate's \textit{fan-out}.

The size of a circuit is the number of gates.

In a Boolean function the result of an operator can be used as an
argument of only one another operator. Fan-out does not make sense
in a Boolean function. Of course, while it is possible to create an
equivalent Boolean function for a circuit with gates with fan-out of
more than one, it would require the introduction of new variables and
operators. If we measure the size of the Boolean function as the
number of operators, then circuits with gates with fan-out of more
than one will require fewer wires (variables) and gates. Alternatively,
a circuit distinguishes between which variable is a primary input and
which not, while in a (single-output) Boolean function a variable is a
input.

From a higher-level standpoint, the main difference between Boolean
functions and circuits is that \textbf{function sharing} is only
supported in circuits. It is possible and straightforward to convert a
circuit to an equivalent Boolean function but the number of operators
in the Boolean function is often larger than the number of gates in
the circuit. The full-adder shown in Figure~\ref{fig:full_adder}, for
example, requires at least six operators:
$\Sigma \leftrightarrow i_1 \oplus i_2 \oplus c_i$ and
$c_o \leftrightarrow i_1 \wedge i_2 \vee (i_1 \oplus i_2) \wedge
c_i$. The XOR gate that adds $i_1$ and $i_2$ is used both in
calculating the sum $\Sigma$ and the carry-output bit $c_o$.  In some
pathological cases, the blow-up can be exponential. The other
direction is trivial: all Boolean functions are also circuits with all
gates having a fan-out of exactly one.

Sometimes we would like to talk about how the nodes in a circuit are
connected, without concerning ourselves with the exact function of
each node. This is referred to as the \textbf{topology} of a circuit.

\begin{definition}[Topology]
  \label{def:topology}

  Given a circuit
  $C = \langle V, E, \alpha, \beta, \chi, \omega \rangle$, the
  topology of $C$ is defined by the $C$ sub-tuple
  $G = \langle V, E, \chi, \omega \rangle$.

\end{definition}

\begin{figure}[hbt]
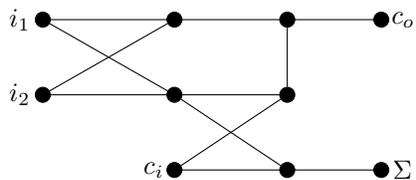

  \centering
  \includestandalone{figures/full_adder_topology}
  \caption{Full adder topology}
  \label{fig:full_adder_topology}
\end{figure}

The graph in Figure~\ref{fig:full_adder_topology} shows the topology
of the full-adder circuits shown in Figure~\ref{fig:full_adder}. There
are three types of nodes: the input nodes $i_1$ and $i_2$, the
internal nodes that correspond to gates, and the ouput nodes $\Sigma$
and $c_o$.

The main purpose of this paper is to present an algorithm for
synthesizing circuits of minimal size.
 \section{Component Selection Problems and the Universal Component Cell}
\label{sec:component_selection}

Suppose we are given a basis $B$, a topology
$G = \langle{V, E, \chi, \omega}\rangle$, and a requirements circuit
$\psi$. The purpose of our first algorithm is, given $B$, $G$, and
$\psi$ to create a circuit $\varphi$, such that $\varphi \equiv \psi$.

Consider the full-adder from Figure~\ref{fig:full_adder} as the
requirements circuit $\psi$. Obtaining the topology $G$ from $\psi$ is
trivial as the circuit topology is a sub-tuple of the circuit (see
Definition~\ref{def:topology}). Let $B$ be the standard basis shown in
Figure~\ref{fig:standard_basis}. Given that the requirements circuit,
itself, uses $B$, there exists at least one full-adder that uses the
standard basis: that is the requirements $\psi$, itself. It is the
trivial solution. We will see that there also exist multiple
non-trivial solutions.

\begin{figure}[hbt]
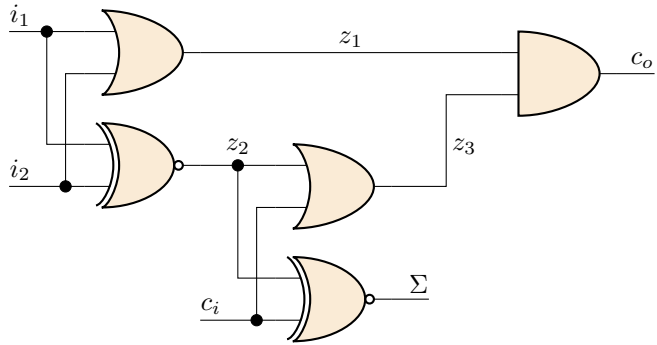

  \centering
  \includestandalone{figures/full_adder_alternative}
  \caption{An alternative implementation of a full-adder}
  \label{fig:full_adder_alternative}
\end{figure}

Figure~\ref{fig:full_adder_alternative} shows an alternative,
non-trivial, implementation $\varphi$ of the full-adder $\psi$ with
gates different from the ones in Figure~\ref{fig:full_adder}. Instead
of using two AND-gates, two XOR-gates, and an OR-gate, the alternative
implementation makes two identical subsystems, each one containing an
OR-gate and an XNOR-gate. The final carry output bit is computed by an
AND-gate.

We can think of the circuit shown in
Figure~\ref{fig:full_adder_alternative} as a symmetrical equivalence
of the circuit shown in Figure~\ref{fig:full_adder}. In what follows,
we present an algorithm that computes and counts these symmetric
circuit alternatives. This algorithm, based on QBF, is surprisingly
efficient. We will see in the empirical results of
Section~\ref{sec:experiments} that circuits implementing common
arithmetic and logical operations have many ``deep'' symmetries.

\begin{problem}[Component Selection Problem]
  \label{problem:component_selection}
  
  Given a basis $B$, topology
  $G = \langle V, E, \chi, \omega \rangle$, and requirements $\psi$,
  construct a circuit
  $\varphi = \langle V, E, \alpha, \beta, \chi, \omega \rangle$, such
  that $\varphi \equiv \psi$.
  
\end{problem}

Problem~\ref{problem:component_selection} is concerned with finding
the type of each component in $\varphi$, or automatically specifying
the functions $\alpha$, and $\beta$. In some papers
\citep{haaswijk2018sat}, Problem~\ref{problem:component_selection} is
referred to as ``labeling'' because one can think of the type of a
gate as a label in a graph-like topology.

A design exploration problem is to count all possible circuit
implementations. Counting has little practical application on its own
but the count is an important factor that characterizes the
performance of circuit synthesis.

\begin{problem}[Counting Component Selection Configurations]
  \label{problem:counting_configurations}

  Given a basis $B$, topology
  $G = \langle V, E, \chi, \omega \rangle$, and requirements $\psi$,
  count the number of distinct circuits
  $\varphi_i = \langle V, E, \alpha_i, \beta_i, \chi, \omega \rangle$,
  $1 \le i \le n$, such that $\varphi_i \equiv \psi$.

\end{problem}

A na\"ive approach to solving Problems
\ref{problem:component_selection} and
\ref{problem:counting_configurations} is to consider all possible
combinations of component types. There is, of course, the need to
perform an equivalence check for each combination of components and
there are exponentially many combinations. Equivalence checking is a
coNP-hard problem but it is often easy in practice
\citep{matsunaga1996efficient}. The problem of equivalence checking
has been largely solved either by using compilation to Ordered Binary
Decision Diagrams (OBDDs) as proposed by \citet{bryant1986graph} or
through resolution methods \citep{marques1999combinational}. Despite
the practical ease of equivalence checking, solving any instance of
Problem~\ref{problem:component_selection} would still require an
exponential number of coNP-hard calls.

The main idea behind our approach for solving Problems
\ref{problem:component_selection} and
\ref{problem:counting_configurations} is the universal component cell:
a component that introduces extra selector inputs allowing the
choosing of which basis operation to perform. Connecting multiple
cells according to the user-specified topology allows the extraction
of one solution of Problem~\ref{problem:component_selection} from the
return value of a single QBF solver call.

\subsection{The Universal Component Cell}

\begin{figure}[hbt]
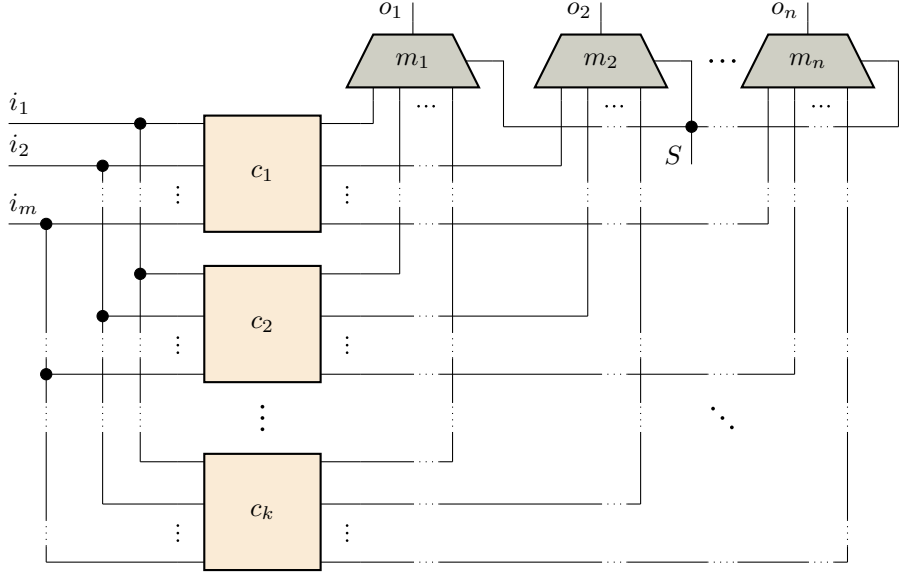

  \centering
  \includestandalone{figures/universal_cell}
  \caption{The universal component cell}
  \label{fig:universal_cell}
\end{figure}

The universal component cell is a Boolean circuit that can be
configured to perform as any of the functions in a basis $B$. It is
shown in Figure~\ref{fig:universal_cell}.

In Figure~\ref{fig:universal_cell} there is a component
$c_1, c_2, \ldots, c_k$ for each component of the basis. Suppose that
each component of the basis has $m$ inputs and $n$ outputs. All
outputs go to a set of $n$ multiplexers $m_1, m_2, \ldots, m_n$.

\begin{figure}[htb]
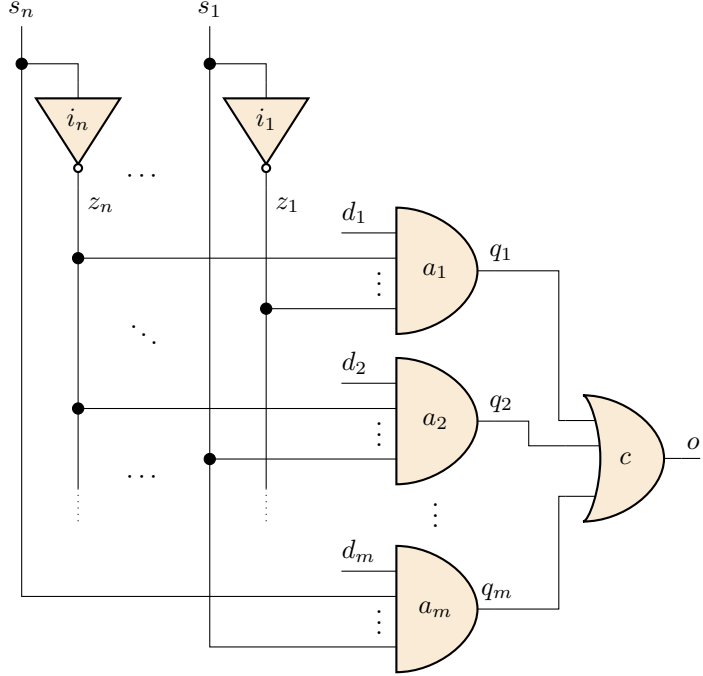

\centering
  \includestandalone{figures/n_mux}\caption{Variable size multiplexer circuit}
  \label{fig:n_mux}
\end{figure}

The configuration of the universal cell is a binary value assigned to
a vector of selector lines $S$. The number of selector inputs is
$|S| = \lceil\log_2{n}\rceil$ where $n$ is the number of distinct
component types in the basis. The actual routing is done by
variable-size multiplexer circuits similar to the ones shown in
Figure~\ref{fig:n_mux}.

Figure~\ref{fig:n_mux} shows a multiplexer of variable size. Suppose
there are $n$ alternative gates and $|S| = \lceil\log_2{n}\rceil$
selector lines. The multiplexer needs $n$ multi-input AND-gates and
$|S|$ inverters. All AND-gates have $|S| + 1$ inputs. The multiplexer
also uses an OR-gate with $n$ inputs. The space complexity of the
circuits is $O(|S| \times n)$ when multi-input gates are realized with
ladders of two-input ones.

When constructing the cells, we take special care if the components in
$B$ have different numbers of inputs and outputs and if $|B| \ne 2^k$,
$k \in \mathbb{N}$. The special care is that we augment the miter
circuit with gates that ``disable'' these hanging wires.

\subsection{An Efficient QBF-Based Algorithm}

In what follows we reduce Problem~\ref{problem:component_selection} to
finding a satisfiable solution of a QBF problem. Most QBF solvers, in
addition to determining if a given QBF is satisfiable or not, also
compute a partial certificate, or a witness: an assignment to the
variables in the outermost quantifier that satisfies or invalidates
the formula. We use this assignment for constructing the solution of
our problem. The circuit whose partial certificate is a solution of
Problem~\ref{problem:component_selection} is shown in
Figure~\ref{fig:miter}.

\begin{figure}[hbt]
  \centering
  \includestandalone{figures/augmented_circuit}
  \caption{Miter}
  \label{fig:miter}
\end{figure}

The two subcircuits shown in Figure~\ref{fig:miter}
illustrate the concept of a miter \citep{brand1993verification}. The
miter is constructed from the requirements circuit $\psi$ and a circuit
$\varphi$ which uses the topology of $\psi$ and instead of gates has
universal cells. The corresponding pairs of primary inputs of
$\varphi$ and $\psi$ are joined together and the primary outputs are
connected to XNOR gates whose outputs are connected to the constant $\top$.

The miter is used for equivalence checking. The basic idea of a miter
is to pairwise tie all inputs and outputs of the two circuits together
and to verify satisfiability. The resulting inputs are
$X = \{x_1, x_2, \ldots, x_n\}$ and the outputs are
$Y = \{y_1, y_2, \ldots, y_n\}$.

The subscircuit on the left side of Figure~\ref{fig:miter}
has universal component cells only. The selector lines of all
universal component cells make the variable set $S$. The solution of
Problem~\ref{problem:component_selection} is a an assignment to all
$S$-variables. All internal variables of the requirements circuit
$\psi$ and all internal variables of the universal component cell go
in the variable set $Z$.

\begin{algorithm}[htb]
  \caption{\textsc{LabelCounter}($B, \psi$)}
  \label{alg:select_gates}

  \SetKwInOut{Input}{Input}
  \SetKwInOut{Output}{Output}

  \Input{$B$, set of Boolean functions, basis \\
         $\psi = \langle V, E, \alpha, \beta, \chi, \omega \rangle$, Boolean circuit, requirements}
  \Output{\textit{count}, integer, number of configurations}

  \BlankLine
       
  $X \gets \{\chi(v) : v \in V\} \setminus \{\star\}$ \\
  $Y \gets \{\omega(v) : v \in V\} \setminus \{\star\}$ \\

  $\textit{count} \gets 0$ \\
  $\textit{miter}, S, Z \gets \textsc{CreateMiter}(B, V, E, X, Y)$ \\
  \While{$\textit{witness} \gets \normalfont\textsc{SolveQBF}(\exists{S}\forall{X} \textit{miter})$} {
    $\textit{miter} \gets\ \textit{miter} \wedge \neg\textit{witness}$ \\
    $\textit{count} \gets\ \textit{count} + 1$
  }
  \textbf{return} $\textit{count}$
\end{algorithm}

The circuit that contains the universal cells and the requirements is
constructed by the \textsc{CreateMiter} subroutine of
Algorithm~\ref{alg:select_gates}. The function copies the
requirements circuit $\psi$ under a new name $\varphi$ ties together
each pair of corresponding primary inputs and outputs and replaces all
components in $\varphi$ with universal cells. Each universal cell
switches between components in $B$.

The method \textsc{SolveQBF} is the actual invocation of the QBF
solver. In the case of non-clausal solvers
\citep{lonsing2017depqbf,janota2016solving}, one can directly feed the
miter as an input. If the solver is clausal, a conversion to
quantified Conjunctive Normal Form is needed (CNF). This conversion
typically introduces a new set of variables that can affect the
performance of the solvers. Clausal QBF solvers benefit from
preprocessing the input formula with approaches such as in \bloqqer
\citep{biere2011blocked}. During preprocessing one should take care
that no selector variables are simplified.

The counting algorithm works by blocking solutions. This is done by
negating a solution and adding a corresponding circuit gates
(inverters, AND-gates, and an OR-gate) to the original miter. The size
of the miter grows linearly with the number of solutions.

The typical miter approach uses XOR gates to compare outputs. The two
functions are different if and only if the miter is satisfiable. This
is dual to using XNOR gates and checking for validity. Notice, that due
to the fact that the XNOR gates are connected to a constant, there is a
some constant-folding that simplifies the job of the QBF solver.
 \section{Brute-Force Circuit Counting}
\label{sec:brute_force}

One can think of Boolean circuit synthesis as having two aspects:
\begin{inparaenum}[(i)]
\item\label{inpe:topology} coming-up with a topology $G$ and
\item\label{inpe:types} determining the type of each node in $G$.
\end{inparaenum} Algorithm~\ref{alg:select_gates} solves only
(\ref{inpe:types}). Arguably, (\ref{inpe:topology}) is the more
difficult part, and in general both (\ref{inpe:topology}) and
(\ref{inpe:types}) must be solved simultaneously. In this section we
combine Algorithm~\ref{alg:select_gates} and an exhaustive
search over all possible topologies of a certain size.

Circuit design is an optimization problem: the objective is to
minimize some property such as primary input to output propagation
time or power (if the circuit is implemented electrically). The
optimization criterion depends on the use-case. The main goal of this
paper is to minimize the complexity of the circuit, i.e., the number
of components.

\begin{problem}[Optimal Circuit Design]
  \label{problem:circuit_design}

  Given a basis $B$ and a requirements circuit $\psi$, compute a
  circuit $\varphi = \langle V, E, \alpha, \beta, \chi, \omega
  \rangle$, such that $\varphi \equiv \psi$ and no other circuit
  $\varphi^\prime = \langle V^\prime, E^\prime, \alpha^\prime,
  \beta^\prime, \chi^\prime, \omega^\prime \rangle$ exists such that
  $\varphi^\prime \equiv \psi$ and $|V^\prime| < |V|$.

\end{problem}

A circuit topology, itself, has two aspects:
\begin{inparaenum}[(i)]
\item how components are connected with each other and
\item how components interface with the outside world in terms of
  primary inputs and outputs ($\chi$ and $\omega$,
  respectively).
\end{inparaenum}
This gives rise to a class of graphs that have three types of nodes:
\begin{inparaenum}[(i)]
\item primary inputs $X$,
\item primary outputs $Y$, and
\item internal nodes $Z$.
\end{inparaenum}
It is assumed that each primary input node $x \in X$ is connected to
one or more internal nodes $Z' \subseteq Z$. Each primary output
$y \in Y$ is connected to a distinct internal node $r \in Z$.

\begin{figure}[hbt]
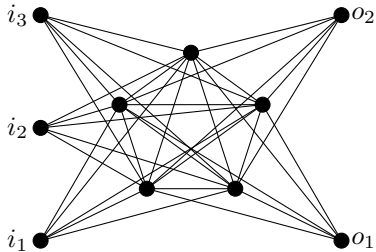

  \centering
  \includestandalone{figures/fully_connected_topology}
  \caption{The fully connected topology $K_{3, 5, 2}$}
  \label{fig:fully_connected_topology}
\end{figure}

Our first approach to solving Problem~\ref{problem:circuit_design} is
to exhaustively enumerate all possible topologies up to a certain
size. The topology that has the fully-connected graph is denoted as
$K$. A fully-connected topology where the primary inputs, outputs, and
internal nodes are partitioned is denoted as $K_{|X|, |Y|, |Z|}$,
where $|X|$ is the number of primary inputs, $|Y|$ is the number of
primary outputs and $|Z|$ is the number of internal variables. A
circuit topology of size $|V| = |X| + |Y| + |Z|$ is always a subgraph
of $K_{|X|, |Y|, |Z|}$. We can skip circuit topologies where two
primary outputs are tied together.

\begin{algorithm}[htb]
  \caption{\textsc{ExhaustiveSearch}($\psi$)}
  \label{alg:exhaustive_search}

  \SetKwInOut{Input}{Input}
  \SetKwInOut{Output}{Output}

  \Input{$B$, set of Boolean functions, basis \\
         $\psi = \langle V, E, \alpha, \beta, \chi, \omega \rangle$, Boolean circuit, requirements}
  \Output{\textit{count}, integer, number of circuits}

  \BlankLine

  $X \gets \{\chi(v) : v \in V\} \setminus \{\star\}$ \\
  $Y \gets \{\omega(v) : v \in V\} \setminus \{\star\}$ \\

  \BlankLine

  $\textit{count} \gets 0$ \\
  $n \gets 1$ \\

  \While{$\textit{count} = 0 \wedge n \le |V|$} {
    \ForAll{$\langle V^\prime, E^\prime \rangle \subseteq K_{|X|, |Y|, n}$\label{line:next_subset}} {
      $\textit{miter}, S, Z \gets \textsc{CreateMiter}(B, V^\prime, E^\prime, X, Y)$ \\
      $\gamma \gets \textsc{CircuitToCNF}(\textit{miter})$ \\
      \While{$\textit{witness} \gets \normalfont\textsc{SolveQBF}(\exists{S}\forall{X} \gamma)$} {
         $\textit{miter} \gets \textit{miter} \wedge \neg\textit{witness}$ \\
         $\textit{count} \gets \textit{count} + 1$
      }
    }
    $n \gets n + 1$\\
  }
\end{algorithm}

The number of circuit topologies of a certain size grows rapidly.  The
number of directed edges in $K_{m, n, k}$ is
$|E| = mk + nk + k(k - 1) = k(m + n + k - 1)$. This results in a total
of $2^{|E|}$ subsets. Consider the topology of the full-adder with
three primary inputs and two primary outputs. The first six elements
of the series $|2^{T_{3, k, 2}}|$ are
$2^5, 2^{12}, 2^{21}, 2^{32}, 2^{45}$, and $2^{60}$.

Algorithm~\ref{alg:exhaustive_search} is the simplest method for
circuit counting. It is guaranteed to terminate as there are
upper-bounds for the number of components and for the number of
subgraphs of $K$. Algorithm~\ref{alg:exhaustive_search} is also
guaranteed to generate a design if all components in $\psi$ correspond
to Boolean functions in the basis $B$.

Algorithm~\ref{alg:exhaustive_search} computes
designs of minimal size. The reason for that is that first all
topologies with one internal node are tried, then all topologies with
two nodes, etc.

Algorithm~\ref{alg:exhaustive_search} solves
Problem~\ref{problem:component_selection} for each candidate topology
$\langle V^\prime, E^\prime, \chi, \omega \rangle$. The number of
invocations of the QBF solver can be significantly reduced if we
consider non-isomorphic graphs only. There is no analytic approach to
enumerating all non-isomorphic graphs of size $k$, the latter is a
problem on its own. The world leaders in graph counting are
\cite{mckay2014isomorphism}.
 \section{Computing Both a Topology and the Component Types from the
  Solution of a Single 2-QBF Problem}
\label{sec:synthesize}
 
The brute-force algorithm of Sec.~\ref{sec:brute_force} is too slow.
It is possible to encode the whole circuit synthesis, both component
selection and topology generation, as a single QBF satisfiability
problem. The difficulty of generating a circuit is then left entirely
to the QBF solver. The approach is shown in
Algorithm~\ref{alg:synthesize}.

\begin{algorithm}[htb]
  \caption{\textsc{SynthesizeCircuit}($B$, $\psi$, $n$)}
  \label{alg:synthesize}

  \SetKwInOut{Input}{Input}
  \SetKwInOut{Output}{Output}
  \SetKwInOut{Local}{Local Variables~}

  \Input{$B$, set of Boolean functions, basis \\
         $\psi$, requirements (inputs $X$, outputs $Y$) \\
         $n$, integer, maximum number of components}
  \Output{$\Phi$, a set of circuits}
  \Local{$X_u$, set of variables, the inputs to all universal cells \\
         $Y_u$, set of variables, the outputs of all universal cells \\
         $S_u$, set of variables, the inputs to all universal cells \\
         $S_t = \left|s_{i, j}\right|$, matrix of variables, the interconnection fabric selectors}

  \BlankLine
  $\Phi \gets \emptyset$ \\
  \For{$k \in \{1, 2, \ldots, n\}$} {
    $\langle{S_u, X_u, Y_u, \varphi_u}\rangle \gets \textsc{UniversalCells}(B, k)$\label{line:universal_cell} \\
    $\langle{S_t, \varphi_t}\rangle \gets \textsc{InterconnectionFabric}(X_u \cup Y, X \cup Y_u)$ \\
    \For{$j \in \{1, 2, \ldots, |X \cup Y_u|\}$} {
      $\varphi_t \gets \varphi \wedge\ \textsc{ColumnCardinalityConstraint}(\{s_{1,j}, s_{2,j}, \ldots, s_{|X_u \cup Y|,j}\})$ \label{line:col_cc} \\
    }
    \For{$i \in \{1, 2, \ldots, |X_u \cup Y|\}$} {
      $\varphi_t \gets \varphi \wedge\ \textsc{RowCardinalityConstraint}(\{s_{i,1}, s_{i,2}, \ldots, s_{i,|X \cup Y_u|}\})$ \label{line:row_cc} \\
    }
    $\varphi \gets \varphi_u \cup \varphi_t$ \\
    \If{$\mathit{witness} \gets \normalfont\textsc{SolveQBF}(\exists{S_u \cup S_t}\forall{X}\varphi)$} {
      $\varphi \gets \varphi \wedge \neg\textit{witness}$ \\
      $\Phi \gets \Phi \cup \textsc{ReconstructCircuit}(\mathit{witness})$ \\
    }
  }  
  \textbf{return} $\Phi$
\end{algorithm}

Similar to Algorithm~\ref{alg:exhaustive_search},
Algorithm~\ref{alg:synthesize} first tries candidate circuits with one
components, then with two, and so on, until an equivalent circuit is
discovered. The \textsc{UniversalCell} subroutine in
line~\ref{line:universal_cell} adds $k$ universal component cells (see
Sec.~\ref{sec:component_selection}). All inputs of the $k$ universal
components are accumulated in $X_u$ and all outputs are accumulated in
$Y_u$. The selector variables for the types of components are
accumulated in $S_u$. The circuit of the universal cells is denoted as
$\varphi_u$.

\subsection{A Configurable Interconnection Fabric}

The step after adding the universal cells is to construct the circuit
$\varphi_t$ that represent the interconnection fabric (the wires
connecting the gates). Denote the elements of $X \cup Y_u$ as
$\{y_1, y_2, \ldots, y_m\}$. Consider a single input $x$ of a
universal cell. The formula
\begin{eqnarray}
  \label{formula:topological_switch}
  x \leftrightarrow \bigvee_{i = 1}^{|X \cup Y_u|}{s_i \wedge y_i}
\end{eqnarray}
\noindent
``connects'' $x$ to all possible outputs of universal cells or primary
inputs depending on the values of the selector variables $s_i$.

Eq.~(\ref{formula:topological_switch}) has to be repeated for every
possible input of a universal cell or primary output. Let us denote
those outputs as $X_u \cup Y = \{x_1, x_2, \ldots, x_n\}$. This leads
to the formula for the configurable interconnection fabric:

\begin{eqnarray}
  \label{formula:all_multiplexers}
  \bigwedge_{j = 1}^{|X_u \cup Y|}{\bigvee_{i = 1}^{|X \cup Y_u|}{x_j \leftrightarrow s_{i, j} \wedge y_i}}
\end{eqnarray}

\begin{figure}[p]
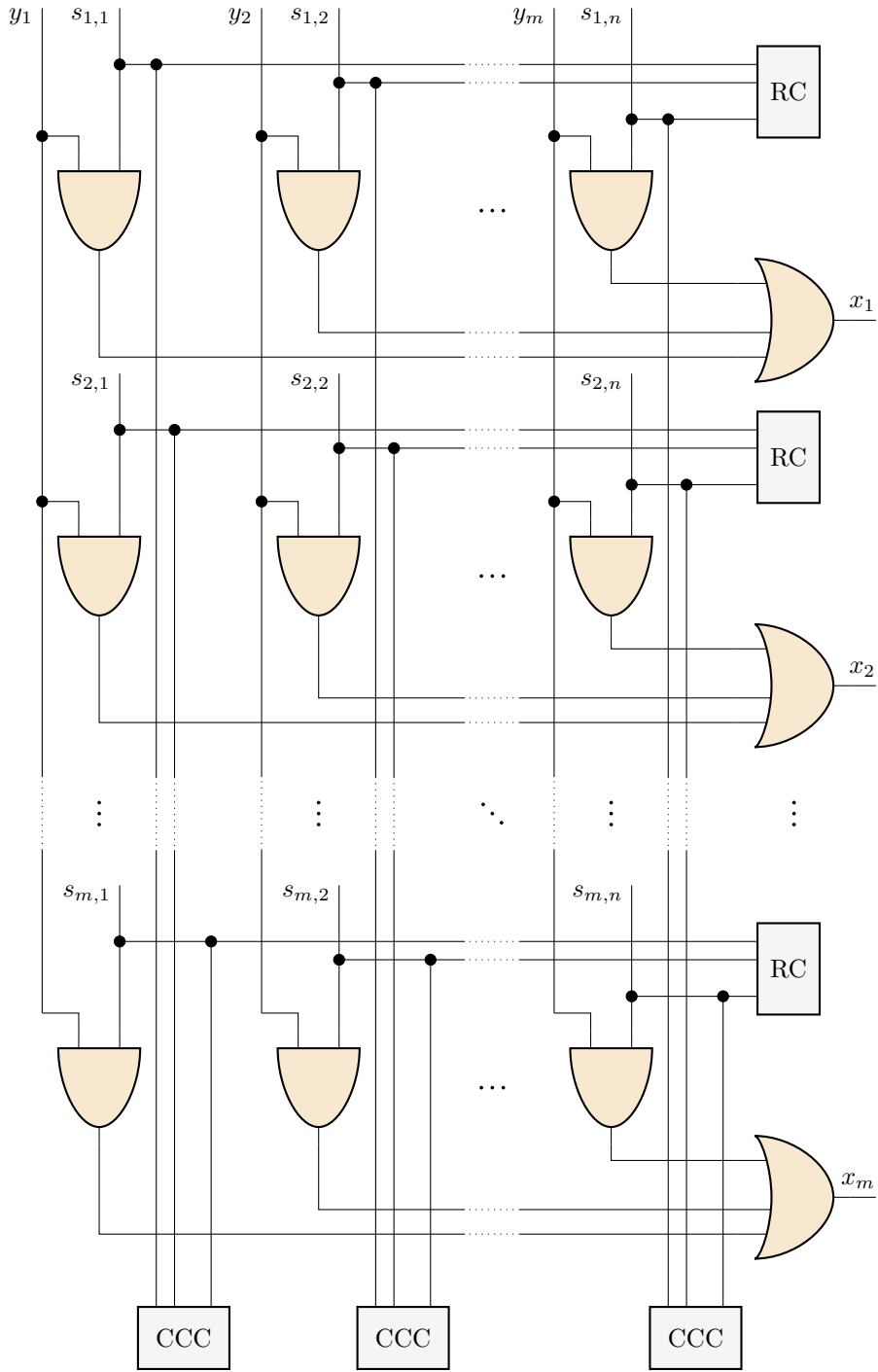

  \centering
  \includestandalone{figures/interconnection_fabric}\caption{Configurable interconnection fabric and
    cardinality constraints\label{fig:interconnection_fabric}}
\end{figure}

Most constraints of Eq.~(\ref{formula:all_multiplexers}) are
implemented as an array of two input AND-gates and are shown in
Figure~\ref{fig:interconnection_fabric}. One of the inputs of each
AND-gate from this array is connected to a selector variable
$s_{i, j}$. Each selector variable $s_{i, j}$ enables or disables the
connection of an output to an input. There is also a multi-input
OR-gate for each row of AND-gates.

\begin{figure}[hbt]
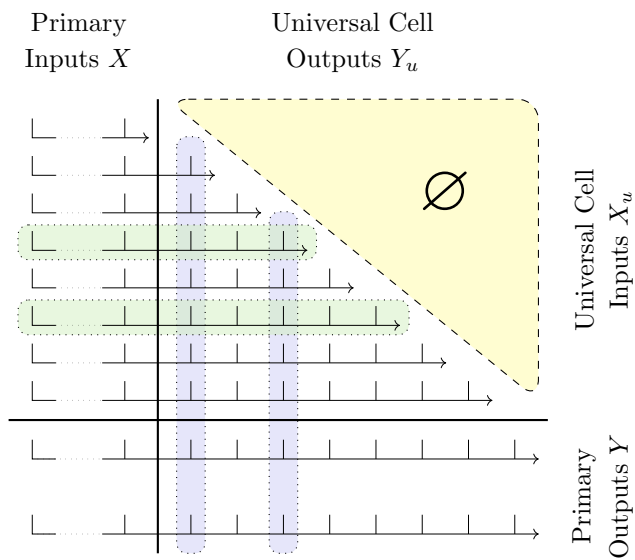

  \centering
  \includestandalone{figures/topology_constraints}
  \caption{Topology constraints\label{fig:topology_constraints}}
\end{figure}

Figure~\ref{fig:topology_constraints} illustrates the high-level
structure of the interconnection topology.

\subsection{Cardinality and Other Constraints}

An unconstrained interconnection fabric would result in malformed
circuits: loops, floating wires, wires that are not
connected to any components, etc. To avoid these malformed topologies,
Algorithm~\ref{alg:synthesize} imposes a number of constraints
(implemented as sub-circuits) on the encoding. Below is a description
of each of these constraint types.

\begin{description}
  \item[Cycle Breaking (CB):]{
      The topology selector variables in the upper right triangle of
      Fig.~\ref{fig:topology_constraints} are all disabled (assigned
      $\perp$). These constraints impose a strict ordering on the
      components and ensures that the outputs of each unversal cell
      are connected to the inputs of a successor universal cell
      only. Instead of assigning $\perp$ to the variables there, we
      simply make the multiplexers of different size and save the
      computational time for constant-folding.}
  \item[Row Cardinality Constraints (RCCs):]{
      An ``exactly-one'' constraint is added to each row of the
      connectivity matrix. These constraints can be implemented either
      with a sorting network, with a multi-operand adder (see
      Appendix~\ref{appendix:alu}), or with a combination of two-input
      AND-gates and multi-input OR-gates. The choice of the
      implementation does not affect the performance because the RCCs
      constitute a relatively small part of the encoding.}
  \item[Column Cardinality Constraints (CCCs):]{
      These constraints can be either ``at-least-one'',
      ``exactly-one'', or a combination of the two. The choice
      determines the type of the synthesized topology. The options
      that are of practical significance are:

      \begin{description}
      \item[Circuit Topology:]{
          All CCCs are of type ``at-least-one''. Notice that an
          ``at-least-one'' constraint is simply a single multi-input
          OR-gate.}
      \item[Boolean Function Topology:]{
          The first $|X|$ CCCs are of type ``at-least-one'', and the
          remaining $|Y_u|$ columns are of type ``one''. This
          combination of CCCs results in a circuit where the fan-out
          of primary inputs to gate inputs in unrestricted while the
          fan-out of each gate is restricted to one. For these
          circuits there are corresponding Boolean functions whose
          number of variables is the same as the number of primary
          inputs in the synthesized circuit. More colloquially: the
          synthesized circuit is a Boolean function.}
      \item[Network Topology:]{
          All CCCs are of type ``exactly one'', same as the RCCs. This
          topology is suitable for synthesizing sorting networks and
          reversible circuits.}
      \end{description}
  }
  \item[Unbalanced Universal Cell Ports (UUCP):]{
      The universal cell does not necessarily combine gates with the
      same number of inputs and outputs, leading to hanging wires. The
      $T_3$ constraints prevent components being connected to them. They
      are implemented as small binary multiplexers that choose which
      extra topology wires go to a pre-selected input/output of the
      universal cell.}
\end{description}

Notice that Algorithm~\ref{alg:synthesize} needs an upper-bound for
the number of components $n$. If the basis of the requirements $\psi$
is the same as the basis of the synthesis, then the number of
components $|\psi|$ can be used as an upper-bound of $n$ as an
existence of a circuit for $n = |\psi|$ is guaranteed. Otherwise, one
can use the size of a canonical form. For example in the standard
basis any formula corresponding to a circuit can be converted to
Disjunctive Normal Form (DNF). It is possible to use the size of a
circuit implementing this DNF as a value for $n$, although $n$ would
be too large. In the case of sorting networks one can take existing
upper bounds, for example, the size of the bitonic sorting network
corresponding to the desired number of inputs.

Having all this in place, we are ready to synthesize some circuits for
better understanding of Algorithm~\ref{alg:synthesize}.

\subsection{Examples of Synthesis}

\begin{figure}[hbt]
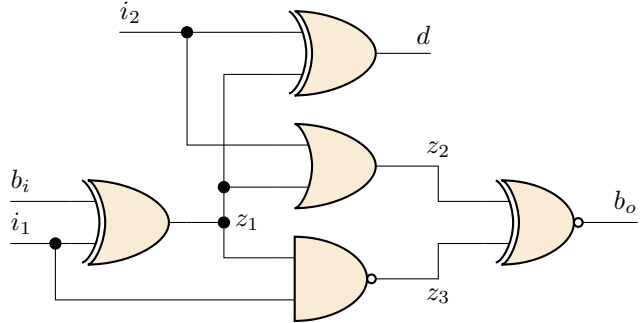

  \centering
  \includestandalone{figures/full_subtractor_alternative}
  \caption{An alternative full-subtractor\label{fig:full_subtractor_alternative}}
\end{figure}

Figure~\ref{fig:full_subtractor_alternative} shows the result of
running Algorithm~\ref{alg:synthesize} with the standard basis and the
full-subtractor shown in Figure~\ref{fig:full_subtractor} as
requirements. The generated circuit has five components only while the
one in the requirements has seven. This is a substantial saving.

\begin{figure}[hbt]
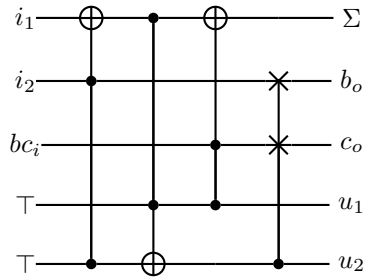

  \centering
  \includestandalone{figures/reversible_adder_subtractor}
  \caption{A reversible full-adder/subtractor\label{fig:reversible_adder_subtractor}}
\end{figure}

Another circuit designed by Algorithm~\ref{alg:synthesize} is the
reversible adder/subtractor shown in
Figure~\ref{fig:reversible_adder_subtractor}. This circuit, using one
CSWAP and three CCNOT gates, has two constant inputs and two garbage
outputs ($u_1$ and $u_2$). Synthesizing reversible circuits given
bases containing reversible gates has application to the standard
model of quantum computing as reversible circuits do not lead to
physical increase of entropy.

\begin{figure}[hbt]
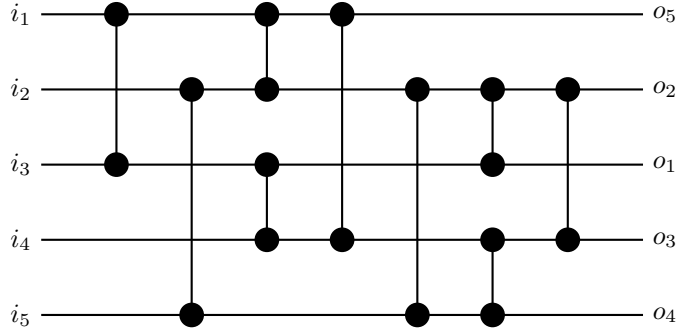

  \centering
  \includestandalone{figures/sorting_network}
  \caption{An optimal five-input sorting network\label{fig:sorting_network}}
\end{figure}

The five-input sorting network shown in
Figure~\ref{fig:sorting_network} is computed by
Algorithm~\ref{alg:synthesize}, configured with a basis containing a
comparator only. The requirements circuit $\psi$ is a bitonic sorting
network. Proving the size of the optimal sorting network for a certain
number of inputs is an open problem \cite{codish2019sorting}. Notice
that we can use Algorithm~\ref{alg:synthesize} for formally proving
minimality of a circuit. For that, we need to prove soundness,
correctness, and optimality of Algorithm~\ref{alg:synthesize} and also
to save and check all resolution-based proofs
\cite{heule2013verifying} of non-existence of circuits of size smaller
to the one synthesized.

\begin{figure}[hbt]
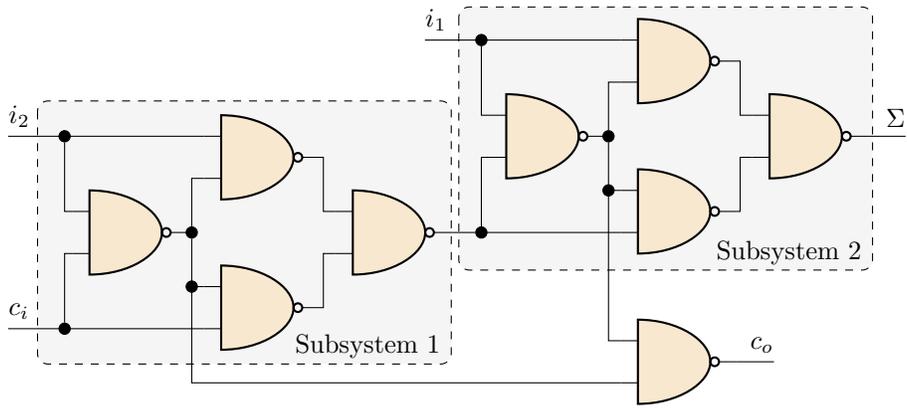

  \centering
  \includestandalone{figures/full_adder_nand}
  \caption{Classical design of a NAND-based full-adder\label{fig:full_adder_nand}}
\end{figure}

Figure~\ref{fig:full_adder_nand} shows a full-adder implemented with
NAND-gates only. The design is the classical one where the two
identical half-adder subsystems are visible.  A full-adder can also be
implemented with NOR-gates only. It has the same topology as the one
shown in Figure~\ref{fig:full_adder_nand}.

\subsection{Symmetry Breaking for Components Whose Inputs Commute}
\label{sec:symmetry_breaking}
Component libraries often have components whose inputs commute. For
example, all inputs in all components in the standard basis
commute. We automatically compute the set of all commuting component
input pairs by building small miters like the one shown in
Figure~\ref{fig:miter}.
\par
Consider a pair of commuting component inputs $x_1$ and $x_2$ and the
set $Y = \{y_1, y_2, \ldots, y_n\}$ of all possible component outputs
and primary outputs that can be connected to $x_1$ and $x_2$. There
are $2n$ selector variables responsible for connecting $x_1$ and $x_2$
to $Y$:
$s_{1, 1}, s_{1, 2}, \ldots, s_{1, n}, s_{2, 1}, s_{2, 2}, \ldots,
s_{2, n}$. We exclude symmetric topologies by adding the following
constraints:
\begin{eqnarray}
  \label{formula:ordering}
  \begin{array}{l@{\ }c@{\ }l}
  s_{2, 1} & \rightarrow & \perp                                                \\
  s_{2, 2} & \rightarrow & s_{1, 1}                                             \\
  s_{2, 3} & \rightarrow & s_{1, 1} \vee s_{1, 2}                               \\
           & \vdots      &                                                      \\
  s_{2, n} & \rightarrow & s_{1, 1} \vee s_{1, 2} \vee \cdots \vee s_{1, n - 1} \\
  \end{array}
\end{eqnarray}
As one can see, Formula~\ref{formula:ordering} essentially orders the
outputs of all possible components when they are connected to a pair
of commuting inputs. Analogous technique works for sets of commuting
inputs of arbitrary size.

\subsection{Algorithm Properties}
\label{sec:properties}

Algorithm~\ref{alg:synthesize} generates a candidate 2-QBF circuit
representing a topology of $k$ components and solves it. Analyzing
this generated circuit answers question about properties such as
soundness and completeness. Due to the applied nature of this
paper we only provide sketches instead of full proofs.

\begin{property}[Soundness]
  Given a requirements circuit $\psi$, for any circuit $\varphi$ produced by
  Algorithm~\ref{alg:synthesize}, it holds that $\psi \equiv \varphi$.
\end{property}
\begin{sketch}
  Proving this property can be done directly by analyzing the miter
  formula $\exists{S}\forall{X} \varphi \equiv \psi$. The formula is
  expanded for every possible assignment to variables in $X$:
  \begin{eqnarray}\label{eqn:expansion}
    \bigwedge_{x \in P^X}{\varphi(x) \leftrightarrow \exists{S} \psi(x)},
  \end{eqnarray}
  where $P^X$ denotes the set of all possible assignments to variables
  in $X$. We also denote as $\varphi(x)$ and $\psi(x)$ the values of
  all primary outputs of the circuits $\varphi$ and $\psi$,
  respectively, given an assignment to all their primary inputs.  We
  need each conjunct in Eq.~\ref{eqn:expansion} to be true. This means
  that we need an assignment to the $S$-variables that makes $\psi$ in
  every conjunct true if $\varphi$ is true and false otherwise.
\end{sketch}

Completeness means that if there exists a circuit that can be
synthesized from the given basis $B$, Algorithm~\ref{alg:synthesize}
is guaranteed to find it.

\begin{property}[Completeness]
  Given a requirements circuit $\psi$, Algorithm~\ref{alg:synthesize}
  is guaranteed to produce a circuit $\varphi \equiv \psi$ if such a
  circuit exists.
\end{property}
\begin{sketch}
  This can be shown with the help of a direct proof by analyzing the
  meta-circuit $\varphi$ generated by Algorithm~\ref{alg:synthesize}.

  The idea is to show that each satisfying assignment of $\varphi$
  corresponds to a circuit and that for any valid circuit and a fixed
  $k$, there exists a corresponding satisfiable assignment.

  Analyzing the 2-QBF circuit $\varphi$ generated by
  Algorithm~\ref{alg:synthesize} is too complex, so the first step is
  to split formula in two:
  \begin {enumerate*}
  \item{topology and topological constraints}, and
  \item{the universal component cells}.
  \end{enumerate*}

  Let us consider a SAT-based algorithm and a circuit $\varphi_t$ that
  generates valid topologies only (see
  Sec.~\ref{sec:brute_force}). This algorithm does not need the
  requirements circuit $\psi$ nor the universal component cells. All
  variables in $\varphi_t$ are existentially qualified, i.e.,
  $varpihi_t$ is a 1-QBF or a regular circuit. Next, we can show that
  each satisfying assignment corresponds to an unique well-formed
  topology graph. In the other direction, we have to show that each
  good topology can be the solution of $\varphi_t$.

  Each topological constraint type must be analyzed separately and all
  topological constraints must be analyzed together to shown that they
  do not allow invalid topologies and that they do not exclude valid
  ones. An invalid topology is, for example, a topology in which two
  component outputs are connected to the same input.

  Representing the topology as an adjacency matrix helps
  with the argument.

  The topology result can be next combined with correctness results of
  Algorithm~\ref{alg:select_gates} which should lead to the final
  conclusion that Algorithm~\ref{alg:synthesize} is complete.
\end{sketch}

An easy property to show is that of optimality. In this paper, the
optimization criterion is the number of components in the
circuit. In the application domain of digital design, for example,
this corresponds to power consumption. It is possible to introduce
other costs and even cost functions in which case
Algorithm~\ref{alg:synthesize} may lose its optimality or
completeness.

\begin{property}[Optimality]
  Given a requirements circuit $\psi$, Algorithm~\ref{alg:synthesize}
  is guaranteed to produce a circuit $\varphi \equiv \psi$ with
  $\varphi$ of minimal size if such a circuit exists.
\end{property}
\begin{sketch}
  Algorithm~\ref{alg:synthesize} attempts to synthesize a circuit for
  an increasing number of components $k$, starting from $k = 0$. If
  the synthesis is sound and complete, then the minimality follows
  directly from the iteration strategy for $k$.
\end{sketch}

Another property of Algorithm~\ref{alg:synthesize} is related to the
notion of universality. There are bases for which
Algorithm~\ref{alg:synthesize} is guaranteed to synthesize a circuit
equivalent to any requirements circuit $\psi$. One such basis is a
basis that contains the NAND-gate only.
 \section{Computational Complexity}

For a certain input, Problem~\ref{problem:circuit_design} becomes the
same as the Minimum Equivalent Expression (MEE) problem, classified by
\cite{buchfuhrer2011complexity}. What follows is reformulation of the
complexity results of \cite{buchfuhrer2011complexity} in the
terminology of this paper.

\begin{theorem}[Complexity of Circuit Synthesis with Fixed Basis,
  Single Output and Gate Fan-Out Restricted to One]
  \label{theorem:restricted_complexity}
Single-output circuit generation with basis
  $B = \{\neg, \wedge, \vee\}$ and gate fan-out restricted to one is
  $\Sigma_2^{P}$-complete.
\end{theorem}
\begin{sketch}
For a Boolean formula $\varphi$ with $n$ literals, there exists an
  $O(n)$ reduction from the Minimum Equivalent Expression (MEE)
  problem over signature $\{\vee, \wedge, \neg\}$. The MEE problem is
  classified as L22 in the polynomial-time hierarchy compendium
  \citep{schaefer2002completeness} and is shown to be in
  $\Sigma_2^{P}$ by \citet{buchfuhrer2011complexity}.

  The MEE problem asks if, given a Boolean formula $\varphi$ and a
  constant $k$ there exists a formula $\psi$ for which
  $\psi \equiv \varphi$, and $|\psi| < k$. The number of literals in
  $\psi$ is denoted as $|\psi|$. The circuit generation problem
  concerns generation of circuits with a minimal number of
  components. For a basis $B = \{\neg, \wedge, \vee\}$, the number of
  literals in the Boolean formula equivalent to the generated circuit
  is equal to the number of literals.
\end{sketch}

The complexity of the general circuit synthesis problem can now be
shown constructively. The idea is that we have partial input that
makes the problem $\Sigma_2^{P}$-hard. On the other hand, we have a
constructive proof (Algorithm~\ref{alg:synthesize}) that can solve
Problem~\ref{problem:circuit_design} by solving a 2-QBF. Of course we
also need soundness and completeness of the algorithm.

\begin{theorem}[Complexity of Circuit Generation]
  \label{theorem:unrestricted_complexity}
Problem~\ref{problem:circuit_design} is $\Sigma_2^{P}$-hard.
\end{theorem}
\begin{sketch}
The lower bound on the worst-time complexity comes from
  Theorem~\ref{theorem:restricted_complexity}. The upper bound comes,
  constructively, from Algorithm~\ref{alg:synthesize} as it reduces
  the problem to an $\exists\forall$ QBF.
\end{sketch}

Notice that having a basis with a NAND-gate only is equivalent to DNF
minimization which is also in $\Sigma_2^{P}$ \cite{umans2001minimum}.
 \section{Experiments}
\label{sec:experiments}

What follows is an empirical analysis of the encodings and methods
introduced in the preceding sections. The high-level miter
construction is implemented in \python while the QBF solving is in
\C/\CPP. We have compared three award-winning
\citep{janota2016qbfgallery} QBF solvers: \qfun
\citep{janota2018towards}, \rareqs \citep{janota2016solving}, and
\depqbf \citep{lonsing2017depqbf}. The \qfun QBF solver is
non-clausal. The QCNF input to \rareqs and \depqbf has been
preprocessed with \bloqqer \citep{biere2011blocked} where we had to
take special precaution not to eliminate selector variables. The
preprocessing step works by eliminating unnecessary clauses and
variables. It performs several other optimizations as well. This gives
significant speed-up.

In addition to the above three QBF solvers, we have implemented a full
expansion of the innermost universal quantifier, resulting in a SAT
problem. The SAT problem is then solved with \kissat
\citep{satcompetition2020solvers}. This approach is suitable for
problems with a smaller number of primary inputs. The resulting
expansion based 2-QBF solver is called \plq. \plq performs constant
folding after the expansion and before converting the input to CNF.

To validate the implementation of the algorithms presented in this
paper we compare with the help of a miter and a SAT solver the
equivalence of each synthesized circuit to the requirements.

All experiments were performed on a $2$-CPU (4 cores per
CPU) Intel Xeon \SI{3.3}{GHz} Linux computer with \SI{1.5}{TiB} of
RAM.

\subsection{Requirement Circuit Benchmarks}

We experiment on three types of circuits. The first type are
arithmetic and logic circuit families of variable size. The second
type are netlists from real-world ICs. Last, we take as requirements
several sets of Boolean functions from exact synthesis
\cite{haaswijk2018sat}.

\subsubsection{Arithmetic and Logic Unit Circuits}

Table~\ref{tbl:synthetic_circuits} shows a scalable synthetic set of
combinational arithmetic circuits. The size of each of the eight
synthetic circuits, described in Table~\ref{tbl:synthetic_circuits},
can be varied by setting a parameter $n$. Each variable-size circuit
shares the same topology. The carry and borrow mechanisms of adders
and subtractors, for example, have bus-like topology, while the adder
networks of the multipliers resemble two-dimensional meshes.

\begin{table}[htb]
  \begin{center}
    \begin{tabularx}{\textwidth}{llX}
      \toprule
      {Name}     & {Description}       & Role of the independent parameter $n$                              \\
      \midrule
      $n$-mux    & Multiplexer         & Number of input bits to be multiplexed, selectors are not counted  \\
      $n$-demux  & Demultiplexer       & Number of output bits                                              \\
      $n$-add    & Full-adder          & Number of inputs in one of the addends, carry input is not counted \\
      $n$-sub    & Full-subtractor     & Number of inputs in the subtrahend, borrow input is not counted    \\
      $n$-cmp    & Comparator          & Number of bits in one of the terms                                 \\
      $n$-shift  & Barrel-shifter      & Number of input bits to be shifted, selectors are not counted      \\
      $n$-moa    & Multi-operand adder & Number of input bits to be added                                   \\
      $n$-mul    & Multiplier          & Number of input bits in the multiplicand                           \\
      \bottomrule
    \end{tabularx}
  \end{center}
  \caption{Role of the $n$ parameter in the ALU-$n$ families}
  \label{tbl:synthetic_circuits}
\end{table}

\ref{appendix:alu} provides a detailed description of the
circuits in this benchmark.

We have generated two sets of circuit families for $1 \le n \le 4$ and
$1 \le n \le 32$. These two benchmark sets are called ALU-4 and
ALU-32, respectively. The former is used to benchmark synthesis while
the latter is used for evaluating the performance of gate selection.

\subsubsection{74XXX Integrated Circuits}

Table~\ref{tbl:iscas85} shows the second set of benchmark
circuits. These are reverse-engineered real-world ICs
\citep{hansen1999unveiling}. The 74XXX circuits can be chained
together into larger Arithmetic Logic Units (ALUs).

\begin{table}[htb]
  \begin{center}
    \def\inputs{\multicolumn{1}{c}{PIs}}
    \def\outputs{\multicolumn{1}{c}{POs}}
    \def\gates{\multicolumn{1}{c}{Gates}}
    \begin{tabular}{llD{.}{.}{0}D{.}{.}{0}D{.}{.}{0}}
      \toprule
      Name  & Description      & \inputs & \outputs & \gates \\
      \midrule
      74182 & 4-bit CLA        &       9 &        5 &     19 \\
      74L85 & 4-bit comparator &      11 &        3 &     33 \\
      74283 & 4-bit adder      &       9 &        5 &     36 \\
      74181 & 4-bit ALU        &      14 &        8 &     65 \\
      \bottomrule
    \end{tabular}
  \end{center}
  \caption{74XXX digital circuits}
  \label{tbl:iscas85}
\end{table}

The number of gates in the 74XXX circuits are still beyond the ability
of the synthesis algorithm. The 74XXX circuits are used for measuring
the performance of Algorithm~\ref{alg:select_gates} only.

\subsubsection{Boolean Functions from Exact Synthesis}

The performance of Algorithm~\ref{alg:synthesize} is compared to the
algorithms devised by \cite{haaswijk2018sat}. The authors of this work
have provided function sets from their study on exact synthesis and
classification
\citep{haaswijk2019sat}. Table~\ref{tbl:boolean_functions_exact_synthesis}
provides an overview of the benchmark.

\begin{table}[htb]
  \begin{center}
    \def\inputs{\multicolumn{1}{c}{PIs}}
    \def\functions{\multicolumn{1}{c}{Functions}}
    \begin{tabular}{llD{.}{.}{0}D{.}{.}{0}}
      \toprule
      Name  & Description                           & \inputs & \functions \\
      \midrule
      NPN4  & Negation-Permutation-Negation classes &       4 &        222 \\
      FDSD6 & Fully-DSD decomposable functions      &       6 &       1000 \\
      PDSD6 & Partially-DSD decomposable functions  &       6 &       1000 \\
      FDSD8 & Fully-DSD decomposable functions      &       8 &        100 \\
      PDSD8 & Partially-DSD decomposable functions  &       8 &        100 \\
      \bottomrule
    \end{tabular}
  \end{center}
  \caption{Boolean function sets from exact synthesis}
  \label{tbl:boolean_functions_exact_synthesis}
\end{table}

We have randomly selected $2422$ functions. The sizes of the subsets
are the same as in \cite{haaswijk2018sat} but the functions, with the
exceptions of the one in the NPN4 class, are different.

\subsection{Gate Selection}

\begin{figure}[htb]
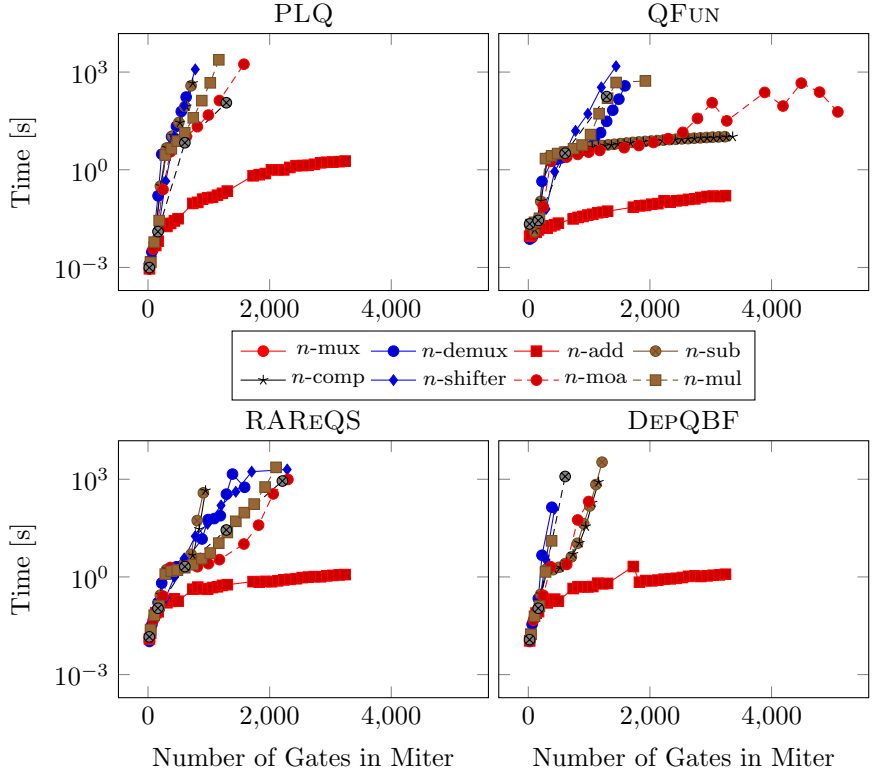

  \includestandalone{figures/select_gates_alu_tts}
  \caption{Component selection time-to-solution
    for ALU-32 circuits}
  \label{fig:select_gates_alu_tts}
\end{figure}

This section empirically evaluates the performance of
Algorithm~\ref{alg:select_gates}. The main results, shown in
Figure~\ref{fig:select_gates_alu_tts}, summarize the QBF performance
for ALU-32 and different QBF solvers. The horizontal axis shows the
number of variables in the problem and the vertical axis is the
time-to-solution.  The performance depends on the topology of the
requirements circuit and, to some extent, on the choice of the QBF
solver.

The plots in Figure~\ref{fig:select_gates_alu_tts} have logarithmic
vertical axes to accommodate the exponential
time-to-solution. Contrary to our intuition, the multiplier circuit is
not the most difficult one and the full-adder is not the easiest. The
performance is best for the demultiplexer, no matter which QBF solver
has been used. In general, the QBF solver performance is better for
large fan-outs. This can be explained with less back-tracking when
there are more outputs.

Table~\ref{tbl:select_gates_74xxx_tts} characterizes the performance
of Algorithm~\ref{alg:select_gates} on the 74XXX circuits. The table
shows the number of solutions found be each solver in
\SI{1}{\hour}. Interestingly, the only solver that found solutions for
74181 is the clausal \rareqs, despite solving a 3-QBF problem.

\begin{table}[htb]
  \centering

\includestandalone{tables/select_gates_74xxx_tts}\caption{Performance of non-clausal QBF solvers in enumerating
    component selection for 74XXX circuits}
  \label{tbl:select_gates_74xxx_tts}
\end{table}

The \qfun solver showed better performance than \plq due to the fact
that \plq spent a lot of time expanding the circuit. Of course, when
counting, there is no need to expand the circuit every time after a
solution is blocked.

\subsection{Circuit Synthesis}

The bulk of our experiments is concerned with evaluation the
performance of Algorithm~\ref{alg:synthesize}.

\subsubsection{Arithmetic and Logic Unit Circuits}

Table~\ref{tbl:synthesize_alu_4_std_optimal_size_bounds} shows the
most important data in this paper. It summarizes the performance of
the \plq and \qfun QBF solvers with and without symmetry breaking. The
bounds are on the number of components in a circuit. An upper bound
value means that Algorithm~\ref{alg:synthesize} has generated a
circuit with a certain number of gates. The lower bound values
show that Algorithm~\ref{alg:synthesize} has proven non-existence
of a circuit of a given size.

Higher values for lower bounds and lower values for upper bounds
indicate better result. The best numbers for every circuit are shown
in bold. For example, the row for the $1$-adder circuit show that all
four solver/symmetry breaking configurations prove the non-existence
of a $4$-component circuit and find a $5$-component one.

In some cases Algorithm~\ref{alg:synthesize} could fully solve a
circuit. This means that Algorithm~\ref{alg:synthesize} found a
satisfiable solution for $k$ components and showed non-satisfiability
for $m$ components, for $0 \le m \le k - 1$.  The names of the
fully-solved circuits are also shown in bold in the leftmost column of
Table~\ref{tbl:synthesize_alu_4_std_optimal_size_bounds}.

\begin{table}[p]
  \includestandalone{tables/synthesize_alu_4_std_optimal_size_bounds}\caption{Optimization performance for ALU-4 circuits and the
    standard basis}
  \label{tbl:synthesize_alu_4_std_optimal_size_bounds}
\end{table}

\begin{figure}[hbt]
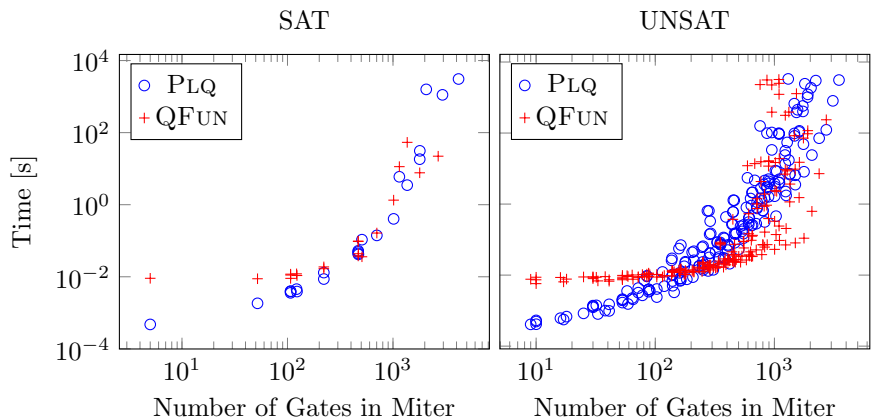

  \centering
  \includestandalone{figures/synthesize_alu_4_tts}
  \caption{Time-to-solution for various candidate circuit sizes in the
    ALU-4 benchmark}
  \label{fig:synthesize_alu_4_tts}
\end{figure}

Figure~\ref{fig:synthesize_alu_4_tts} shows the times-to-solution of
the QBF solvers for ALU-4. There is significantly more UNSAT data
because the search is from small to large number of components. The
time-to-solution increases exponentially. The most difficult calls are
just one component below the smallest circuit size. Once a circuit has
been found it becomes easier for a while and then, when increasing the
number of components the QBF solver starts timing out again.

\subsubsection{Reversible Circuits}
\label{sec:reversible_circuits}

Table~\ref{tbl:synthesize_alu_4_reversible_optimal_size_bounds}
summarizes the results for synthesizing the ALU-4 reversible circuit
from the reversible basis shown in
Fig.~\ref{fig:reversible_basis}. The data in the table should be
interpreted similarly to the data in
Table~\ref{tbl:synthesize_alu_4_std_optimal_size_bounds}, except that
there is no symmetry breaking and that in addition to number of gates,
there is also number of ancillary inputs.

\begin{table}[p]
  \includestandalone{tables/synthesize_alu_4_reversible_optimal_size_bounds}\caption{Optimization performance for ALU-4 circuits and the reversible basis}
  \label{tbl:synthesize_alu_4_reversible_optimal_size_bounds}
\end{table}

Coincidentally, what Algorithm~\ref{alg:synthesize}
could synthesize with the reversible basis, is close to what
Algorithm~\ref{alg:synthesize} could synthesize
with the standard basis. For example all eight circuits
in the multiplexer and demultiplexer families could be synthesized and
proven minimal.

Notice that $1$-moa is simply a wire and $1$-mul is a single two-input
AND-gate.

Representatives of successfully synthesized ALU-4 reversible circuits
are shown in Appendix~\ref{appendix:reversible_circuits}.

\subsubsection{Boolean Functions from Exact Synthesis}
\label{sec:exact_synthesis}

\begin{figure}[hbt]
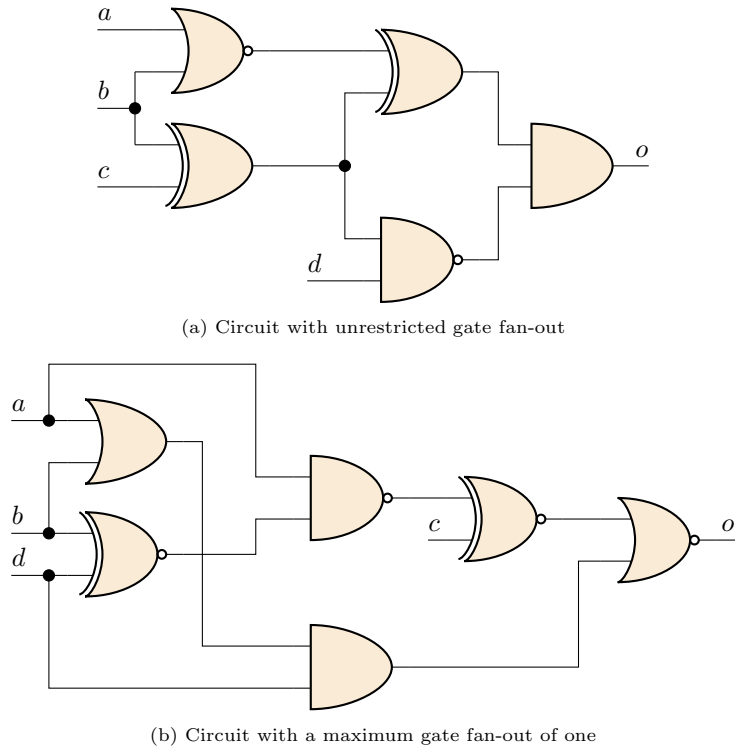

  \begin{subfigure}{\textwidth}
    \centering
    \includestandalone{figures/0x12D_circuit}\caption{Circuit with unrestricted gate fan-out}
    \label{fig:0x12D_circuit}
  \end{subfigure}

  \vspace{8pt}

  \begin{subfigure}{\textwidth}
    \centering
    \includestandalone{figures/0x12D_boolean_function}\caption{Circuit with a maximum gate fan-out of one}
    \label{fig:0x12D_boolean_function}
  \end{subfigure}

  \caption{Minimal implementations of the NPN4 Boolean function with
    truth-table 0x12D}
  \label{fig:0x12D}
\end{figure}

We next analyze the performance of Algorithm~\ref{alg:synthesize} on
the function sets from exact synthesis.  Each experiment has been
repeated twice, for two different topologies of the synthesized
circuit.  The first topology is the Boolean function one where the
gate fan-out is restricted to one. In the second set of experiments
there is no restriction on the fan-out. The difference is illustrated
in Figure~\ref{fig:0x12D_circuit} and
Figure~\ref{fig:0x12D_boolean_function}. Both figures are equivalent
to the NPN4 circuit with truth table 0x12D. Notice that the circuit in
Figure~\ref{fig:0x12D_circuit} has one gate less compared to the
circuit shown in Figure~\ref{fig:0x12D_boolean_function}.

Table~\ref{tbl:synthesize_exact_synthesis_tts_plq} summarizes the
experimental results for the \plq solver. The T/O column shows the
number of experiments in which the QBF solver timed out. The time out
for each experiment was been set-up to \SI{5}{\min}. Symmetry-breaking
was enabled. For the solved problems, we have the mean time $\mu$ in
\si{\s} and the standard deviation $\sigma$. Columns 2-5 are for
circuits with gate fan-out restricted to one. This is the same fan-out
as in the experiments of \cite{haaswijk2018sat}. Columns 6-9 are for
circuits with unrestricted gate fan-out.

\begin{table}[htb]
  \centering

\includestandalone{tables/synthesize_exact_synthesis_tts_plq}

  \caption{Solved instances by \plq and time-to-solution for Boolean
    function sets from exact synthesis}
  \label{tbl:synthesize_exact_synthesis_tts_plq}
\end{table}

Table~\ref{tbl:synthesize_exact_synthesis_tts_qfun} shows the synthesis
results for \qfun. Its layout is the same as
Table~\ref{tbl:synthesize_exact_synthesis_tts_qfun}. The performance
of \qfun is worse compared to the one of \plq.

\begin{table}[htb]
  \centering

\includestandalone{tables/synthesize_exact_synthesis_tts_qfun}

  \caption{Solved instances by \qfun and time-to-solution for Boolean
    function sets from exact synthesis}
  \label{tbl:synthesize_exact_synthesis_tts_qfun}
\end{table}

In the whole benchmark, \plq found $10$ instances in which an NPN04
circuit could be synthesized with one less gate due to allowing
unrestricted fan-out. The \qfun solver found $4$ such cases. In the
larger sets, \plq found $26$ cases for PDSD06 and $2$ for FDSD06 while
\qfun did not find any. This initial evidence shows that for the
studied function sets, unrestricted fan-out leads to a small size
reduction (one gate) in rare cases.
 \section{Related Work}

Circuit design is related to diagnostic reasoning
\citep{dekleer1987diagnosing}. Consider
Problem~\ref{problem:component_selection} and
Algorithm~\ref{alg:select_gates}. The requirements circuit $\psi$ can
be thought of as an observation. Instead of augmenting $\psi$ to
create $\phi$, as done in Algorithm~\ref{alg:select_gates}, we can
augment the buggy system description. The failure modes are ``mistaken
gate identity'', for example, the modeler has used an AND-gate in
place of an OR-gate. Algorithm~\ref{alg:select_gates} then computes
minimal changes in the system description that explain the observed
circuit.

The General Diagnostic Engine (GDE) of \cite{dekleer1987diagnosing}
can diagnose wiring errors and generate topology. When the problem is
reduced to QBF, however, it is easier to avoid ``don't cares'' by
universally quantifying the primary inputs. Combined with the
``connect to successor components only'' (see
Sec.~\ref{sec:synthesize}), our approach is more efficient in avoiding
loops and exploring the design space.

Some of the motivation for our work comes from
\citet{arthur2006evolution}. The authors of this work show that
the evolutionary design of a multi-bit adder takes significantly less
steps than anticipated. This ``ease'' made us attempt a complete
algorithm on a seemingly very difficult problem.

The problem of circuit synthesis has been first introduces by
\citet{roth1962minimization}. The authors use a very early computer,
an IBM 7090, to solve decomposition problems of four variables in
approximately ten minutes. For larger problems they propose a
heuristics that would sacrifice the algorithm completeness. Our QBF
algorithm, on the other hand, could solve problems of more than 30
variables. This was, of course, done on computers that are orders of
magnitude faster but we expect that the difficulty of the
synthesis/decomposition problems is at least in the second level of
the polynomial hierarchy \citep{stockmeyer1977polynomial}. Another
distinct advantage of our algorithm is that the
synthesis/decomposition is in terms of multi-output Boolean functions
while the paper of \citet{roth1962minimization} supports single output
functions only.

The use of the $\exists\forall\exists$-quantified miter has been
proposed for FPGA synthesis \citep{ling2005fpga}. This paper, however,
addresses the component placement problem only and does not consider
wiring, routing, and topology. Our paper demonstrates that the
combined placement/routing problem can also be solved with a single
QBF call and, thus, we have provided a fully automatic solution to the
circuit synthesis problem.

There is a large body of work on logic synthesis related to model
checking \cite{clarke2018model}. Typically this type of synthesis is
concerned with reasoning about temporal logic. \cite{bloem2014sat},
for example, uses SAT and QBF for circuit synthesis with emphasis on
safety properties.

Problem~\ref{problem:component_selection} is closely related to logic
synthesis for Filed Programmable Analog Arrays (FPGAs). FPGAs
typically consist of array of LookUp Tables (LUTs) and an
interconnection network. Programming an FPGA consists of synthesizing
the logic elements and configuring the interconnection network. There
are multiple methods for doing that \citep{cong1996combinational} but
due to the sizes of the problem, most methods are sub-optimal
\citep{cong2007optimality}.
 \section{Discussion}

Modern digital designs such as the Pentium CPUs have millions of
components. All algorithms in this paper are far from being able to
synthesize and enumerate such designs. Large Integrated Circuits
(ICs), however, are far from being optimal at the top-level. Companies
that make digital circuits integrate subsystems with the designer of
each subsystem focusing on the integrity and optimality of his or her
own subsystem. This results in globally suboptimal designs that also
have bugs, vulnerabilities and inefficiencies.

The problems we have defined are of industrial interest and create a
benchmark that is useful in the QBF competition
\citep{janota2016qbfgallery}. If accepted the benchmark will help the
QBF community to create faster QBF solvers that have practical
application. This can be achieved by noticing the structure of the
circuit design problems.

We can, at any time, sacrifice completeness and turn the algorithms
proposed in this paper into heuristic or stochastic ones. The
easiest way to do that is to replace the complete QBF search with
stochastic \citep{gent2003using}.

The algorithms in this paper can be adopted to analog designs and
design with state. The electronic designs that pose biggest challenge
and are of significant practical and theoretical interests are
hybrid. It is possible for our synthesis algorithms to work on
analogue designs by using QBF modulo theory solvers. These are similar
to satisfiability modulo theory solvers
\citep{barrett2018satisfiability} and do not exist at the time of
writing of this. The theories can be Ordinary Differential Equations
(ODEs) or Differential Algebraic Equations (DAEs). Similarly, the
algorithms of this paper, can work for geometric and physical designs
with QBF modulo Partial Differential Equations (PDEs).
 \section{Conclusion}

This paper proposes novel and generic solution to the problem of
circuit design and exploration. The problem of generating a circuit
that is equivalent to a goal is solved similar to how electronic and
logic designers solve it: first the component a chosen and placed, and
second they are connected with wires. We have given empirical evidence that
the complexity of the problem is determined, to a large extent, by
the component selection part.

We have proposed a reduction to QBF for solving a difficult
problem. We believe that this is the first practical sound and compete
algorithm for circuit design and enumeration. The built-in heuristics,
compilation and learning in the QBF solvers gives us several orders of
magnitude improvement over a baseline graph generation algorithm.

Our method is more generic than anything proposed in literature as it
considers arbitrary component libraries, such as ones consisting of
reversible gates.
 \section*{Acknowledgments}

We extend our gratitude to Matthew Klenk and John Maxwell from PARC
for many discussions and for reviewing this paper. We would also like
to thank Florian Lonsing from TU Wien for providing and supporting
\depqbf and for tutoring us on the use of QBF. Thanks to
Mikol{\'a}{\v{s}} Janota from University of Lisbon for providing
\rareqs and useful discussions. Thanks to Martina Seidl from Johannes
Kepler University for providing and supporting \bloqqer. Thanks to
Marijn Heule from The University of Texas at Austin for useful
discussion and reviewing the paper.
 
\bibliography{aij-digital-design}

\appendix
\section{The ALU Circuit Families}
\label{appendix:alu}

Table~\ref{tbl:alu_circuit_sizes} gives the number of primary inputs
(PIs), primary outputs (POs), and gates as a function of the parameter
$n$. Some of the circuits use a proxy parameter $k$ to avoid the use
of logarithms.

\begin{table}[htb]
  \begin{center}
    \begin{tabular}{llccc}
      \toprule
      Name       & Notes                     & PIs       & POs      & Gates                            \\
      \midrule
      $n$-mux    & $n = 2^k$, $k \ge 1$      & $2^k + k$ & $1$      & $2^k + k + 1$                    \\
      $n$-demux  & $n = 2^k$, $k \ge 1$      & $k + 1$   & $2^k$    & $2^k + k$                        \\
      $n$-add    & $n \ge 1$                 & $2n + 1$  & $n + 1$  & $5n$                             \\
      $n$-sub    & $n \ge 1$                 & $2n + 1$  & $n + 1$  & $7n$                             \\
      $n$-cmp    & $n \ge 1$                 & $2n$      & $3$      & $3n + 4$                         \\
      $n$-shift  & $n \ge 1$                 & $2^n + n$ & $2^n$    & $2^n(3n - 2) + n + 2$            \\
      $n$-moa    & $n = 2^k - 1$, $k \ge 2$  & $2^k - 1$ & $k$      & $2^{k + 1}(k - 2) + 2^k - k + 3$ \\
      $n$-mul    & $n \ge 2$                 & $n$       & $2n$     & $6n^2 - 8n$                      \\
      \bottomrule
    \end{tabular}
  \end{center}
  \caption{Size of the circuits in the ALU-$n$ families}
  \label{tbl:alu_circuit_sizes}
\end{table}

The multiplexer (see Figure~\ref{fig:n_mux}) is the same as the one
used in the universal component cell. The demultiplexer is similar to
the multiplexer and its architecture is shown in
Figure~\ref{fig:n_demux}. Both can be generated for an arbitrarily
sized input/output word.

\begin{figure}[htb]
\centering
  \includestandalone{figures/n_demux}\caption{Variable size demultiplexer circuit}
  \label{fig:n_demux}
\end{figure}

\begin{figure}[p]
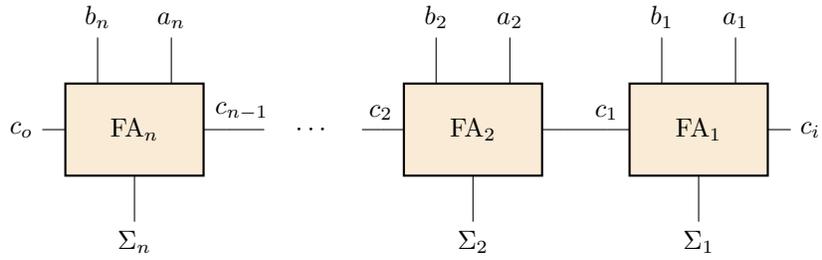
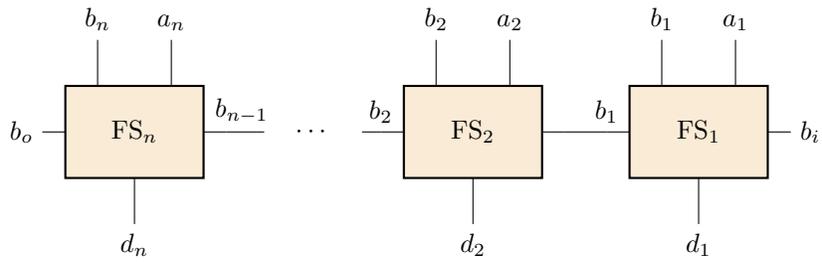
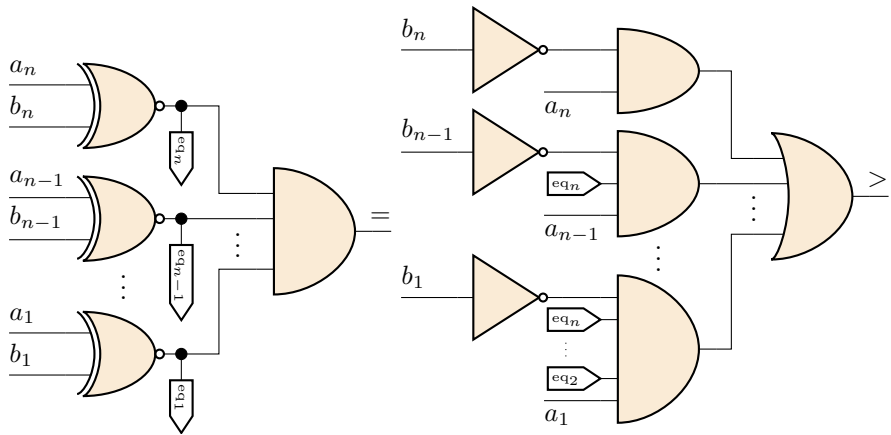
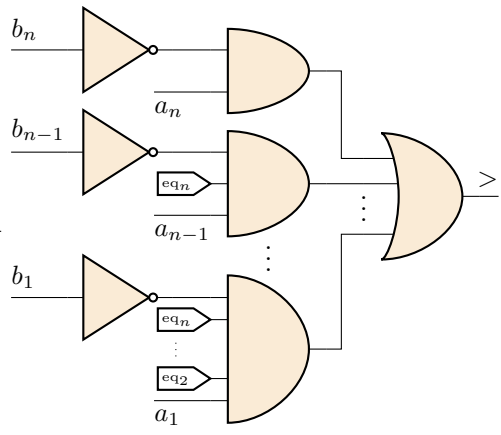

  \centering
  \begin{subfigure}{\textwidth}
    \centering
    \includestandalone{figures/n_adder}\caption{A ripple-carry adder}
    \label{fig:ripple_carry_adder}
  \end{subfigure}

  \vspace{12pt}

  \begin{subfigure}{\textwidth}
    \centering
    \includestandalone{figures/n_subtractor}\caption{A ripple-carry subtractor}
    \label{fig:ripple_carry_subtractor}
  \end{subfigure}

  \vspace{12pt}

  \begin{subfigure}[b]{0.425\textwidth}
    \centering
    \includestandalone{figures/n_comp_eq}
    \caption{Equality comparator}
    \label{fig:comp_eq}
  \end{subfigure}\begin{subfigure}[b]{0.55\textwidth}
    \centering
    \includestandalone{figures/n_comp_gt}
    \caption{``Greater-than'' comparator}
    \label{fig:comp_gt}
  \end{subfigure}
  \caption{Variable size adder, subtractor, and comparator}
  \label{fig:ripple_carry_adder_subtractor}
\end{figure}

The adder, shown in Figure~\ref{fig:ripple_carry_adder} and the
subtractor, shown in Figure~\ref{fig:ripple_carry_subtractor}, are
both ripple-carry. Due to the long propagation of carry, they are not
used in the design of modern ICs. Used as a requirements circuit and
with a sufficiently fast QBF solver Algorithm~\ref{alg:synthesize}
should be able to enumerate all parallel adders and subtractors. An
example of a real-world four-bit adder with carry look-ahead design is
the 74283 IC, which is discussed later.

The $n$-bit comparator, shown in Figure~\ref{fig:comp_eq} and
Figure~\ref{fig:comp_gt}, uses $n$ XNOR gates to check for equality,
and inverters and AND-gates to check for ``greater than''.  The ``less
than'' signal is derived from the other two outputs with the help of
an OR-gate and another inverter.

\begin{figure}[p]
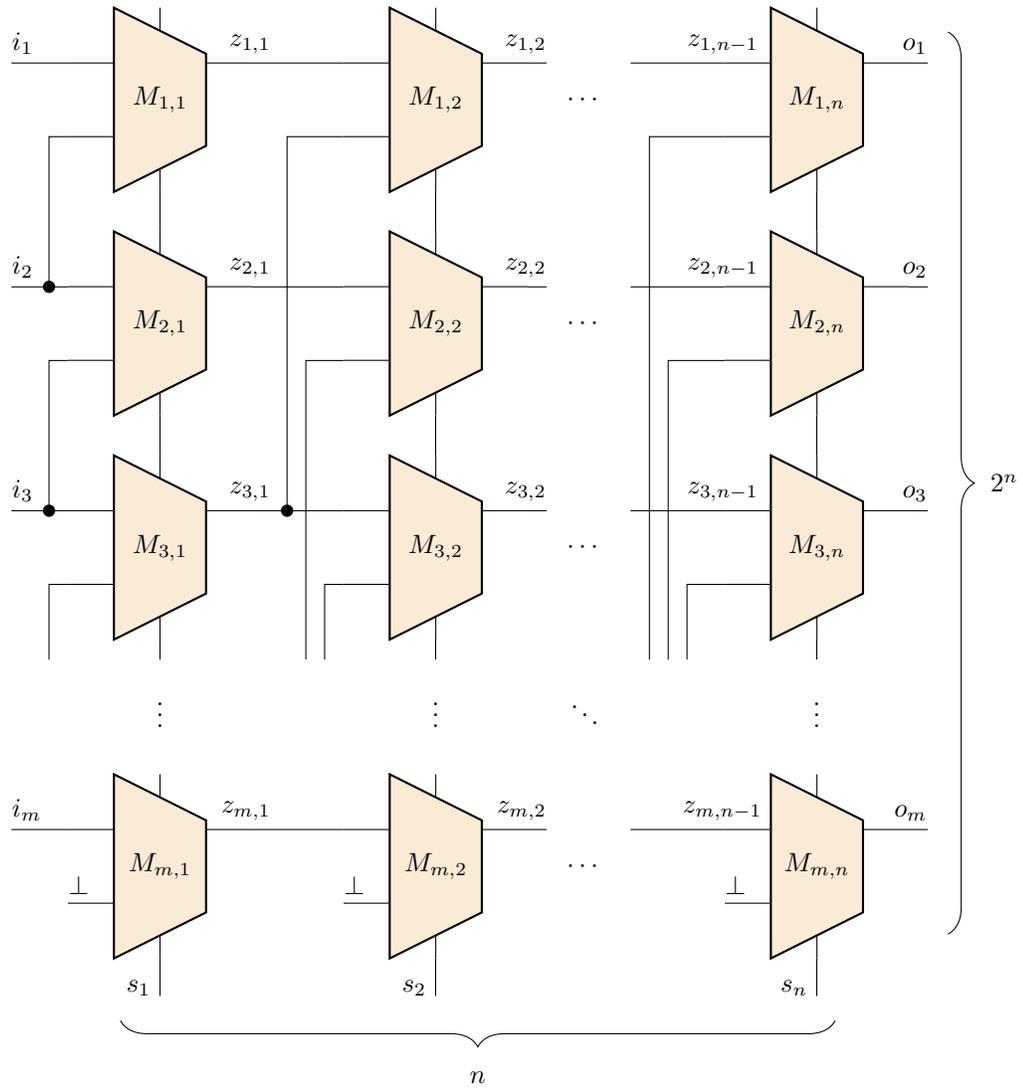

  \centering
  \includestandalone{figures/n_shifter}
  \caption{Variable size barrel-shifter}
  \label{fig:barrel_shifter}
\end{figure}

Barrel-shifters are used for shifting or rotating the bits in a
bit-word and have important application in the design Floating-Point
Units (FPUs) and cryptography cores. Figure~\ref{fig:barrel_shifter}
shows a variable-size barrel-shifter. It shifts the input word to the
right, losing the least-significant bits.

The barrel-shifter shown in Figure~\ref{fig:barrel_shifter} uses a
cascade of multiplexers with two inputs and one output. The amount of
shifting is specified as a binary number on the selector lines
$s_1, s_2, \ldots, s_n$. The total number of multiplexers is
$2^n \times n$. There are some multiplexers with an input tied to
ground on each column of the array shown in
Figure~\ref{fig:barrel_shifter}. We have $2^{n - 1}$ such multiplexers
per column where $n$ is the column number. Each such multiplexer loses
an AND-gate and an OR-gate. This reduces the number of gates as
accounted for in Table~\ref{tbl:alu_circuit_sizes}. All multiplexers
of a barrel-shifter reuse the same $n$ inverters. The inverters are
not shown in Figure~\ref{fig:barrel_shifter}.

\begin{figure}[hp]
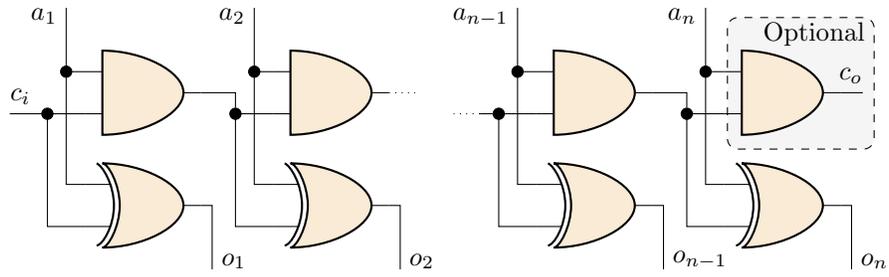
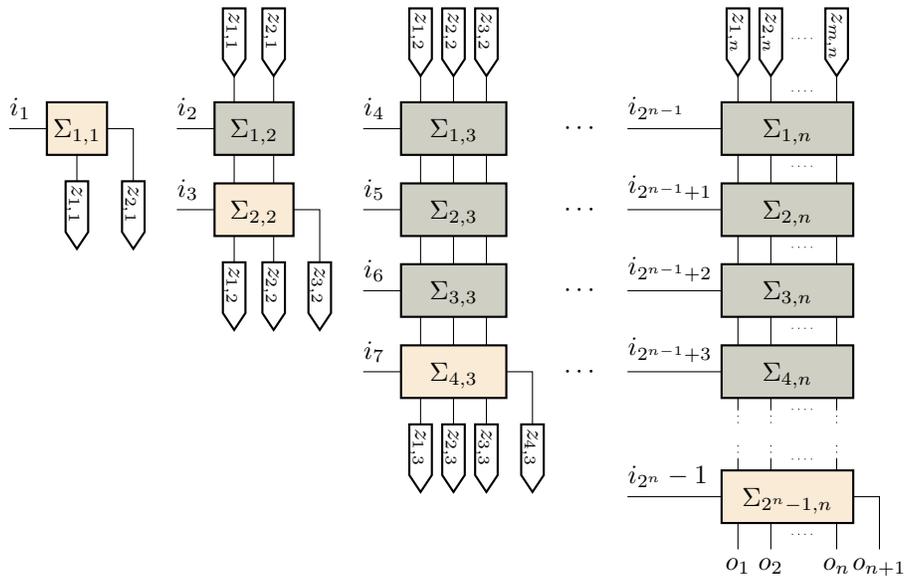

  \centering
  \begin{subfigure}{\textwidth}
    \centering
    \includestandalone{figures/ha_chain}\caption{A half-adder chain with an optional carry out bit}
    \label{fig:multi_operand_full_adder}
  \end{subfigure}
  \par\bigskip
  \begin{subfigure}{\textwidth}
    \centering
    \includestandalone{figures/n_multi_operand_adder}\caption{A ladder of half-adder-chains for multi-operand addition}
    \label{fig:multi_operand_adder_chain}
  \end{subfigure}
  \caption{A binary multi-operand adder of variable size}
  \label{fig:multi_operand_adder}
\end{figure}

The $n$-bit multi-operand adder circuit, shown in
Figure~\ref{fig:multi_operand_adder}, adds $n$ single-bit numbers. A
digital circuit that implements multi-operand addition is useful as a
stand-alone circuit and also has application in multipliers
\cite{wallace1964suggestion}.  Multi-operand addition of single-bit
numbers is also known as bit-counting or binary vector
addition. Applications of satisfiability to optimization use
bit-counting for implementing ``at-least-$k$'' or ``at-most-$k$''
constraints \citep{fu2006solving}.

The multi-operand adder is implemented as a chain of multi-operand
full-adders (see Figure~\ref{fig:multi_operand_full_adder}). Each
full-adder adds one bit to a binary number and consists of $k$
half-adders where $k$ equals the number of bits necessary for
representing the binary number (see
Figure~\ref{fig:multi_operand_full_adder}). The full-adders can be
implemented without a carry-out bit, which saves one AND-gate. The
multi-operand adder uses full-adders of increasing size. The first
adder has one input, the second and third have two inputs, the next
four have three inputs, etc.

This particular implementation of a multi-operand adder has no
application in digital electronics due to the long primary inputs to
outputs propagation time, but it is useful in constraint
programming. The chained multi-operand adder can be used as a
requirements circuit to allow the automatic discovery of advanced
topologies such as the one in \cite{wallace1964suggestion} or
\cite{dadda1965some} trees.

\begin{figure}[p]
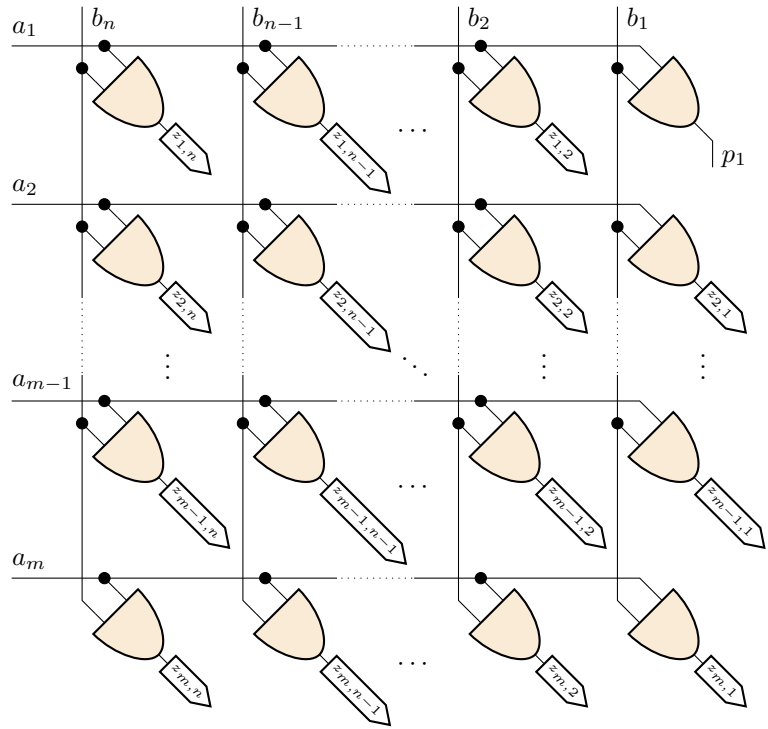
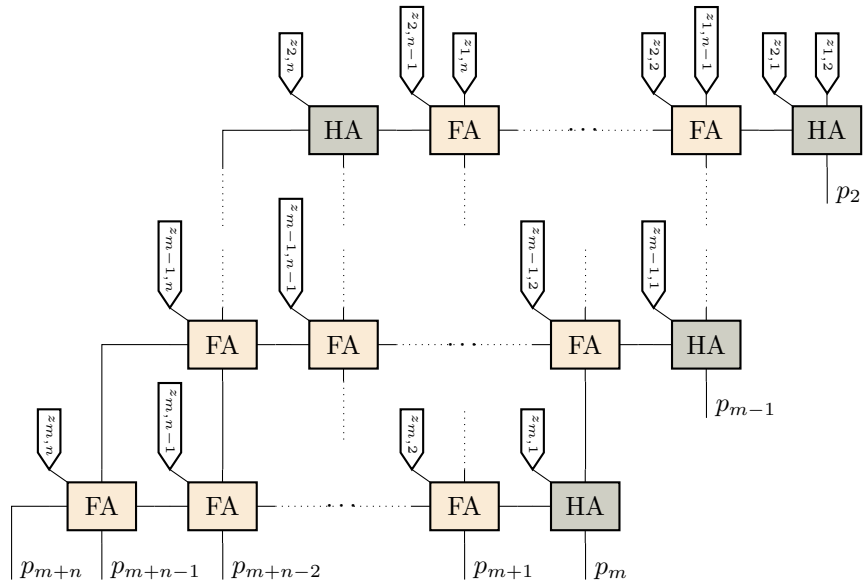

  \centering
  \begin{subfigure}{\textwidth}\centering
    \includestandalone{figures/n_mul_pp}\caption{Partial products}\label{fig:n_mul_pp}\end{subfigure}

  \vspace{4pt}

  \begin{subfigure}{\textwidth}\centering
    \includestandalone{figures/n_mul_an}\caption{Adder network}\label{fig:n_mul_an}\end{subfigure}

  \caption{Variable size multiplier}
  \label{fig:mul}
\end{figure}

Figure~\ref{fig:mul} shows the architecture of a variable size
multiplier that implements the standard ``pen and paper'' method. The
multiplier consists of two subsystems: an array of AND-gates that
computes partial products (see Figure~\ref{fig:n_mul_pp} and a network
of adders that sum the partial products (see Figure~\ref{fig:n_mul_an}).

 \section{Reversible Circuits from the ALU Families}
\label{appendix:reversible_circuits}

\begin{sidewaysfigure}[p]
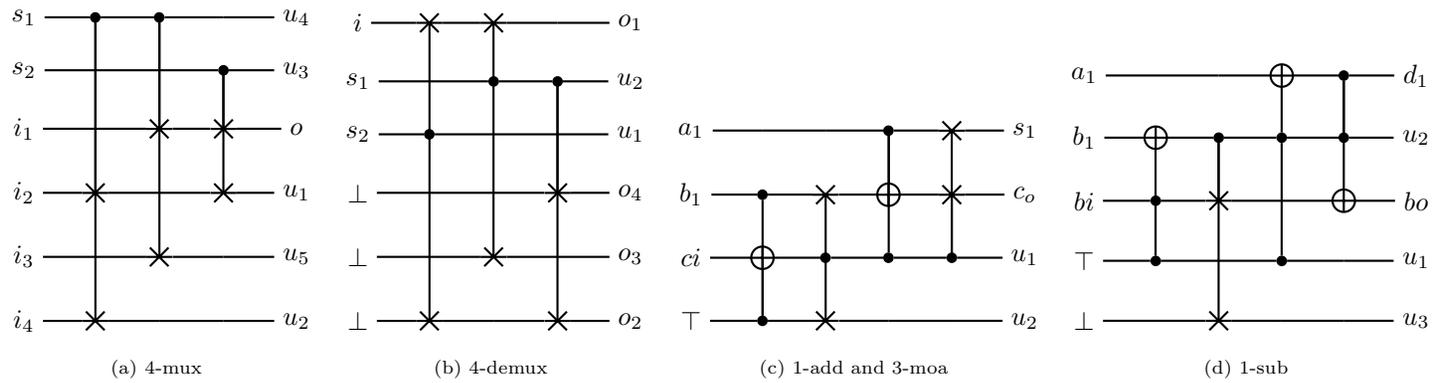
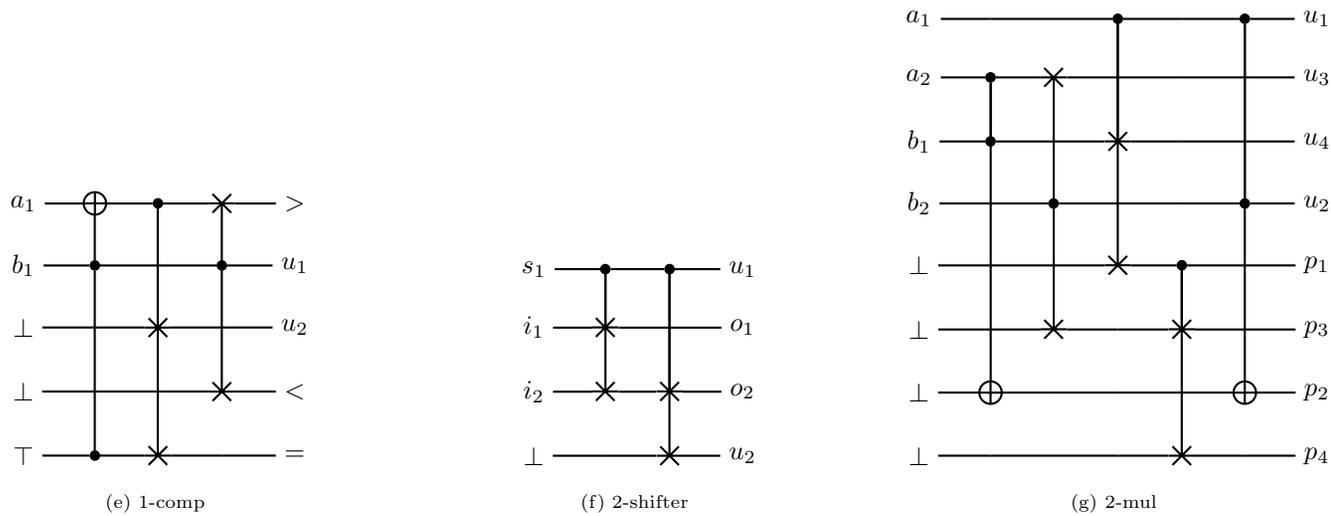

  \begin{subfigure}[b]{0.23\textwidth}
    \centering
    \includestandalone{figures/reversible_mux_4}
    \caption{$4$-mux}
    \label{fig:mux_4_3_1}
  \end{subfigure}\begin{subfigure}[b]{0.23\textwidth}
    \centering
    \includestandalone{figures/reversible_demux_4}
    \caption{$4$-demux}
    \label{fig:demux_4_3_1}
  \end{subfigure}\begin{subfigure}[b]{0.27\textwidth}
    \centering
    \includestandalone{figures/reversible_adder_1}
    \caption{$1$-add and $3$-moa}
    \label{fig:adder_1_4_1}
  \end{subfigure}\begin{subfigure}[b]{0.27\textwidth}
    \centering
    \includestandalone{figures/reversible_subtractor_1}
    \caption{$1$-sub}
    \label{fig:subtractor_1_4_1}
  \end{subfigure}

  \vspace{12pt}

  \begin{subfigure}[b]{0.33\textwidth}
    \centering
    \includestandalone{figures/reversible_comp_1}
    \caption{$1$-comp}
  \end{subfigure}\begin{subfigure}[b]{0.33\textwidth}
    \centering
    \includestandalone{figures/reversible_barrel_shifter_2}
    \caption{$2$-shifter}
  \end{subfigure}\begin{subfigure}[b]{0.33\textwidth}
    \centering
    \includestandalone{figures/reversible_mul_2}
    \caption{$2$-mul}
  \end{subfigure}\caption{Representatives of optimal reversible ALU-4 circuits}
  \label{fig:reversible_alu_4_circuits}
\end{sidewaysfigure}

Figure~\ref{fig:mux_4_3_1} shows a 4-to-1 multiplexer. Its functioning
can be verified by analyzing the circuit. It consists of three CSWAP
gates. If the two selector lines $s_1$ and $s_2$ are both low, then
there are no swapped values and the input $i_1$ is coped to the output
$o$. If the left-most CSWAP gate is activated with a value of one on
$s_1$, then $i_4$ goes to $i_2$. If the second selector $s_2$ is also
high, then the value of $i_4$ from $i_2$ will go to $i_1$ and
$o$. This is correct because when both $s_1$ and $s_2$ are selected we
expect $i_4$ to send its value to $o$. The remaining two combinations
can be checked in a similar manner.

Figure~\ref{fig:demux_4_3_1} shows a 1-to-4 demultiplexer. Similar to
the multiplexer from Figure~\ref{fig:mux_4_3_1} it is also made of
three CSWAP gates. Similar the standard basis multiplexer and
demultiplexer, the ones from the reversible basis are very similar.

Figure~\ref{fig:adder_1_4_1} and Figure~\ref{fig:subtractor_1_4_1}
show a full-adder and a full-subtractor, respectively. They are both
made of four gates but the similarities end there. A multi-operand
adder with three inputs is equivalent to a regular full-adder, hence
Figure~\ref{fig:adder_1_4_1} shows both.
 
\end{document}

%% file: tables/select_gates_74xxx_tts.tex
\begin{tabular}{lD{.}{.}{0}D{.}{.}{0}D{.}{.}{0}D{.}{.}{0}}
\toprule
Name & \textsc{PLQ} & \textsc{QFun} & \textsc{RAReQS} & \textsc{DepQBF} \\ 
\midrule
74182 & 154 & 512 & 512 & {-} \\
74L85 &  10 & 675 & 134 & {-} \\
74283 &  31 & 691 & 406 & {-} \\
74181 & {-} &   7 &  21 & {-} \\
\bottomrule
\end{tabular}

%% file: tables/synthesize_alu_4_std_optimal_size_bounds.tex
\begin{threeparttable}

\def\upperbound{\multicolumn{4}{c}{Upper Bound}}
\def\lowerbound{\multicolumn{4}{c}{Lower Bound}}
\def\plq{\multicolumn{2}{c}{\textsc{PLQ}}}
\def\qfun{\multicolumn{2}{c}{\textsc{QFun}}}
\def\name{\multicolumn{1}{c}{Name}}
\def\inputs{\multicolumn{1}{c}{PIs}}
\def\gates{\multicolumn{1}{c}{Gates}}
\def\sb{\multicolumn{1}{c}{SB}}
\def\nsb{\multicolumn{1}{c}{No SB}}

\addtolength{\tabcolsep}{-2pt}
\begin{tabular}{l
                D{.}{.}{0}
                D{.}{.}{0}
                D{.}{.}{0}
                D{.}{.}{0}
                D{.}{.}{0}
                D{.}{.}{0}
                D{.}{.}{0}
                D{.}{.}{0}
                D{.}{.}{0}
                D{.}{.}{0}}
\toprule
      &         &        & \upperbound             & \lowerbound             \\ 
      &         &        & \plq       & \qfun      & \plq       & \qfun      \\ 
\name & \inputs & \gates & \sb & \nsb & \sb & \nsb & \sb & \nsb & \sb & \nsb \\ 
\midrule
\textbf{1-mux}     & 2 &  2 &  \textbf{2} &  \textbf{2} &  \textbf{2} &  \textbf{2} &  \textbf{1} & \textbf{1} & \textbf{1} & \textbf{1} \\
\textbf{2-mux}     & 3 &  4 &  \textbf{3} &  \textbf{3} &  \textbf{3} &  \textbf{3} &  \textbf{2} & \textbf{2} & \textbf{2} & \textbf{2} \\
3-mux              & 5 &  6 &       {{-}} &       {{-}} &       {{-}} &       {{-}} &  \textbf{6} & \textbf{6} & \textbf{6} &          5 \\
4-mux              & 6 &  7 &       {{-}} &       {{-}} &       {{-}} &       {{-}} &  \textbf{6} & \textbf{6} & \textbf{6} &          5 \\
\textbf{1-demux}   & 2 &  2 &  \textbf{2} &  \textbf{2} &  \textbf{2} &  \textbf{2} &  \textbf{1} & \textbf{1} & \textbf{1} & \textbf{1} \\
\textbf{2-demux}   & 2 &  3 &  \textbf{2} &  \textbf{2} &  \textbf{2} &  \textbf{2} &  \textbf{1} & \textbf{1} & \textbf{1} & \textbf{1} \\
\textbf{3-demux}   & 3 &  5 &  \textbf{5} &  \textbf{5} &  \textbf{5} &  \textbf{5} &  \textbf{4} & \textbf{4} & \textbf{4} & \textbf{4} \\
\textbf{4-demux}   & 3 &  6 &  \textbf{6} &  \textbf{6} &  \textbf{6} &  \textbf{6} &  \textbf{5} & \textbf{5} & \textbf{5} & \textbf{5} \\
\textbf{1-add}     & 3 &  5 &  \textbf{5} &  \textbf{5} &  \textbf{5} &  \textbf{5} &  \textbf{4} & \textbf{4} & \textbf{4} & \textbf{4} \\
2-add              & 5 & 10 & \textbf{10} & \textbf{10} & \textbf{10} &       {{-}} &  \textbf{8} &          7 &          7 &          6 \\
3-add              & 7 & 15 &       {{-}} &       {{-}} &       {{-}} &       {{-}} &  \textbf{8} & \textbf{8} &          7 &          7 \\
4-add              & 9 & 20 &       {{-}} &       {{-}} &       {{-}} &       {{-}} &  \textbf{7} & \textbf{7} & \textbf{7} & \textbf{7} \\
\textbf{1-sub}     & 3 &  7 &  \textbf{5} &  \textbf{5} &  \textbf{5} &  \textbf{5} &  \textbf{4} & \textbf{4} & \textbf{4} & \textbf{4} \\
2-sub              & 5 & 14 & \textbf{10} & \textbf{10} &          13 & \textbf{10} &  \textbf{7} & \textbf{7} & \textbf{7} & \textbf{7} \\
3-sub              & 7 & 21 & \textbf{15} &       {{-}} &       {{-}} &       {{-}} &  \textbf{8} & \textbf{8} &          6 &          6 \\
4-sub              & 9 & 28 &       {{-}} &       {{-}} &       {{-}} &       {{-}} &  \textbf{8} &          7 &          7 &          6 \\
\textbf{1-comp}    & 2 &  5 &  \textbf{3} &  \textbf{3} &  \textbf{3} &  \textbf{3} &  \textbf{2} & \textbf{2} & \textbf{2} & \textbf{2} \\
\textbf{2-comp}    & 4 & 10 &  \textbf{8} &  \textbf{8} &  \textbf{8} &  \textbf{8} &  \textbf{7} & \textbf{7} & \textbf{7} & \textbf{7} \\
3-comp             & 6 & 13 & \textbf{13} & \textbf{13} &       {{-}} &       {{-}} &  \textbf{8} &          7 &          7 &          7 \\
4-comp             & 8 & 16 &       {{-}} &       {{-}} &       {{-}} &       {{-}} &  \textbf{8} &          7 &          7 &          6 \\
\textbf{1-shifter} & 2 &  2 &  \textbf{2} &  \textbf{2} &  \textbf{2} &  \textbf{2} &  \textbf{1} & \textbf{1} & \textbf{1} & \textbf{1} \\
\textbf{2-shifter} & 3 &  5 &  \textbf{5} &  \textbf{5} &  \textbf{5} &  \textbf{5} &  \textbf{4} & \textbf{4} & \textbf{4} & \textbf{4} \\
3-shifter          & 5 & 14 & \textbf{11} &          13 &       {{-}} &          14 &  \textbf{7} & \textbf{7} & \textbf{7} &          6 \\
4-shifter          & 6 & 20 &       {{-}} &       {{-}} &       {{-}} &       {{-}} &  \textbf{8} & \textbf{8} &          7 &          7 \\
1-moa              & 1 &  1 &  \textbf{0} &  \textbf{0} &  \textbf{0} &  \textbf{0} &  \textbf{1} &      {{-}} & \textbf{1} &      {{-}} \\
\textbf{2-moa}     & 2 &  2 &  \textbf{2} &  \textbf{2} &  \textbf{2} &  \textbf{2} &  \textbf{1} & \textbf{1} & \textbf{1} & \textbf{1} \\
\textbf{3-moa}     & 3 &  5 &  \textbf{5} &  \textbf{5} &  \textbf{5} &  \textbf{5} &  \textbf{4} & \textbf{4} & \textbf{4} & \textbf{4} \\
4-moa              & 4 &  9 &  \textbf{9} &  \textbf{9} &  \textbf{9} &  \textbf{9} &  \textbf{7} & \textbf{7} & \textbf{7} &          6 \\
\textbf{1-mul}     & 2 &  1 &  \textbf{1} &  \textbf{1} &  \textbf{1} &  \textbf{1} &  \textbf{0} & \textbf{0} & \textbf{0} & \textbf{0} \\
\textbf{2-mul}     & 4 &  8 &  \textbf{7} &  \textbf{7} &  \textbf{7} &  \textbf{7} &  \textbf{6} & \textbf{6} & \textbf{6} & \textbf{6} \\
3-mul              & 6 & 30 &       {{-}} &       {{-}} &       {{-}} &       {{-}} &  \textbf{9} & \textbf{9} &          8 &          8 \\
4-mul              & 8 & 64 &       {{-}} &       {{-}} &       {{-}} &       {{-}} & \textbf{11} &         10 &          9 &          8 \\
\bottomrule
\end{tabular}
\addtolength{\tabcolsep}{2pt}
\end{threeparttable}

%% file: tables/synthesize_alu_4_reversible_optimal_size_bounds.tex
\begin{threeparttable}

\def\upperbound{\multicolumn{4}{c}{Upper Bound}}
\def\lowerbound{\multicolumn{4}{c}{Lower Bound}}
\def\plq{\multicolumn{2}{c}{\textsc{PLQ}}}
\def\qfun{\multicolumn{2}{c}{\textsc{QFun}}}
\def\name{\multicolumn{1}{c}{Name}}
\def\inputs{\multicolumn{1}{c}{PIs}}
\def\gates{\multicolumn{1}{c}{Gates}}
\def\garbage{\multicolumn{1}{c}{Ancil.}}

\addtolength{\tabcolsep}{-3pt}
\begin{tabular}{l
                D{.}{.}{0}
                D{.}{.}{0}
                D{.}{.}{0}
                D{.}{.}{0}
                D{.}{.}{0}
                D{.}{.}{0}
                D{.}{.}{0}
                D{.}{.}{0}
                D{.}{.}{0}
                D{.}{.}{0}}
\toprule
      &         &        & \upperbound                           & \lowerbound                           \\ 
      &         &        & \plq              & \qfun             & \plq              & \qfun             \\ 
\name & \inputs & \gates & \gates & \garbage & \gates & \garbage & \gates & \garbage & \gates & \garbage \\ 
\midrule
\textbf{1-mux}     & 2 &  2 & \textbf{2} & \textbf{1} & \textbf{2} & \textbf{1} & \textbf{2} & \textbf{0} &  \textbf{2} & \textbf{0} \\
\textbf{2-mux}     & 3 &  4 & \textbf{2} & \textbf{0} & \textbf{2} & \textbf{0} & \textbf{1} & \textbf{2} &  \textbf{1} & \textbf{2} \\
\textbf{3-mux}     & 5 &  6 & \textbf{3} & \textbf{1} & \textbf{3} & \textbf{1} & \textbf{3} & \textbf{0} &  \textbf{3} & \textbf{0} \\
\textbf{4-mux}     & 6 &  7 & \textbf{3} & \textbf{0} & \textbf{3} & \textbf{0} & \textbf{2} & \textbf{4} &  \textbf{2} & \textbf{4} \\
\textbf{1-demux}   & 2 &  2 & \textbf{2} & \textbf{1} & \textbf{2} & \textbf{1} & \textbf{2} & \textbf{0} &  \textbf{2} & \textbf{0} \\
\textbf{2-demux}   & 2 &  3 & \textbf{2} & \textbf{1} & \textbf{2} & \textbf{1} & \textbf{2} & \textbf{0} &  \textbf{2} & \textbf{0} \\
\textbf{3-demux}   & 3 &  5 & \textbf{3} & \textbf{3} & \textbf{3} & \textbf{3} & \textbf{3} & \textbf{2} &  \textbf{3} & \textbf{2} \\
\textbf{4-demux}   & 3 &  6 & \textbf{3} & \textbf{3} & \textbf{3} & \textbf{3} & \textbf{3} & \textbf{2} &  \textbf{3} & \textbf{2} \\
\textbf{1-add}     & 3 &  5 & \textbf{4} & \textbf{1} & \textbf{4} & \textbf{1} & \textbf{4} & \textbf{0} &  \textbf{4} & \textbf{0} \\
2-add              & 5 & 10 &      {{-}} &      {{-}} &      {{-}} &      {{-}} & \textbf{5} & \textbf{5} &           4 &          8 \\
3-add              & 7 & 15 &      {{-}} &      {{-}} &      {{-}} &      {{-}} &          4 &          4 &  \textbf{4} & \textbf{8} \\
4-add              & 9 & 20 &      {{-}} &      {{-}} &      {{-}} &      {{-}} &          3 &          4 &  \textbf{4} & \textbf{8} \\
\textbf{1-sub}     & 3 &  7 & \textbf{4} & \textbf{2} & \textbf{4} & \textbf{2} & \textbf{4} & \textbf{1} &  \textbf{4} & \textbf{1} \\
2-sub              & 5 & 14 &      {{-}} &      {{-}} &      {{-}} &      {{-}} & \textbf{6} & \textbf{1} &           5 &          0 \\
3-sub              & 7 & 21 &      {{-}} &      {{-}} &      {{-}} &      {{-}} & \textbf{5} & \textbf{3} &           5 &          1 \\
4-sub              & 9 & 28 &      {{-}} &      {{-}} &      {{-}} &      {{-}} &          3 &          5 &  \textbf{7} & \textbf{0} \\
\textbf{1-comp}    & 2 &  5 & \textbf{3} & \textbf{3} & \textbf{3} & \textbf{3} & \textbf{3} & \textbf{2} &  \textbf{3} & \textbf{2} \\
2-comp             & 4 & 10 &      {{-}} &      {{-}} &      {{-}} &      {{-}} & \textbf{6} & \textbf{1} &           5 &          1 \\
3-comp             & 6 & 13 &      {{-}} &      {{-}} &      {{-}} &      {{-}} &          5 &          5 &  \textbf{8} & \textbf{0} \\
4-comp             & 8 & 16 &      {{-}} &      {{-}} &      {{-}} &      {{-}} &          4 &          2 & \textbf{16} & \textbf{0} \\
\textbf{1-shifter} & 2 &  2 & \textbf{2} & \textbf{1} & \textbf{2} & \textbf{1} & \textbf{2} & \textbf{0} &  \textbf{2} & \textbf{0} \\
\textbf{2-shifter} & 3 &  5 & \textbf{2} & \textbf{1} & \textbf{2} & \textbf{1} & \textbf{2} & \textbf{0} &  \textbf{2} & \textbf{0} \\
3-shifter          & 5 & 14 &      {{-}} &      {{-}} &      {{-}} &      {{-}} & \textbf{5} & \textbf{7} &           5 &          0 \\
4-shifter          & 6 & 20 &      {{-}} &      {{-}} &      {{-}} &      {{-}} & \textbf{5} & \textbf{3} &           4 &          8 \\
\textbf{1-moa}     & 1 &  1 & \textbf{0} & \textbf{0} & \textbf{0} & \textbf{0} &      {{-}} &      {{-}} &       {{-}} &      {{-}} \\
\textbf{2-moa}     & 2 &  2 & \textbf{2} & \textbf{2} & \textbf{2} & \textbf{2} & \textbf{2} & \textbf{1} &  \textbf{2} & \textbf{1} \\
\textbf{3-moa}     & 3 &  5 & \textbf{4} & \textbf{1} & \textbf{4} & \textbf{1} & \textbf{4} & \textbf{0} &  \textbf{4} & \textbf{0} \\
4-moa              & 4 &  9 &      {{-}} &      {{-}} &      {{-}} &      {{-}} & \textbf{6} & \textbf{1} &           5 &          0 \\
\textbf{1-mul}     & 2 &  1 &      {{-}} &      {{-}} &      {{-}} &      {{-}} & \textbf{1} & \textbf{2} &  \textbf{1} & \textbf{2} \\
\textbf{2-mul}     & 4 &  8 & \textbf{5} & \textbf{4} &      {{-}} &      {{-}} & \textbf{5} & \textbf{3} &           5 &          2 \\
3-mul              & 6 & 30 &      {{-}} &      {{-}} &      {{-}} &      {{-}} & \textbf{5} & \textbf{4} &           5 &          1 \\
4-mul              & 8 & 64 &      {{-}} &      {{-}} &      {{-}} &      {{-}} &          3 &          5 &  \textbf{9} & \textbf{0} \\
\bottomrule
\end{tabular}
\addtolength{\tabcolsep}{3pt}
\end{threeparttable}

%% file: tables/synthesize_exact_synthesis_tts_plq.tex
\def\bf{\multicolumn{4}{c}{Fan-Out $= 1$}}
\def\circuits{\multicolumn{4}{c}{Fan-Out $\ge 1$}}

\addtolength{\tabcolsep}{-1.5pt}
\begin{tabular}{lD{.}{.}{0}D{.}{.}{0}D{.}{.}{2}D{.}{.}{2}D{.}{.}{0}D{.}{.}{0}D{.}{.}{2}D{.}{.}{2}}
\toprule
     &                                    \bf        & \circuits                                     \\ 
Name & \textrm{Solved} & \textrm{T/O} & \mu & \sigma & \textrm{Solved} & \textrm{T/O} & \mu & \sigma \\ 
\midrule
NPN04  & 221 &   1 &    3.3 &  6.67 & 222 &   0 &  15.58 & 49.76 \\
PDSD06 & 922 &  78 &  47.64 & 44.62 & 908 &  92 &  66.51 & 68.37 \\
FDSD06 & 999 &   1 &  10.53 & 16.64 & 999 &   1 &  11.85 & 24.45 \\
PDSD08 &   0 & 100 &    {-} &   {-} &   0 & 100 &    {-} &   {-} \\
FDSD08 &  53 &  47 & 190.75 & 39.62 &  62 &  38 & 167.85 & 60.48 \\
\bottomrule
\end{tabular}
\addtolength{\tabcolsep}{1.5pt}

%% file: tables/synthesize_exact_synthesis_tts_qfun.tex
\def\bf{\multicolumn{4}{c}{Fan-Out $= 1$}}
\def\circuits{\multicolumn{4}{c}{Fan-Out $\ge 1$}}

\addtolength{\tabcolsep}{-1.5pt}
\begin{tabular}{lD{.}{.}{0}D{.}{.}{0}D{.}{.}{2}D{.}{.}{2}D{.}{.}{0}D{.}{.}{0}D{.}{.}{2}D{.}{.}{2}}
\toprule
     &                                    \bf        & \circuits                                     \\ 
Name & \textrm{Solved} & \textrm{T/O} & \mu & \sigma & \textrm{Solved} & \textrm{T/O} & \mu & \sigma \\ 
\midrule
NPN04  & 208 &  14 &   9.1 &  39.9 & 207 &  15 & 10.37 & 16.27 \\
PDSD06 & 498 & 502 & 46.93 & 57.16 & 474 & 526 & 56.81 &  62.4 \\
FDSD06 & 941 &  59 & 10.04 & 22.19 & 938 &  62 & 11.88 & 29.09 \\
PDSD08 &   0 & 100 &   {-} &   {-} &   0 & 100 &   {-} &   {-} \\
FDSD08 &  41 &  59 & 59.38 & 72.14 &  43 &  57 & 44.72 & 65.29 \\
\bottomrule
\end{tabular}
\addtolength{\tabcolsep}{1.5pt}